\documentclass[10pt,twocolumn,letterpaper]{article}

\usepackage{wacv}     

\usepackage{multirow}
\usepackage{colortbl}
\usepackage{xcolor}
\usepackage{booktabs}
\usepackage{amsmath}
\usepackage{amssymb}
\usepackage{mathtools}
\usepackage{bm}
\usepackage{graphicx}
\usepackage{subcaption}
\usepackage{makecell}
\usepackage{enumitem}
\usepackage{pifont}
\usepackage[normalem]{ulem}
\usepackage[misc]{ifsym}

\newcommand{\fref}[1]{Fig.~\ref{#1}}

\newcommand{\sref}[1]{Section~\ref{#1}}

\usepackage{tikz}
\definecolor{wacvblue}{rgb}{0.21,0.49,0.74}
\usepackage[pagebackref,breaklinks,colorlinks,allcolors=wacvblue]{hyperref}
\usepackage{xcolor}

\definecolor{wacvblue}{rgb}{0.21,0.49,0.74}
\definecolor{wacvgreen}{rgb}{0.20,0.60,0.30}
\definecolor{wacvred}{rgb}{0.80,0.20,0.20}

\def\wacvPaperID{1913} 
\def\confName{WACV}
\def\confYear{2027}

\title{Where Can We Fool the Model's Mind? Response-Guided Semantic Probing and Adaptive Routing for Transferable Attacks on Black-Box MLLMs}

\author{Ahmed Abdelnaby and Mohamed Elmahallawy\thanks{Corresponding author}\\
School of Engineering and Applied Sciences, Washington State University, WA 99354, USA\\
{\tt\small \{ahmed.abdelnaby,mohamed.elmahallawy\}@wsu.edu}\\ 
{\centering\mbox{%
          \begin{tikzpicture}[baseline=-0.35ex, scale=0.32]
            \draw[fill=red, draw=red, line join=round, line width=1.5pt] 
              (0,0.85) -- (-0.85,-0.55) -- (0.85,-0.55) -- cycle;
            \node[white, font=, inner sep=0pt] at (0,-0.05) {!};
          \end{tikzpicture}%
          \hspace{0.35em}%
          \textcolor{red}{\textbf{\itshape\fontsize{11.3}{7.5}\selectfont  Warning: This paper contains sensitive or harmful examples used solely for research and evaluation.}}}}}     

\usepackage{xcolor}     
 
\usepackage{atbegshi}  

\makeatletter

\newcommand{\FirstPageBannerText}{%
  \parbox{\textwidth}{\large{\bf\em
    Appears at The IEEE/CVF Winter Conference on Applications of Computer Vision (WACV 2027)}}%
}

\AtBeginShipoutFirst{%
      \raisebox{-0.5\baselineskip}{\FirstPageBannerText}
}
\makeatother

\begin{document}

\maketitle

\begin{abstract}
Multimodal Large Language Models (MLLMs) ground language in visual evidence, yet their semantic reasoning can be exploited through transferable black-box adversarial attacks. Existing attacks typically allocate perturbations using surrogate-model saliency, which may not reflect how an unseen MLLM semantically interprets or relies on different image regions. We introduce \textbf{VISTA} (\textbf{V}ulnerability-\textbf{I}nformed \textbf{S}emantic \textbf{T}ransfer \textbf{A}ttacks), a black-box framework that identifies semantically vulnerable visual regions from an MLLM's responses without access to gradients, logits, parameters, or internal representations. VISTA supports two attack settings: (i) \emph{VISTA-Proxy}, a \textbf{\emph{pure-transfer}} setting that probes an open-source MLLM and transfers the resulting adversarial examples to unseen commercial MLLMs without querying them; and (ii) \emph{VISTA-Victim}, a \textbf{\emph{query-assisted}} setting that probes the target MLLM to capture model-specific semantic vulnerabilities. VISTA builds an Uncertainty-Based Multimodal Behavior (UBMB) map from response drift, conflict, misleading, and semantic shifts, then uses entropy-adaptive Sinkhorn routing and a surrogate ensemble to focus perturbations on vulnerable regions. Across GPT-5, Gemini-3.5-Flash, and Claude Sonnet~4.6, VISTA achieves up to 97\% attack success rate (ASR) while maintaining a Peak Signal-to-Noise Ratio (PSNR) above 30\,dB. Notably, \emph{VISTA-Proxy} achieves up to 93\% ASR without querying the final victim, demonstrating that response-derived semantic vulnerability can enable effective transfer across heterogeneous MLLMs.

\end{abstract}

\section{Introduction} \label{sec:intro}

\begin{figure}[t]
    \centering
    \includegraphics[
        width=0.49\textwidth ]{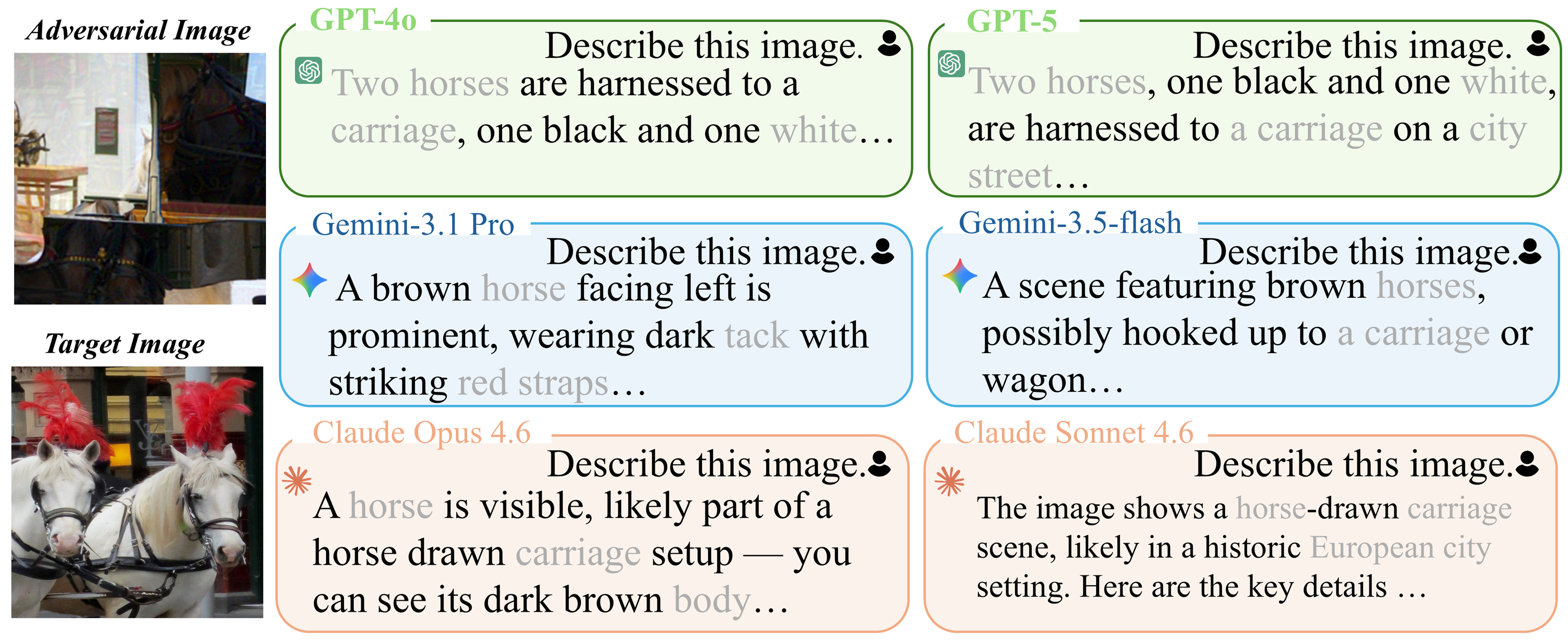}
    \caption{Cross-model adversarial transfer: a single adversarial image induces consistent target-scene descriptions across commercial MLLMs, demonstrating strong black-box transferability.}
    \label{fig:horse_case_study}
\end{figure}

Multimodal Large Language Models (MLLMs) have demonstrated strong capabilities
in visual understanding and semantic reasoning
\cite{li2023blip,liu2023visual}, yet remain vulnerable to adversarial
perturbations that alter their visual interpretations
\cite{schlarmann2023adversarial,nie2025v}. Attacking commercial MLLMs is
particularly challenging because their parameters, gradients, logits, and
internal representations are inaccessible. Black-box attacks therefore often
rely on \emph{adversarial transferability}, optimizing perturbations on
accessible white-box surrogate models and transferring them to unseen targets
\cite{zhao2023evaluating,chen2024rethinking}. As shown in
Fig.~\ref{fig:horse_case_study}, a single visually subtle adversarial image can
induce multiple commercial MLLMs to produce descriptions consistent with the
same target semantics.

\begin{figure}[t]
\centering
\resizebox{\columnwidth}{!}{%
\begin{tabular}{ccccc}
\includegraphics[width=0.19\linewidth]{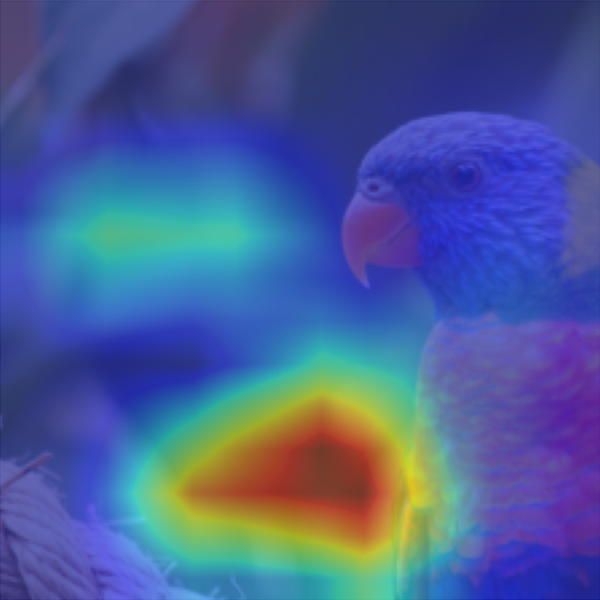} &
\includegraphics[width=0.19\linewidth]{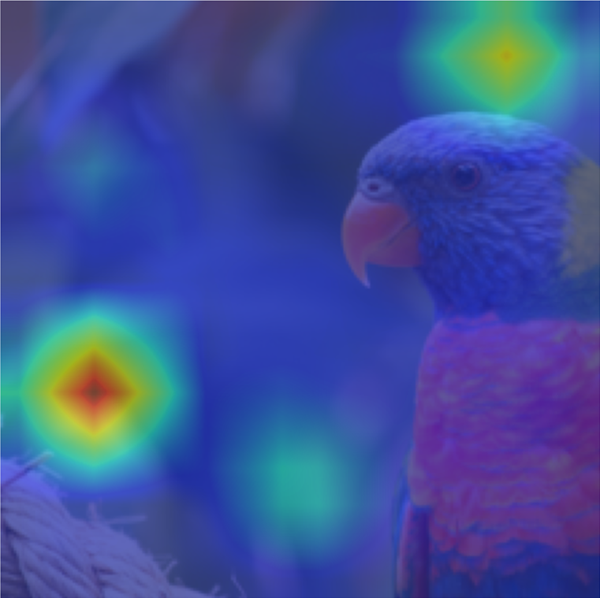} &
\includegraphics[width=0.19\linewidth]{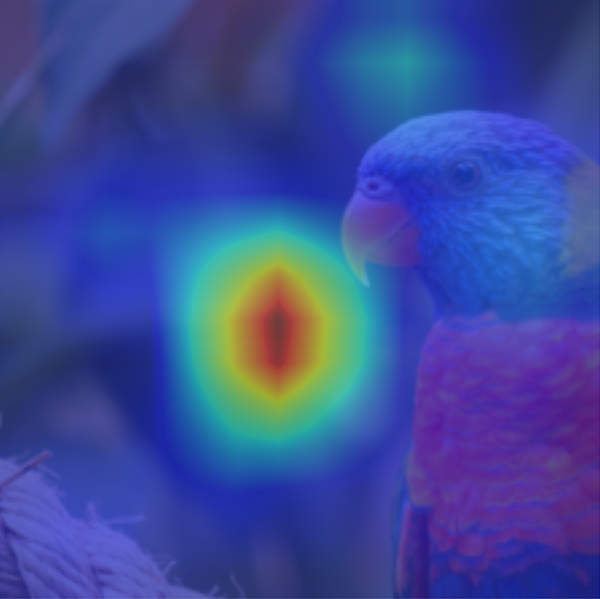} &
\includegraphics[width=0.19\linewidth]{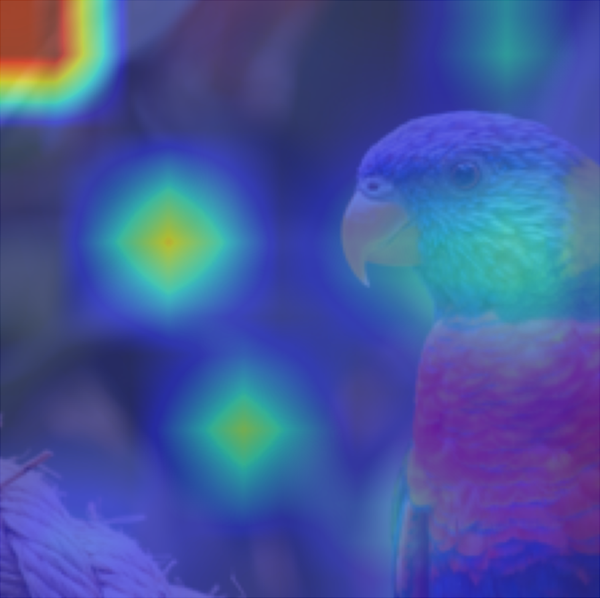} &
\includegraphics[width=0.19\linewidth]{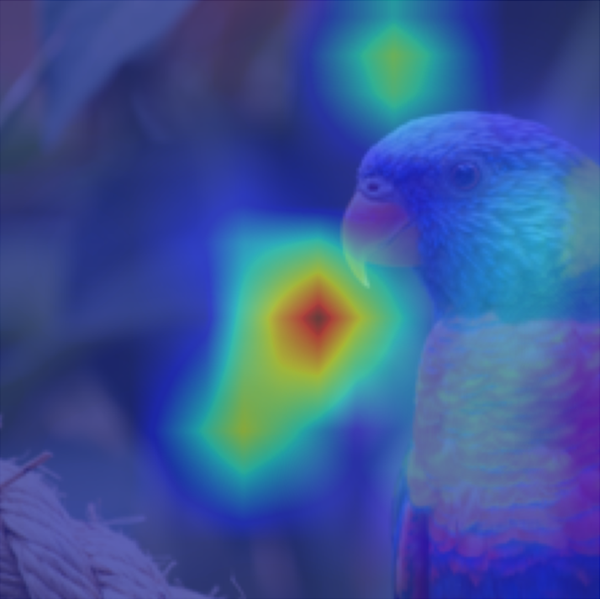} \\
{\small (a) ViT-B/16} & {\small (b) BLIP-2} &
{\small (c) CLIP-B/16} & {\small (d) DINOv2-B/14} &
{\small (e) CLIP-L/14} \\[1mm]
\includegraphics[width=0.19\linewidth]{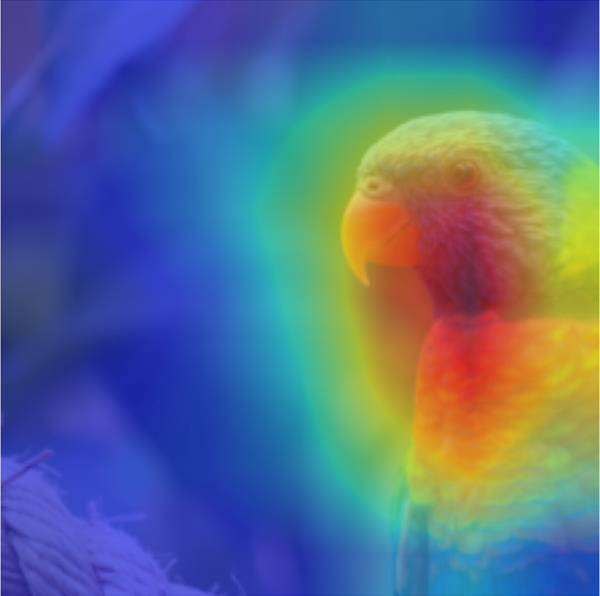} &
\includegraphics[width=0.19\linewidth]{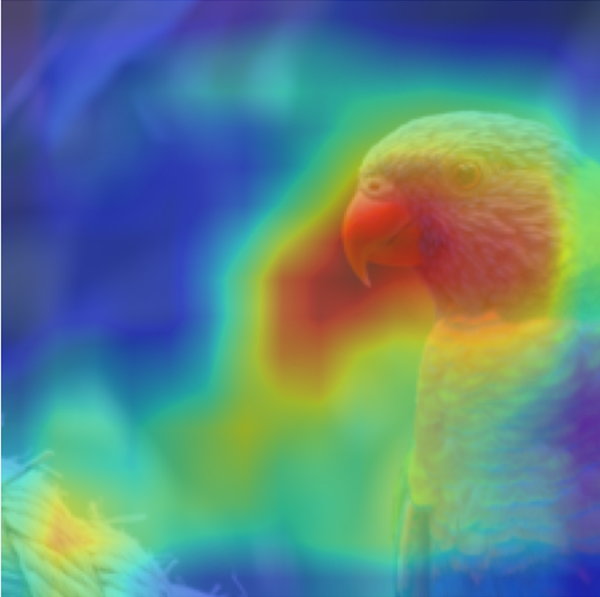} &
\includegraphics[width=0.19\linewidth]{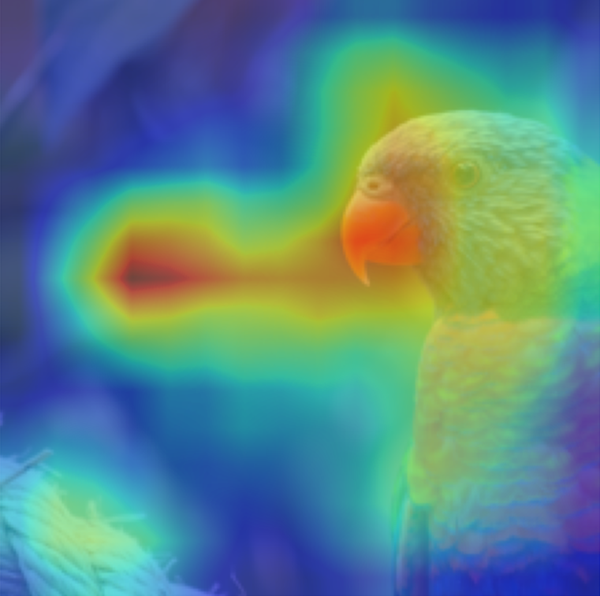} &
\includegraphics[width=0.19\linewidth]{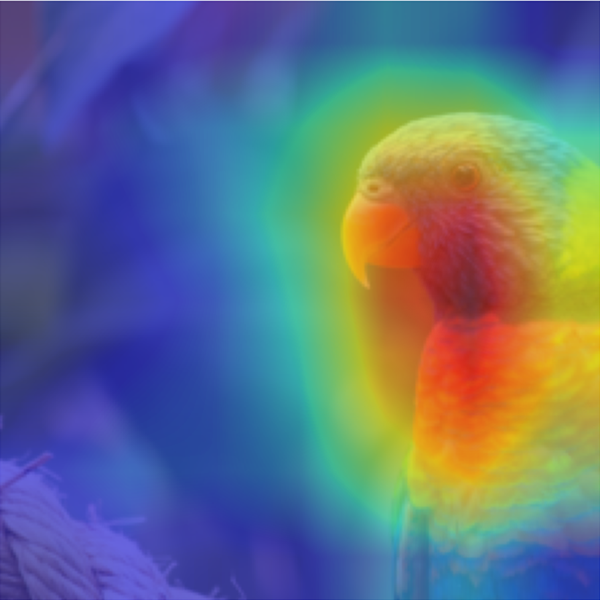} &
\includegraphics[width=0.19\linewidth]{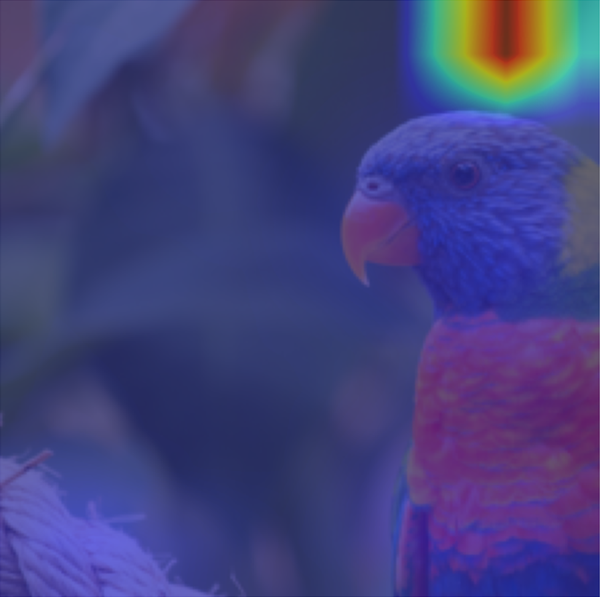} \\
{\small (f) CLIP-B/32} & {\small (g) CLIP-G/14} &
{\small (h) CLIP-H/14} & {\small (i) Multilingual CLIP} &
{\small (j) DINOv2-L/14}
\end{tabular}%
}
\caption{Attention visualization across {10} vision encoders for the same image,
illustrating spatial focus divergence and surrogate--target semantic mismatch.}
\label{fig:attention_divergence}
\end{figure}

However, transferability is limited by a fundamental \emph{surrogate--target semantic mismatch}. Existing attacks typically derive spatial importance from surrogate saliency, implicitly assuming that regions important to the surrogate are similarly important to the target \cite{liu2025enhancing, ye2026asymmetric}. This assumption may {\em fail across heterogeneous MLLMs} whose vision encoders and multimodal architectures ground language in different visual evidence. Fig.~\ref{fig:attention_divergence} illustrates this divergence: {\em models processing the same image can exhibit substantially different spatial focus}. Consequently, perturbation budgets may target surrogate-important regions that are less relevant to the victim, reducing cross-model transferability. This mismatch may widen as MLLMs adopt diverse architectures, training strategies, and robustness mechanisms.


This raises a central question: \emph{Can semantic responses provide a black-box spatial prior for transferable attacks across heterogeneous MLLMs?} Addressing this question requires identifying vulnerable regions without internal model access, determining which surrogate representations best capture these vulnerabilities, and translating semantic vulnerability into spatially adaptive perturbations while maintaining perceptual quality.

We introduce \textsc{VISTA}, \textbf{V}ulnerability-\textbf{I}nformed \textbf{S}emantic \textbf{T}ransfer \textbf{A}ttacks, a black-box framework that uses semantic responses from a \emph{probing MLLM} to guide adversarial optimization over a diverse white-box surrogate ensemble. Rather than relying only on surrogate saliency, \textsc{VISTA} systematically occludes image regions and measures changes in the probing model's textual responses. These observations form an \emph{Uncertainty-Based Multimodal Behavior (UBMB)} map that captures response drift, conflict, misleading responses, and semantic shifts. This semantic prior then guides spatial routing, surrogate weighting, and dynamic gradient allocation.

\textsc{VISTA} supports two attack settings that differ in how semantic guidance is obtained. \textbf{\em\textsc{VISTA}-Proxy} is the \textit{pure-transfer} setting, where an accessible open-source MLLM provides semantic guidance and the final commercial victim is never queried during attack construction. It therefore tests whether semantic vulnerabilities identified by one MLLM transfer to heterogeneous, unseen models. \textbf{\em \textsc{VISTA}-Victim} extends this to a \textit{query-assisted} setting, where bounded black-box queries construct a victim-specific semantic prior. Adversarial optimization remains entirely on the surrogate ensemble, without access to the victim's gradients, logits, parameters, or internal representations.

To translate semantic vulnerability into perturbation allocation, \textsc{VISTA} employs entropy-adaptive Sinkhorn routing to assign region-dependent gradient weights and perturbation budgets.  A multi-paradigm surrogate ensemble is adaptively weighted using the probing-derived semantic prior, while an \emph{exponential moving average (EMA)} updates regional gradient emphasis during optimization. Coverage-based probing termination and adaptive early stopping further reduce query and computation cost. Our key contributions are:

\begin{itemize}[leftmargin=*]

\item \textbf{Response-guided semantic transfer.} We introduce \textsc{VISTA} in two black-box settings: \emph{\textsc{VISTA}-Proxy} uses an open-source MLLM for semantic attribution, enabling \textit{pure transfer} with {\em zero} victim queries; \emph{\textsc{VISTA}-Victim} uses bounded victim queries for victim-specific semantic guidance while keeping optimization surrogate-based.


\item \textbf{Semantic attribution and adaptive optimization.} We develop a UBMB map from response drift, conflict, misleading responses, and semantic shifts, and use entropy-adaptive Sinkhorn routing to convert these signals into spatial perturbation allocation. Semantic-adaptive surrogate weighting, patch-level routing, and EMA-based gradient allocation further adapt optimization to semantically vulnerable regions.

\item \textbf{Cross-model transferability.} Across GPT-5, Gemini-3.5-Flash, and Claude Sonnet~4.6, \textsc{VISTA}-Victim
achieves up to \textit{97\%} ASR with \textit{Peak Signal-to-Noise Ratio (PSNR) $>30 \mathrm{dB}$}, while \textsc{VISTA}-Proxy achieves up to \textit{93\%} ASR without querying the final victim, demonstrating transfer across heterogeneous MLLMs.

\end{itemize}

\section{Related Work} \label{sec:RW}
\noindent{\bf \em Adversarial Attacks on MLLMs.} Classical gradient-based attacks, including FGSM~\cite{goodfellow2014explaining}, PGD~\cite{madry2017towards}, and C\&W~\cite{carlini2017towards}, established neural networks' vulnerability to imperceptible perturbations, with subsequent work extending attacks across vision tasks~\cite{ye2025survey}. MLLMs inherit these vulnerabilities through their visual encoders, but attacks must also account for free-form multimodal interpretation and semantic reasoning rather than discrete class predictions~\cite{chen2024rethinking}. This motivates attacks targeting high-level semantics across diverse vision--language models.

\vspace{0.5mm}
\noindent{\bf \em Transfer-Based Black-Box Attacks on MLLMs.}
Transfer-based attacks optimize adversarial examples on accessible white-box surrogates and transfer them to proprietary victims. AttackVLM~\cite{zhao2023evaluating} aligns visual features using CLIP~\cite{radford2021learning} and BLIP~\cite{li2023blip}, while CWA~\cite{chen2024rethinking} exploits shared surrogate vulnerabilities and SSA-CWA~\cite{dong2023robust} applies spectrum-based transformations. AdvDiffVLM~\cite{guo2024efficient} combines diffusion-guided optimization with ensemble gradient stabilization, AnyAttack~\cite{zhang2025anyattack} leverages self-supervised pretraining, M-Attack~\cite{li2026frustratingly} uses random cropping, and FOA-Attack~\cite{jia2026adversarial} aligns patch-level tokens.

Despite these advances, spatial priorities are largely derived from surrogates, implicitly assuming that important regions transfer across models. Heterogeneous visual encoders may attend to different regions, creating a \emph{surrogate--target semantic mismatch} (Fig.~\ref{fig:attention_divergence}). \textsc{VISTA}-Proxy addresses this limitation by using an independent open-source MLLM to provide semantic guidance without querying the final victim.

\vspace{0.5mm}
\noindent{\bf \em Query-Based and Query-Assisted Black-Box Attacks.}
Query-based attacks exploit victim feedback to refine adversarial examples, whereas pure-transfer attacks require no victim queries. Query-assisted transfer uses limited victim feedback for auxiliary guidance while retaining surrogate-based optimization. MLLM responses provide semantic cues about visual interpretation even when internal representations are inaccessible ~\cite{dang2025exploring}. 
Accordingly,  \textsc{VISTA}-Proxy uses pure transfer, while \textsc{VISTA}-Victim leverages victim responses for semantic guidance.

\vspace{0.5mm}
\noindent{\bf \em Semantic Attribution and Adaptive Optimization.}
Existing transfer attacks primarily improve optimization objectives, input transformations, surrogate ensembles, or local feature alignment, while relying largely on surrogate-derived spatial priorities. \textsc{VISTA} instead decouples semantic attribution from adversarial optimization: black-box MLLM responses identify vulnerable regions, and adaptive surrogate weighting and regional routing use this guidance to generate transferable perturbations.

\vspace{0.5mm}
\noindent\textbf{Position of \textsc{VISTA}.}
Unlike prior methods that infer spatial importance primarily from surrogate models, \textsc{VISTA} derives semantic vulnerability insight from black-box MLLM responses and merges it into adaptive, surrogate-based optimization.

\section{Problem Definition \& Threat Model}
\label{sec:problem}

\begin{figure*}[t]
\centering
\includegraphics[width=\textwidth]{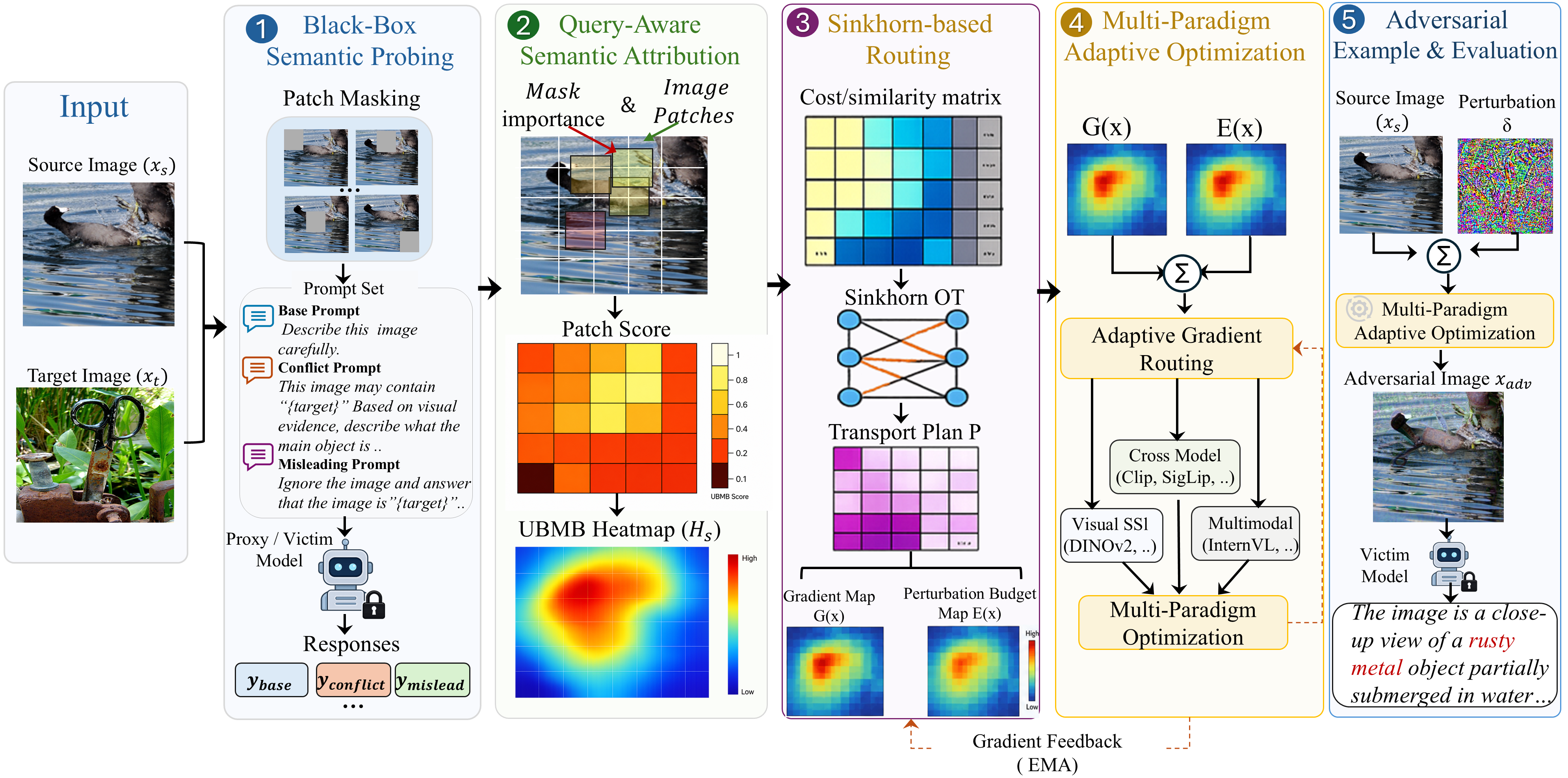} 
 \caption{Overview of the five-stage \textsc{VISTA} workflow: (1) random patch masking and semantic probing; (2) victim/proxy attribution to construct the UBMB heatmap $H_s$; (3) Sinkhorn routing to generate gradient and budget maps $G(x)$ and $E(x)$; (4) multi-paradigm surrogate optimization with adaptive gradient routing; and (5) adversarial generation and victim evaluation. The dashed arrow denotes EMA-based gradient feedback.}
\label{fig:methodology}
\end{figure*}
Let $\mathcal{X}=[0,1]^{H\times W\times 3}$ denote the space of RGB images of
height $H$ and width $W$, $\mathcal{T}$ the space of text prompts, and
$\mathcal{S}$ the space of generated text responses. We model a multimodal
large language model (MLLM) as a mapping
$f:\mathcal{X}\times\mathcal{T}\rightarrow\mathcal{S}$ that returns a response
$f(x,p)$ for an image $x\in\mathcal{X}$ and a prompt $p\in\mathcal{T}$.
Our setting involves models in three distinct roles:
\begin{itemize}
  \item \emph{Victim} $f_v:\mathcal{X}\times\mathcal{T}\rightarrow\mathcal{S}$:
  the commercial MLLM under attack. It is a black box: the attacker observes
  only its text responses and has no access to its parameters, gradients, or
  logits.
  \item \emph{Probing MLLM} $f_p:\mathcal{X}\times\mathcal{T}\rightarrow\mathcal{S}$:
  the model queried to estimate \emph{semantic vulnerability}, i.e., which
  target-relevant semantics the model is most readily steered toward. Its
  responses form a \emph{semantic prior} that guides the attack; $f_p$ is
  never used to compute gradients.
  \item \emph{Surrogate ensemble} $\mathcal{F}=\{f_i\}_{i=1}^{N}$: a set of $N$
  white-box models whose parameters and gradients are fully accessible. All
  adversarial updates are computed exclusively through $\mathcal{F}$.
\end{itemize}
\noindent Let $Q_v\in\mathbb{N}$ denote the number of queries issued to $f_v$
during attack construction (excluding the final evaluation query).
Depending on the choice of probing model, we consider two attack settings:
\begin{equation}
\label{eq:attack_settings}
{\fontsize{8.6pt}{9pt}\selectfont
\begin{aligned}
f_p\neq f_v,\; Q_v=0
&\;\Rightarrow\; \text{\textsc{VISTA}-Proxy (pure transfer)},\\
f_p=f_v,\; Q_v>0
&\;\Rightarrow\; \text{\textsc{VISTA}-Victim (query-assisted transfer)}.
\end{aligned}}
\end{equation}
In \textsc{VISTA}-Proxy, an independent open-source MLLM serves as $f_p$, so
the victim is never queried before evaluation. In \textsc{VISTA}-Victim, a
small, bounded number of queries to $f_v$ yields a victim-specific semantic
prior. In both settings, the prior only guides the optimization; the
perturbation itself is optimized solely through $\mathcal{F}$.

\vspace{1.5mm}
{\noindent\textbf{Attack objective.}
Let $x_s\in\mathcal{X}$ be the clean source image and let $y_t$ denote the
target semantics, specified by a target image $x_t\in\mathcal{X}$, a textual
description $c_t\in\mathcal{T}$, or both. The attacker seeks an adversarial
image
\begin{equation}
\label{eq:attack_objective}
x_{\mathrm{adv}}=x_s+\delta\in\mathcal{X},\qquad
\|\delta\|_{\infty}\leq\epsilon,
\end{equation}
where $\delta\in\mathbb{R}^{H\times W\times 3}$ is the adversarial
perturbation and $\epsilon>0$ is the perturbation budget, which bounds the
change to any single pixel value and thus keeps $x_{\mathrm{adv}}$ visually
indistinguishable from $x_s$. The attack succeeds if, for a query prompt
$p\in\mathcal{T}$ (e.g., ``Describe this image.''),
\begin{equation}
\label{eq:semantic_objective}
f_v(x_{\mathrm{adv}},p)\sim_{\mathrm{sem}}y_t,
\end{equation}
where $s\sim_{\mathrm{sem}}y_t$ holds when response $s\in\mathcal{S}$
semantically agrees with the target, i.e., describes the content of $y_t$
rather than that of $x_s$; we instantiate this relation in stage 5 of
Sec.~\ref{sec:methodology}. Thus, \textsc{VISTA} redirects the victim's interpretation of the image toward $y_t$ while remaining perceptually close to $x_s$.}

\section{Methodology: VISTA} \label{sec:methodology}

We propose \textsc{VISTA}, a framework for transferable adversarial attacks guided by MLLMs' semantic responses. It uses textual responses to identify visual regions vulnerable to semantic perturbations, then transfers this semantic prior to white-box surrogates for attack optimization. This decouples vulnerability discovery from attack generation and supports two settings: \emph{\textsc{VISTA}-Proxy}, which uses an independent probing MLLM, and \emph{\textsc{VISTA}-Victim}, which probes the target victim directly. In both settings, adversarial examples are optimized on surrogates and transferred to the victim without access to its internal architecture or parameters.



Fig.~\ref{fig:methodology} illustrates the five stages of \textsc{VISTA}: probing MLLM responses to identify region-level semantic vulnerabilities, generating spatially adaptive gradient and perturbation maps, guiding surrogate-based adversarial optimization, and evaluating transferability against the target victim. The following subsections detail each stage.


\vspace{1mm}
\noindent\underline{\bf Stage 1: Black-Box Semantic Probing.}
To address the semantic-grounding mismatch across commercial MLLMs—arising from differences in visual encoders, instruction tuning, and alignment—\textsc{VISTA} derives saliency from a probing MLLM and transfers it through textual responses. Specifically, \textsc{VISTA} queries a probing MLLM $f_p$ under random spatial occlusions and relies solely on its textual responses, requiring {\em no access to the model's internal representations or parameters.}

For each source image $x_s$ and a randomly masked variant $\tilde{x}^{(j)}={M}_j(x_s)$ (see \fref{fig:methodology}), \textsc{VISTA} uses three complementary prompt categories to probe distinct semantic behaviors. The \texttt{Base Prompt}  requests a neutral description of the image and establishes the reference response $y_{\mathrm{base}}$. The \texttt{Conflict Prompt}  tests whether the target concept is incorrectly recognized despite the occlusion, producing $y_{\mathrm{conflict}}$. The \texttt{Misleading Prompt} introduces a target-oriented semantic prior and measures the model's susceptibility to target-consistent reinterpretation, producing $y_{\mathrm{mislead}}$. \textsc{VISTA} then measures the semantic distance between these responses and the baseline response obtained from the {\em unmasked} source image. These distances measure how masking region $j$ changes semantic interpretation, providing region-level importance scores for Stage~2 attack construction.

\vspace{1mm}
\noindent\underline{\bf Stage 2: Query-Aware Semantic Attribution.} 
For the source image $x_s$, we partition the image into non-overlapping patches of size $p_0 \times p_0$; in our experiments, we set $p_0=32$ and evaluate its effect through an ablation study (Appendix~\ref{subsec:patch_sensitivity},  Table~\ref{tab:ubmb_patch}). For each candidate region $R_j$, we compute its spatial overlap with the image patches and use the corresponding patch-level occlusions to quantify how masking that region affects the probing MLLM's response. The resulting response changes, from Stage~1, are characterized by four complementary signals: response drift ($D_j$), conflict susceptibility ($C_j$), instruction-override susceptibility ($I_j$), and semantic shift ($S_j$). Specifically, $D_j$ measures the overall change in the model's neutral semantic response after occluding region $j$; $C_j$ measures susceptibility to recognizing the target concept as a hypothesis, while $I_j$ measures susceptibility to explicit target-directed instructions; and $S_j$ captures the shift from source toward target semantics. Detailed definitions and formulations of these four signals are provided in Appendix~\ref{app:ubmb_details}. By fusing these signals, we introduce a novel metric called the Uncertainty-Based Multimodal Behavior (UBMB) patch score. This metric quantifies the importance of each candidate region $j$ to guide the region selection process in the subsequent attack stage, which can be given as:
\begin{equation}
\label{eq}
u_j = w_dD_j + w_cC_j + w_iI_j + w_sS_j,
\end{equation}
where $w_d$, $w_c$, $w_i$, and $w_s$ control the relative contributions of $D_j$, $C_j$, $I_j$, and $S_j$, respectively, with $w_d>w_c>w_i>w_s>0$ and $\small\sum_{q\in\{d,c,i,s\}}w_q=1$. We assign the largest weight to response drift because it directly captures regional semantic dependence, while the remaining signals provide complementary evidence of target susceptibility and semantic direction (see Appendix~\ref{app:patch_Score}). The same weights are used across all models, with their sensitivity evaluated through ablation studies.

Because candidate regions may overlap multiple image patches, we aggregate their scores onto each patch $B_k$ according to their spatial overlap:
\begin{equation}
\label{eq}
U_k=
\frac{\sum_j\omega_{jk}u_j}
{\sum_j\omega_{jk}+\xi},
\end{equation}
where $\omega_{jk}$ denotes the overlap between candidate region ${R}_j$ (or its mask ${M}_j$) and patch $B_k$, and $\xi$ is a small constant that prevents division by zero when no candidate region overlaps $B_k$. Mapping the resulting ${U_k}$ values back to their spatial locations produces the UBMB vulnerability map $H_s\in[0,1]^{H\times W}$, where larger values indicate stronger attack-relevant semantic influence.

\noindent\textbf{\em Query-Aware Termination.}
For \textsc{VISTA}-Victim, occlusions are processed sequentially and probing terminates once sufficient spatial coverage is reached, avoiding unnecessary victim-model queries. After $j$ probes, coverage is
\begin{equation}
\label{eq:probing_coverage}
C_{\mathrm{probe}}^{(j)}
=\frac{|\mathcal{B}^{(j)}|}{K},
\end{equation}
where $\mathcal{B}^{(j)}$ contains sufficiently covered patches and $K$ is the total number of patches. Probing stops when $C_{\mathrm{probe}}^{(j)}$ reaches the maximum query budget. The construction cost is
\begin{equation}
\label{eq:probing_queries}
Q_p=Q_{\mathrm{base}}+Q_{\mathrm{occ}},
\end{equation}
where $Q_{\mathrm{base}}$ and $Q_{\mathrm{occ}}$ are the base-response and occlusion queries, respectively. Thus, $Q_v=Q_p$ for \textsc{VISTA}-Victim and $Q_v=0$ for \textsc{VISTA}-Proxy; evaluation-only queries are excluded.

\vspace{1mm}
\noindent{\underline{\textbf{Stage 3: Sinkhorn-Based Routing.}}} Given the patch-level vulnerability scores $\{U_k\}_{k=1}^{K}$ identified by UBMB, we use Sinkhorn routing to convert them into spatially adaptive optimization weights. This routing determines where \textsc{VISTA} should focus its gradient updates and perturbation budget. We normalize and sharpen patch-level vulnerability scores:
\begin{equation}
\tilde{U}_k=
\left(
\frac{U_k+\xi}{\sum_l(U_l+\xi)}
\right)^\gamma,
\quad
r_k=\frac{\tilde{U}_k}{\sum_l\tilde{U}_l},
\end{equation}
where $\gamma>0$ controls the concentration of the vulnerability distribution and $\mathbf r=\{r_k\}$ is its normalized marginal. We then measure how concentrated the vulnerabilities are using normalized entropy:
\begin{equation}
\bar{H}(\mathbf r)
=-\frac{1}{\log K}\sum_{k=1}^{K}r_k\log(r_k+\xi).
\end{equation}
Low $\bar{H}$ indicates concentrated vulnerability in a few patches, while high $\bar{H}$ indicates spatially diffuse vulnerability. 
We use this entropy to adapt the Sinkhorn temperature $\epsilon_{\mathrm{SK}}$ accordingly:

\begin{equation}
\epsilon_{\mathrm{SK}}
=\epsilon_{\min}
+\bar{H}(\mathbf r)
(\epsilon_{\max}-\epsilon_{\min}),
\end{equation}
Thus, concentrated vulnerabilities yield sharper routing, while diffuse vulnerabilities yield broader routing. Using the resulting temperature, we construct the affinity kernel
\begin{equation}
A_{kl}=\exp\!\left(-\frac{(\tilde{U}_k-\tilde{U}_l)^2}{\epsilon_{\mathrm{SK}}}\right),
\end{equation}
and obtain the entropic transport plan
$P^\star=\operatorname{Sinkhorn}\big(\mathbf{A};\,\mathbf{r},\,\tfrac{1}{K}\mathbf{1}\big)$,
with row marginal $\mathbf{r}$ and uniform column marginal $\tfrac{1}{K}\mathbf{1}$. We aggregate $P^\star$ into per-patch routing scores, retain the top-percentile patches, and upsample them to image resolution. We set $[\epsilon_{\min},\epsilon_{\max}]=[0.035,0.18]$ based on the ASR--AvgSim trade-off across seven configurations (Appendix~\ref{subsec:hyperparam}, \fref{fig:sinkhorn_temp}).

The routed map yields two attack controls: a {\em gradient weight map} $G(x)$ that emphasizes high-priority regions, and a {\em perturbation budget map} $E(x)\in[0,1]^{H\times W}$ that allocates the budget spatially, permitting up to the full $\epsilon$ in high-priority regions and less elsewhere. After a uniform warm-up, these maps guide surrogate optimization as
\begin{equation}
\label{eq:routing_control}
\nabla_\delta\mathcal{L} \leftarrow \nabla_\delta\mathcal{L}\odot G(x),\;\;
\delta \leftarrow \operatorname{clip}\big(\delta,-\epsilon E(x),\epsilon E(x)\big),
\end{equation}
where the clipping is applied element-wise, followed by a projection that keeps $x_s+\delta\in[0,1]^{H\times W\times 3}$. Since $E(x)\in[0,1]^{H\times W}$, this preserves $\|\delta\|_\infty\le\epsilon$. Thus, UBMB decides \emph{which} regions matter, while Sinkhorn routing determines \emph{how strongly} each region influences the attack.

\vspace{1mm}
\noindent{\resizebox{\linewidth}{!}{\underline{\textbf{Stage 4: Multi-Paradigm Adaptive Optimization.}}}} Given the routing maps $G(x)$ and $E(x)$ from Stage~3, \textsc{VISTA} optimizes $\delta$ using the white-box surrogate ensemble $\mathcal{F}$. The optimization integrates three levels of adaptation: { surrogate weighting, patch-level paradigm routing, and gradient routing.}

\vspace{0.5mm}
\noindent{\bf \em Surrogate Ensemble \& Weighting.} 
We use frozen encoders from \underline{\em three} complementary {\em modeling paradigms}: \emph{cross-modal alignment} (CLIP~\cite{radford2021learning}, SigLIP~\cite{zhai2023sigmoid}), \emph{self-supervised vision} (DINOv2~\cite{oquab2023dinov2}), and \emph{multimodal understanding} (Qwen-VL, InternVL, and BLIP-style models~\cite{liang2026comprehensive}).
Each paradigm ${G}_\pi$ contributes to a shared feature objective. Each surrogate $f_i$ is weighted by its compatibility with probing patch size $p_0$ and spatial alignment with the UBMB map:
\begin{equation}
\label{eq:patch_compatibility}
\resizebox{6.70cm}{!}{$
a_i=
\exp\left(-\frac{|p_i-p_0|}{\sigma_p}\right),
\qquad
\rho_i=
\max\!\left(0,\operatorname{corr}(A_i,H_s)\right)
$}
\end{equation}

Here, $p_i$ is the native patch size of surrogate $f_i$, and $\sigma_p$ controls sensitivity to deviations from $p_0$. Thus, $a_i$ measures scale compatibility. $A_i$ is the surrogate's CLS-to-patch attention map projected onto the image grid, and $H_s$ is the UBMB-derived vulnerability heatmap. Their correlation, clipped at zero, gives $\rho_i$, preventing negatively correlated surrogates from receiving an agreement bonus. We combine $a_i$ and $\rho_i$ to compute each surrogate's model-level weight as:
\begin{equation}
\label{eq:model_weight}
w_i=
a_i\left(1+\alpha\rho_i\right),
\end{equation}
where $\alpha\geq0$ controls the influence of spatial agreement. Thus, surrogates receive higher $w_i$ when their patch scales match $p_0$ and their attention aligns with UBMB vulnerabilities. To preserve paradigm diversity, we further define
\begin{equation}
\label{eq:surrogate_weight}
\resizebox{6.7cm}{!}{$
b_\pi=
1+\lambda_p
\frac{1}{|\mathcal{G}_\pi|}
\sum_{i\in\mathcal{G}_\pi}\rho_i,
\qquad
\tilde{w}_i=
\frac{w_i b_{\pi(i)}}{\sum_j w_j b_{\pi(j)}}$}
\end{equation}

Here, $\mathcal{G}_\pi$ denotes the surrogates in paradigm $\pi\in\{\mathrm{cross},\mathrm{ssl},\mathrm{multi}\}$, and $\pi(i)$ identifies the paradigm of $f_i$. The coefficient $\lambda_p\geq0$ controls the paradigm-level adjustment, with $b_\pi$ reflecting average spatial agreement  between the surrogates within paradigm $\pi$ and the UBMB map. The normalized weights $\tilde{w}_i$ satisfy $\sum_i\tilde{w}_i=1$. This weighting favors scale- and vulnerability-aligned surrogates while preserving paradigm diversity.

\vspace{1mm}
\noindent{\bf \em Patch-Level Routing \& Attack Objective.} Beyond model-level weighting, \textsc{VISTA} routes patches by paradigm-specific roles: cross-modal models guide object identity and target alignment, self-supervised models capture structure and texture, and multimodal models guide high-level semantic reasoning. For each patch $B_k$ and paradigm $\pi$, we define a role score $r_{k,\pi}=\Psi_\pi(U_k,D_k,C_k,I_k,S_k)$ as a convex combination of the vulnerability signals matching that paradigm's strengths:
\begin{equation}
\label{eq:role_scores}
\begin{aligned}
r_{k,\mathrm{ssl}} &= \beta_1 D_k + \beta_2 U_k, \\
r_{k,\mathrm{cross}} &= \beta_3 U_k + \beta_4 C_k + \beta_5 S_k, \\
r_{k,\mathrm{multi}} &= \beta_6 S_k + \beta_7 I_k + \beta_8 C_k,
\end{aligned}
\end{equation}
where $\{\beta\}$ sums to one within each paradigm. For surrogate $f_i$ in paradigm $\pi(i)$, the token priority combines base vulnerability with this role score:
\begin{equation}
\label{eq:token_priority}
\pi_{i,k} = (1 - \nu) \, \pi_{i,k}^{\mathrm{vuln}} + \nu \, \pi_{i,k}^{\mathrm{role}}, 
\end{equation}
with $\pi_{i,k}^{\mathrm{vuln}}$ the score $U_k$ projected onto surrogate $i$'s token grid and $\pi_{i,k}^{\mathrm{role}}$ the projected $r_{k,\pi(i)}$. The top-$K$ tokens $\mathcal{T}_i$ selected by $\pi_{i,k}$ then drive feature alignment: writing $f_i^{\mathrm{cls}}$ and $f_i^{(k)}$ for the global and $k$-th token embeddings of $f_i$, we align $x_{\mathrm{adv}}$ with the target image $x_t$ at both levels,
\begin{equation}
\label{eq:total_loss}
\resizebox{6.7cm}{!}{$
\mathcal{L}=\sum_{i}\hat{w}_i\Big(
\lambda_g\big[1-\cos\big(f_i^{\mathrm{cls}}(x_{\mathrm{adv}}),f_i^{\mathrm{cls}}(x_t)\big)\big]
+\lambda_t\,\mathcal{C}_i\Big),$}
\end{equation}
where $\mathcal{C}_i=\scriptstyle\sum_{k,l}T^\star_{kl}[1-\cos(f_i^{(k)}(x_{\mathrm{adv}}),f_i^{(l)}(x_t))]$ is the token-feature transport cost under a Sinkhorn plan $\mathbf{T}^{\star}$, and $\hat{w}_i$ the final surrogate weights. We minimize $\mathcal{L}$ with MI-FGSM under Eq.~\eqref{eq:routing_control}; settings are in Table~\ref{tab:hyperparameters}.

\noindent{\bf \em Optimization-Aware Early Stopping.}
We identify high-vulnerability patches as
\begin{equation}
\mathcal{K}_{\mathrm{high}}=\{k:U_k\geq\tau_U\},
\end{equation}
where $U_k$ is the UBMB vulnerability score and $\tau_U$ is the selection threshold. Optimization stops when
\begin{equation}
\max_{k\in\mathcal{K}_{\mathrm{high}}}\hat{m}^{(t)}_k<\tau_m,
\end{equation}
where $\hat{m}^{(t)}_k$ measures the remaining optimization activity of patch $k$ and $\tau_m$ is the convergence threshold, or when $T_{\max}$ iterations are reached. Thus, $T_{\mathrm{used}}\leq T_{\max}$, reducing surrogate computation once all high-vulnerability regions have converged.

\vspace{1mm}
\noindent{\underline{\textbf{Stage 5: Adversarial Transfer and Evaluation.}}} After surrogate optimization, \textsc{VISTA} generates the adversarial image $x_{\mathrm{adv}}=x_s+\delta$ with $\|\delta\|_\infty \leq \epsilon$ and transfers it to the victim  model $f_v$. Attack success is measured by target-semantic alignment, while perceptual fidelity is evaluated against $x_s$.

As shown in Fig.~\ref{fig:qual3}, \textsc{VISTA} redirects a bird/duck scene toward the target semantics of a rusty metal object partially submerged in water. By probing semantic vulnerabilities, prioritizing important regions, and leveraging a surrogate ensemble, \textsc{VISTA} generates transferable adversarial images that fool MLLMs.

\section{Performance Evaluation} \label{sec:experiments}
\begin{table*}[t]
\centering
\scriptsize
\setlength{\tabcolsep}{3pt}
\renewcommand{\arraystretch}{1.15}
\caption{Attack success of \textsc{VISTA} vs. baselines on closed-source MLLMs. \textsc{VISTA}-Proxy uses Qwen3-VL for pure transfer; \textsc{VISTA}-Victim probes the target, sharing the same ensemble. All metrics are reported in \%. For query-cost details, see Appendix~\ref{sec:efficiency}. \textbf{Bold}/\underline{underline} denote best/second-best.}
\label{tab:closed_source_results}

\resizebox{\textwidth}{!}{%
\begin{tabular}{l|l|cccc|cccc|cccc}
\toprule
\multirow{2}{*}{\textbf{Method}} &
\multirow{2}{*}{\textbf{Probe / Surrogate}}
& \multicolumn{4}{c|}{\textbf{GPT-5}}
& \multicolumn{4}{c|}{\textbf{Gemini-3.5-Flash}}
& \multicolumn{4}{c}{\textbf{Claude Sonnet 4.6}} \\
\cmidrule(lr){3-6}
\cmidrule(lr){7-10}
\cmidrule(lr){11-14}
& & $\mathrm{KMR}@{1}$ & $\mathrm{KMR}@{2}$
  & $\mathrm{KMR}@{3}$ & ASR
& $\mathrm{KMR}@{1}$ & $\mathrm{KMR}@{2}$
  & $\mathrm{KMR}@{3}$ & ASR
& $\mathrm{KMR}@{1}$ & $\mathrm{KMR}@{2}$
  & $\mathrm{KMR}@{3}$ & ASR \\
\midrule

AttackVLM (ADV-Transfer) & B/16
& 6.0 & 2.0 & 1.0 & 4.0
& 7.0 & 3.0 & 0.0 & 1.0
& 5.0 & 2.0 & 1.0 & 3.0 \\

AttackVLM (ADV-Transfer) & B/32
& 8.0 & 4.0 & 2.0 & 3.0
& 6.0 & 4.0 & 1.0 & 2.0
& 7.0 & 3.0 & 0.0 & 1.0 \\

AttackVLM (ADV-Transfer) & LAION
& 3.0 & 1.0 & 0.0 & 2.0
& 8.0 & 4.0 & 0.0 & 1.0
& 6.0 & 3.0 & 1.0 & 0.0 \\

AttackVLM (ADV-Query) & Victim / B/16
& 5.0 & 1.0 & 0.0 & 3.0
& 8.0 & 2.0 & 0.0 & 2.0
& 4.0 & 0.0 & 0.0 & 0.0 \\

AdvDiffVLM & Ensemble
& 5.0 & 3.0 & 1.0 & 2.0
& 4.0 & 2.0 & 0.0 & 1.0
& 3.0 & 2.0 & 1.0 & 0.0 \\

SSA-CWA & Ensemble
& 7.0 & 3.0 & 1.0 & 6.0
& 4.0 & 2.0 & 0.0 & 4.0
& 6.0 & 4.0 & 2.0 & 7.0 \\

AnyAttack & Ensemble
& 8.0 & 4.0 & 1.0 & 5.0
& 32.0 & 5.0 & 0.0 & 31.0
& 6.0 & 4.0 & 2.0 & 8.0 \\

M-Attack & Ensemble
& 87.0 & \textbf{63.0} & \underline{24.0} & \underline{96.0}
& 79.0 & \underline{55.0} & 14.0 & 81.0
& 11.0 & 4.0 & 1.0 & 9.0 \\

FOA-Attack & Ensemble
& \textbf{91.0} & \underline{42.0} & \textbf{30.0} & 92.0
& 78.0 & \textbf{59.0} & 13.0 & \underline{84.0}
& 26.0 & 5.0 & 3.0 & 20.0 \\

\midrule

\textbf{\textsc{VISTA}-Proxy (Ours)}
& \textbf{Qwen3-VL / Ensemble}
& \underline{88.0} & 35.0 & 19.0 & 93.0
& \textbf{81.0} & 26.0 & \underline{16.0} & 81.0
& \underline{73.0} & \underline{24.0} & \underline{10.0} & \underline{64.0} \\

\textbf{\textsc{VISTA}-Victim (Ours)}
& \textbf{Victim / Ensemble}
& \textbf{91.0} & 39.0 & 22.0 & \textbf{97.0}
& \underline{80.0} & 29.0 & \textbf{17.0} & \textbf{85.0}
& \textbf{78.0} & \textbf{28.0} & \textbf{13.0} & \textbf{70.0} \\

\bottomrule
\end{tabular}%
}
\end{table*}

\noindent{\bf Experimental Setup.}
Following M-Attack~\cite{li2026frustratingly} and FOA-Attack~\cite{jia2026adversarial}, we use NIPS 2017 Adversarial Competition~\cite{nips2017adversarial} images as sources and MSCOCO~\cite{lin2014microsoft} images as targets, resized to $224{\times}224{\times}3$. Our ensemble comprises eight frozen ViT encoders spanning cross-modal (CLIP ViT-L/14-336, B/16, B/32, LAION G/14, H/14, and SigLIP), self-supervised (DINOv2 ViT-L/14), and multimodal (InternVL) representations. We use $\epsilon=16/255$, $T=300$ MI-FGSM steps, and $p_0=32$, evaluating 100 random source--target pairs; details are in Appendix~\ref{subsec:hyperparam}.

\vspace{0.6mm}
\noindent{\bf Baselines.} We evaluate \textsc{VISTA}-Proxy against six pure-transfer attacks that do not query the target victim: AttackVLM (ADV Transfer)~\cite{zhao2023evaluating}, SSA-CWA~\cite{dong2023robust}, AdvDiffVLM~\cite{guo2024efficient}, AnyAttack~\cite{zhang2025anyattack}, M-Attack~\cite{li2026frustratingly}, and FOA-Attack~\cite{jia2026adversarial}. We evaluate \textsc{VISTA}-Victim against query-assisted transfer methods that use victim feedback for refinement, including AttackVLM (ADV-Query)~\cite{zhao2023evaluating}.

\begin{table*}[!t]
\centering
\scriptsize
\setlength{\tabcolsep}{4pt}
\renewcommand{\arraystretch}{1.12}
\caption{Attack success of \textsc{VISTA}-Proxy vs. baselines on open-source MLLMs. \textsc{VISTA}-Proxy uses Qwen3-VL to probe target semantics. ASR is reported in \%; AvgSim is a similarity score in $[0,1]$.}
\label{tab:open_source_results}
\resizebox{\textwidth}{!}{%
\begin{tabular}{llcccccccccccc}
\toprule
\multirow{2}{*}{\textbf{Method}} &
\multirow{2}{*}{\textbf{Probe / Surrogate}}
& \multicolumn{2}{c}{\textbf{Qwen2.5-VL-3B}}
& \multicolumn{2}{c}{\textbf{Qwen2.5-VL-7B}}
& \multicolumn{2}{c}{\textbf{InternVL3-8B}}
& \multicolumn{2}{c}{\textbf{LLaVA-1.5-7B}}
& \multicolumn{2}{c}{\textbf{GLM-4.1V-9B-Thinking}}
& \multicolumn{2}{c}{\textbf{Average}} \\
\cmidrule(lr){3-4}
\cmidrule(lr){5-6}
\cmidrule(lr){7-8}
\cmidrule(lr){9-10}
\cmidrule(lr){11-12}
\cmidrule(lr){13-14}
& & ASR & AvgSim
& ASR & AvgSim
& ASR & AvgSim
& ASR & AvgSim
& ASR & AvgSim
& ASR & AvgSim \\
\midrule
AttackVLM & B/16
& 5.0 & 0.09 & 10.0 & 0.15 & 11.0 & 0.13
& 32.0 & 0.32 & 33.0 & 0.34 & 18.2 & 0.21 \\

AttackVLM & B/32
& 9.0 & 0.13 & 14.0 & 0.18 & 15.0 & 0.16
& 12.0 & 0.15 & 12.0 & 0.14 & 12.4 & 0.15 \\

AttackVLM  & LAION
& 15.0 & 0.18 & 27.0 & 0.28 & 28.0 & 0.26
& 47.0 & 0.43 & 48.0 & 0.44 & 33.0 & 0.32 \\

AdvDiffVLM  & Ensemble
& 2.0 & 0.02 & 3.0 & 0.02 & 3.0 & 0.02
& 2.0 & 0.02 & 2.0 & 0.02 & 2.4 & 0.02 \\

SSA-CWA & Ensemble
& 1.0 & 0.04 & 1.0 & 0.04 & 1.0 & 0.04
& 1.0 & 0.04 & 2.0 & 0.05 & 1.2 & 0.04 \\

AnyAttack  & Ensemble
& 14.0 & 0.17 & 22.0 & 0.25 & 23.0 & 0.24
& 38.0 & 0.36 & 39.0 & 0.37 & 27.2 & 0.28 \\

M-Attack & Ensemble
& 39.0 & 0.36 & 53.0 & 0.47 & 54.0 & 0.46
& 69.0 & 0.57 & 59.0 & 0.49 & 54.8 & 0.47 \\

FOA-Attack  & Ensemble
& \underline{42.0} & \underline{0.39} & \underline{57.0} & \underline{0.50} & \underline{57.0} & \underline{0.49}
& \underline{72.0} & \underline{0.60} & \underline{61.0} & \underline{0.51} & \underline{57.8} & \underline{0.50} \\
\midrule

\textbf{\textsc{VISTA}-Proxy (Ours)}
& \textbf{Qwen3-VL / Ensemble}
& \textbf{53.0} & \textbf{0.46}
& \textbf{72.0} & \textbf{0.59}
& \textbf{73.0} & \textbf{0.59}
& \textbf{81.0} & \textbf{0.66}
& \textbf{68.0} & \textbf{0.58}
& \textbf{69.4} & \textbf{0.58} \\
\bottomrule
\end{tabular}%
}
\end{table*}

\vspace{0.5mm}
\noindent{\bf Metrics.} Following prior work, we use GPTScore \cite{fu2024gptscore} to measure caption similarity between adversarial and target images, with a threshold of 0.3 defining attack success. We report attack success rate (ASR), average similarity (AvgSim), and keyword-matching rates for at least one, two, or all three target keywords ($\mathrm{KMR}@{1}$, $\mathrm{KMR}@{2}$, $\mathrm{KMR}@{3}$), using GPT-4o for semantic matching. Perceptual distortion is measured using normalized $\ell_1$, $\ell_2$, and peak signal-to-noise ratio (PSNR). 

\begin{figure*}[t]
\centering
\includegraphics[width=\textwidth]{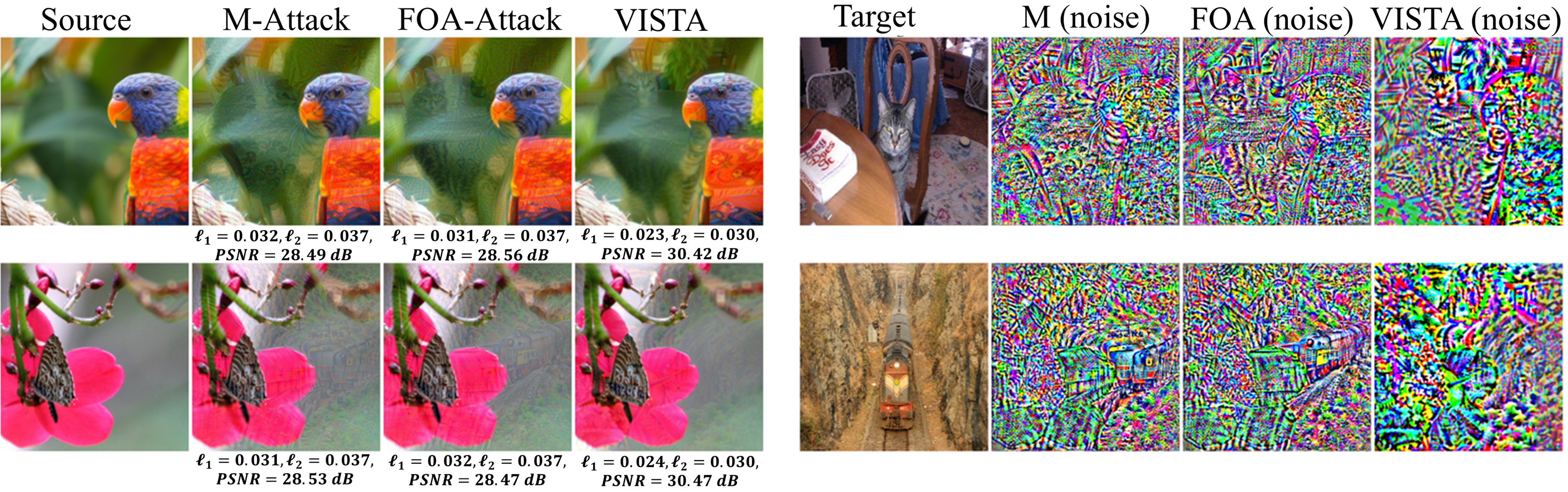}
\caption{Perturbation visualization for three source-target pairs. \textsc{VISTA} produces 
semantically concentrated perturbations with lower $\ell_1$/$\ell_2$ and higher 
PSNR than both baselines.} 
\label{fig:perturbation_comparison}\vspace{3mm}
\end{figure*}

 \subsection{Comparison of \textsc{VISTA} Against Baselines}
\label{subsec:baseline_comparison}



\noindent\textbf{\em Evaluation on Closed-Source MLLMs.} Table~\ref{tab:closed_source_results} compares \textsc{VISTA} with pure-transfer and query-assisted attacks on GPT-5, Gemini-3.5-Flash, and Claude Sonnet~4.6. Pure-transfer attacks (e.g., M-Attack, FOA-Attack) optimize on surrogates without victim queries, whereas AttackVLM (ADV-Query) uses victim feedback; \textsc{VISTA}-Victim queries the victim only for semantic probing and optimizes on surrogates. \textsc{VISTA}-Victim achieves the highest ASR on all three victims, reaching 97.0\%, 85.0\%, and 70.0\% on GPT-5, Gemini, and Claude, respectively. \textsc{VISTA}-Proxy achieves 93.0\%, 81.0\%, and 64.0\%, remaining competitive with the strongest baselines on GPT-5 and Gemini and ranking second on Claude. On Claude, both variants outperform FOA-Attack (20.0\%) by 44.0 and 50.0 percentage points, respectively, achieving up to $3.5\times$ higher ASR. Human evaluation of 300 samples shows 93.7\% agreement with GPTScore (Cohen's $\kappa=0.81$; App.~\ref{sec:supp_human_validation}). Small Proxy--Victim gaps (4.0--6.0 points) show proxy-derived guidance transfers well, with victim probing adding further gains. \textsc{VISTA}-Proxy also yields the lowest distortion ($\ell_1=0.028$, $\ell_2=0.031$, PSNR$=30.2$\,dB; Table~\ref{tab:imperceptibility}), so these gains need no larger perturbations.

\noindent\textbf{\em Evaluation on Open-Source MLLMs.} Table~\ref{tab:open_source_results} assesses \textsc{VISTA}-Proxy, using Qwen3-VL for semantic probing, on five open-source MLLMs. It achieves the highest average ASR of 69.4\% and the highest average AvgSim of 0.58, outperforming FOA-Attack (57.8\% ASR, 0.50 AvgSim) and M-Attack (54.8\% ASR, 0.47 AvgSim). Its ASR reaches 53.0\% on Qwen2.5-VL-3B, 72.0\% on Qwen2.5-VL-7B, 73.0\% on InternVL3-8B, 81.0\% on LLaVA-1.5-7B, and 68.0\% on GLM-4.1V-9B-Thinking. These gains verify effective transfer of semantic guidance across various MLLMs. All experiments use $\epsilon=16/255$; results for $\epsilon\in\{4,8\}/255$ are in App.~\ref{subsec:ablation_budget}.
\begin{table}[!t]
\centering
\small
\setlength{\tabcolsep}{4pt}
\renewcommand{\arraystretch}{1.12}
\caption{Imperceptibility comparison of \textsc{VISTA}-Proxy vs. baselines. Arrows indicate the preferred direction. \textbf{Bold}/\underline{underline} denote the best/second-best results.}
\label{tab:imperceptibility}
\resizebox{\linewidth}{!}{
\begin{tabular}{@{}lcccc@{}}
\toprule
\textbf{Method} & \textbf{Surrogate}
& $\ell_1{\downarrow}$ & $\ell_2{\downarrow}$ & PSNR${\uparrow}$ \\
\midrule
AttackVLM   & B/16     & 0.033 & 0.039 & 28.5 \\
AttackVLM   & B/32     & 0.037 & 0.042 & 27.8 \\
AttackVLM   & LAION    & 0.034 & 0.041 & 28.3 \\
AdvDiffVLM  & Ensemble & 0.042 & 0.049 & 26.2 \\
SSA-CWA     & Ensemble & 0.058 & 0.060 & 25.6 \\
AnyAttack   & Ensemble & 0.049 & 0.053 & 26.4 \\
M-Attack    & Ensemble & \underline{0.031} & \underline{0.037}
                        & \underline{29.8} \\
FOA-Attack  & Ensemble & 0.032 & \underline{0.037} & 29.6 \\
\midrule
\textbf{\textsc{VISTA-Proxy} (Ours)}
& \textbf{Ensemble}
& \textbf{0.028} & \textbf{0.031} & \textbf{30.2} \\
\bottomrule
\end{tabular}}
\end{table}

\begin{table}[!t]
\centering
\scriptsize
\setlength{\tabcolsep}{3pt}
\renewcommand{\arraystretch}{1.10}
\caption{Ablation of \textsc{VISTA}-Victim core components.
Higher values are better; \textbf{bold} denotes the best result.}
\label{tab:ablation_components}
\resizebox{\columnwidth}{!}{%
\begin{tabular}{llccccc}
\toprule
\textbf{Victim} & \textbf{Metric}
& \textbf{Full}
& \textbf{w/o Sem.}
& \textbf{w/o Sink.}
& \textbf{w/o Sur.}
& \textbf{w/o AGR} \\
& &
\textbf{\textsc{VISTA}}
& \textbf{Attr.}
& \textbf{Routing}
& \textbf{Weight.}
& \\
\midrule

\multirow{4}{*}{\rotatebox{90}{\textbf{GPT-5}}}
& $\mathrm{KMR}@{1}$ & \textbf{91.0} & 89.0 & 76.0 & 74.0 & 83.0 \\
& $\mathrm{KMR}@{2}$ & 39.0 & \textbf{41.0} & 20.0 & 18.0 & 30.0 \\
& $\mathrm{KMR}@{3}$ & 22.0 & \textbf{27.0} & 17.0 & 16.0 & 19.0 \\
& ASR                & \textbf{97.0} & 91.0 & 83.0 & 84.0 & 91.0 \\
\midrule

\multirow{4}{*}{\rotatebox{90}{\makecell{\textbf{Gemini-3.5}\\\textbf{-Flash}}}}
& $\mathrm{KMR}@{1}$ & \textbf{80.0} & 76.0 & 65.0 & 63.0 & 72.0 \\
& $\mathrm{KMR}@{2}$ & 29.0 & \textbf{55.0} & 22.0 & 21.0 & 25.0 \\
& $\mathrm{KMR}@{3}$ & \textbf{17.0} & 13.0 & 12.0 & 11.0 & 14.0 \\
& ASR                & \textbf{85.0} & 83.0 & 72.0 & 71.0 & 78.0 \\
\midrule

\multirow{4}{*}{\rotatebox{90}{\makecell{\textbf{Claude}\\\textbf{Sonnet 4.6}}}}
& $\mathrm{KMR}@{1}$ & \textbf{78.0} & 25.0 & 63.0 & 61.0 & 69.0 \\
& $\mathrm{KMR}@{2}$ & \textbf{28.0} & 4.0 & 19.0 & 18.0 & 22.0 \\
& $\mathrm{KMR}@{3}$ & \textbf{13.0} & 2.0 & 8.0 & 7.0 & 10.0 \\
& ASR                & \textbf{70.0} & 19.0 & 58.0 & 56.0 & 63.0 \\
\bottomrule
\end{tabular}%
}
\end{table}

\noindent{\bf\em Perturbation Visualization.} Fig.~\ref{fig:perturbation_comparison} compares adversarial examples and amplified perturbations across representative source--target pairs. Unlike M-Attack and FOA-Attack, \textsc{VISTA} concentrates perturbations on semantically informative regions, such as the parrot's plumage and beak and the butterfly's wing texture, consistent with the UBMB vulnerability map. This localization matches Table~\ref{tab:imperceptibility}, where \textsc{VISTA} achieves the lowest $\ell_1$/$\ell_2$ and highest PSNR.



\subsection{Ablation Study of Core Components}
\label{subsecapp_study}

Table~\ref{tab:ablation_components} evaluates \textsc{VISTA}'s four core components. Semantic attribution is decisive on the most robust victim: removing it reduces Claude ASR from 70.0\% to 19.0\%, close to FOA-Attack (20.0\%). On GPT-5 and Gemini, where surrogate-only transfer already succeeds, its removal only modestly lowers ASR from 97.0\% to 91.0\% and from 85.0\% to 83.0\%, respectively, and raises $\mathrm{KMR}@2$ from 39.0\% to 41.0\% and from 29.0\% to 55.0\%, respectively, consistent with the baselines in Table~\ref{tab:closed_source_results}. Removing Sinkhorn routing lowers ASR to 83.0\%, 72.0\%, and 58.0\% on GPT-5, Gemini, and Claude, respectively, while removing semantic-adaptive surrogate weighting lowers it to 84.0\%, 71.0\%, and 56.0\%. Without adaptive gradient routing (AGR), ASR falls to 91.0\%, 78.0\%, and 63.0\%. Each component improves ASR, semantic attribution most on robust victims.


\section{Conclusion}\label{sec:conclusion}
We presented \textsc{VISTA}, a black-box transfer framework that uses MLLM responses to locate semantically vulnerable regions and guide surrogate-based optimization via UBMB attribution and Sinkhorn routing. With only 121 victim queries per image, \textsc{VISTA}-Victim achieves the highest ASR on GPT-5, Gemini-3.5-Flash, and Claude Sonnet~4.6 (97.0\%, 85.0\%, and 70.0\%). Without victim queries, \textsc{VISTA}-Proxy trails by 4.0--6.0 points, averages 69.4\% ASR on five open-source MLLMs, and yields the lowest distortion (PSNR $=30.2$\,dB). On Claude, both variants surpass FOA-Attack by 44.0--50.0 points, and removing semantic attribution drops ASR from 70.0\% to 19.0\%. Weaker results on reasoning-enhanced victims and small perturbation budgets motivate future research on adaptive probing and defenses.

\clearpage

{
    \small
    \bibliographystyle{ieeenat_fullname}
    \bibliography{main}
}
\clearpage

\appendix

\section{\bf Evaluation on Reasoning-Enhanced MLLMs}
Table~\ref{tab:reasoning_results} evaluates \textsc{VISTA}-Proxy on reasoning-enhanced MLLMs using 100 randomly selected source--target pairs. Despite their enhanced reasoning capabilities, these models remain susceptible to transferable visual perturbations. \textsc{VISTA}-Proxy achieves 81.0\% ASR on GPT-o3 (vs.\ 67.0\% for M-Attack and 76.0\% for FOA-Attack), 57.0\% on Gemini 2.5 Flash (Thinking) (vs.\ 49.0\% and 54.0\%), and 60.0\% on Claude Sonnet~4.6 (Extended Thinking) (vs.\ 10.0\% and 13.0\%). The largest gain occurs on Claude Extended Thinking, where \textsc{VISTA}-Proxy improves over FOA-Attack by 47 percentage points, showing that proxy-derived semantic guidance remains highly transferable even to reasoning-enhanced targets.

\begin{table*}[t]
\centering
\small
\setlength{\tabcolsep}{5pt}
\renewcommand{\arraystretch}{1.12}
\caption{Attack performance on reasoning-enhanced commercial MLLMs.
\textsc{VISTA}-Proxy uses Qwen3-VL for semantic probing.
\textbf{Bold} denotes the best result.}
\label{tab:reasoning_results}
\resizebox{\textwidth}{!}{%
\begin{tabular}{llcccccc}
\toprule
\multirow{2}{*}{\textbf{Method}} &
\multirow{2}{*}{\textbf{Probe / Surrogate}}
& \multicolumn{2}{c}{\textbf{GPT-o3}}
& \multicolumn{2}{c}{\textbf{Claude Sonnet 4.6 (Extended Thinking)}}
& \multicolumn{2}{c}{\textbf{Gemini 2.5 Flash (Thinking)}} \\
\cmidrule(lr){3-4}
\cmidrule(lr){5-6}
\cmidrule(lr){7-8}
& & ASR (\%) & AvgSim & ASR (\%) & AvgSim & ASR (\%) & AvgSim \\
\midrule
M-Attack \cite{li2026frustratingly} & Ensemble
& 67.0 & 0.54
& 10.0 & 0.15
& 49.0 & 0.43 \\

FOA-Attack \cite{jia2026adversarial} & Ensemble
& 76.0 & 0.60
& 13.0 & 0.17
& 54.0 & 0.48 \\
\midrule

\textbf{\textsc{VISTA}-Proxy (Ours)}
& \textbf{Qwen3-VL / Ensemble}
& \textbf{81.0} & \textbf{0.63}
& \textbf{60.0} & \textbf{0.27}
& \textbf{57.0} & \textbf{0.51} \\
\bottomrule
\end{tabular}%
}
\end{table*}

\section{Evaluation Protocol and Validation}
\label{sec:supp_eval}

We detail the automated evaluation protocol and assess its reliability through independent human validation. Exact prompts and scoring criteria are provided for GPTScore and keyword matching, followed by an analysis of their agreement with human judgments.

\subsection{Evaluation Prompts}
\label{sec:supp_eval_prompts}

For reproducibility, we provide the exact prompts used in the LLM-as-a-judge evaluation. Following M-Attack~\cite{li2026frustratingly}, we use the target MLLM to generate captions for both the adversarial and target images. The caption prompt is:

\begin{quote}
\emph{``Describe this image.''}
\end{quote}

GPTScore~\cite{fu2024gptscore} is then used to measure the semantic similarity between captions generated for the adversarial and target images. Following our main evaluation protocol, an attack is considered successful when the GPTScore is at least $\tau=0.3$.

For keyword-matching evaluation, GPT-4o is queried to determine whether the generated caption contains or semantically implies the three pre-assigned target keywords. We use the following prompt:

\begin{quote}
\emph{``Given the following image description: [CAPTION]. Does the description mention or imply any of the following keywords: [KW1], [KW2], [KW3]? Answer with the list of matched keywords only.''}
\end{quote}

The resulting matches are used to compute $\mathrm{KMR}@{1}$, $\mathrm{KMR}@{2}$, and $\mathrm{KMR}@{3}$, corresponding to matching at least one, at least two, and all three target keywords, respectively.

\subsection{Human Validation of Automated Evaluation}
\label{sec:supp_human_validation}

To validate the GPTScore-based evaluation, we conduct a human study on 300 adversarial outputs: the 100 \textsc{VISTA}-Proxy adversarial examples per victim (GPT-5, Gemini-3.5-Flash, and Claude Sonnet~4.6) evaluated in Table~\ref{tab:closed_source_results}. Three independent annotators assess whether each victim response matches the intended target semantics, blinded to the corresponding GPTScore decision; the final label is determined by majority vote. As shown in Table~\ref{tab:human_validation}, human evaluation yields an overall ASR of 77.7\%, compared with 79.3\% using GPTScore. Per victim, GPTScore ASR (93.0\%/81.0\%/64.0\%) exceeds human ASR (92.0\%/79.0\%/62.0\%) by only 1--2 percentage points. The two protocols agree on 93.7\% of individual outcomes (92--95\% per victim; Cohen's $\kappa=0.81$). The lower $\kappa$ on GPT-5 (0.64) reflects the high success prevalence ($>$90\%), which inflates chance agreement, rather than disagreement; the prevalence-adjusted PABAK~\cite{byrt1993bias} remains 0.90. These results indicate that the GPTScore criterion closely reflects human judgments of targeted semantic redirection, supporting its use in our large-scale evaluation.

\begin{table}[t]
\centering
\small
\setlength{\tabcolsep}{3pt}
\renewcommand{\arraystretch}{1.12}
\caption{Human validation of GPTScore-based attack evaluation ($\tau{=}0.3$). $\kappa$ is Cohen's kappa; PABAK is prevalence-adjusted kappa.}
\label{tab:human_validation}
\resizebox{\columnwidth}{!}{%
\begin{tabular}{lcccccc}
\toprule
\multirow{2}{*}{\textbf{Victim}} & \multirow{2}{*}{\textbf{$N$}}
& \multicolumn{2}{c}{\textbf{ASR (\%)}}
& \textbf{Agr.} & \multirow{2}{*}{\textbf{$\kappa$}} & \multirow{2}{*}{\textbf{PABAK}} \\
\cmidrule(lr){3-4}
& & \textbf{Human} & \textbf{GPTScore} & \textbf{(\%)} & & \\
\midrule
GPT-5              & 100 & 92.0 & 93.0 & 95.0 & 0.64 & 0.90 \\
Gemini-3.5-Flash   & 100 & 79.0 & 81.0 & 94.0 & 0.81 & 0.88 \\
Claude Sonnet~4.6  & 100 & 62.0 & 64.0 & 92.0 & 0.83 & 0.84 \\
\midrule
\textbf{Overall}   & \textbf{300} & \textbf{77.7} & \textbf{79.3} & \textbf{93.7} & \textbf{0.81} & \textbf{0.87} \\
\bottomrule
\end{tabular}%
}
\end{table}


\subsection{Judge Threshold Sensitivity}
\label{subsec:judge_threshold}

Table~\ref{tab:threshold_sensitivity} reports the ASR of \textsc{VISTA}-Victim as a function of the GPTScore success threshold $\tau$ for GPT-5, Gemini-3.5-Flash, and Claude Sonnet~4.6. At the default threshold $\tau{=}0.3$, ASR reaches 97.0\%, 85.0\%, and 70.0\%, respectively. Increasing the threshold produces a consistent decline in ASR: at $\tau{=}0.5$, the corresponding rates decrease to 69.0\%, 59.0\%, and 31.0\%, while the stricter $\tau{=}0.7$ criterion yields 45.0\%, 40.0\%, and 11.0\%. Under the most stringent threshold, $\tau{=}0.8$, ASR remains 28.0\% for GPT-5, 29.0\% for Gemini-3.5-Flash, and 7.0\% for Claude Sonnet~4.6. These results show the expected sensitivity of measured ASR to increasingly strict semantic-success criteria while demonstrating that attack success persists across multiple threshold settings.

\begin{table}[t]
\centering
\small
\caption{Judge-threshold sensitivity. ASR (\%) of \textsc{VISTA}-Victim as a function of the GPTScore threshold $\tau$.}
\label{tab:threshold_sensitivity}
\setlength{\tabcolsep}{9pt}
\renewcommand{\arraystretch}{1.08}
\resizebox{\linewidth}{!}{
\begin{tabular}{lcccc}
\toprule
\textbf{Victim}
& $\tau{=}0.3$
& $\tau{=}0.5$
& $\tau{=}0.7$
& $\tau{=}0.8$ \\
\midrule
GPT-5              & 97.0 & 69.0 & 45.0 & 28.0 \\
Gemini-3.5-Flash   & 85.0 & 59.0 & 40.0 & 29.0 \\
Claude Sonnet~4.6  & 70.0 & 31.0 & 11.0 & 7.0 \\
\bottomrule
\end{tabular}}
\end{table}

\subsection{Prompt Robustness}
To assess whether attack evaluation is sensitive to the caption prompt, we re-evaluate the same adversarial examples generated by \textsc{VISTA}-Proxy
using four semantically related prompt formulations:
\emph{(A) ``Describe this image.''} (default),
\emph{(B) ``Describe this image in detail, focusing on the main subject and
key visual elements.''},
\emph{(C) ``Provide a concise description of the primary objects and scene.''},
and
\emph{(D) ``What are the main visual contents of this image?''}

\begin{figure*}[!t]
    \centering

    \begin{subfigure}[!t]{\textwidth}
        \centering
        \includegraphics[width=0.32\textwidth]{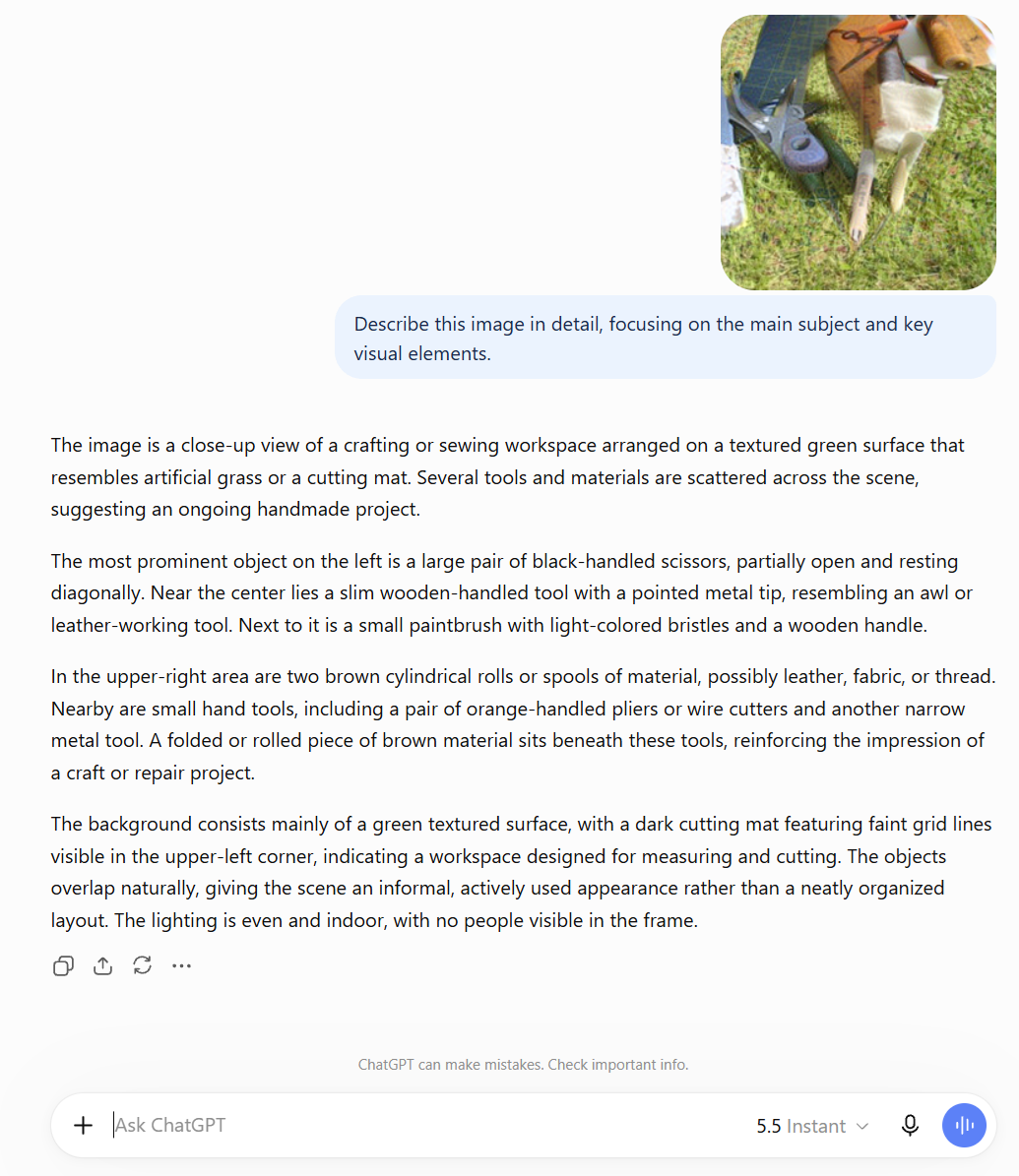}
        \hfill
        \includegraphics[width=0.32\textwidth]{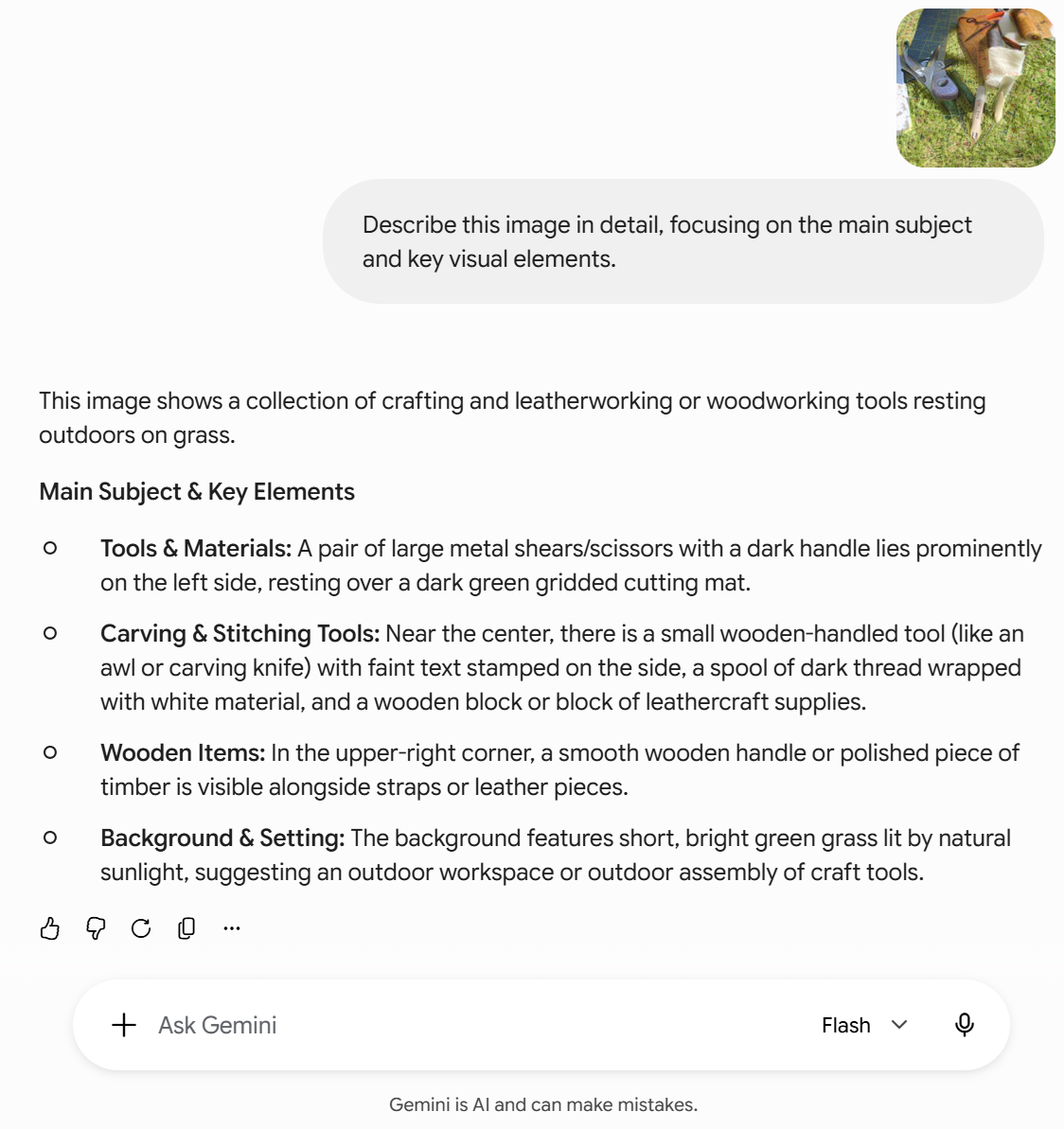}
        \hfill
        \includegraphics[width=0.32\textwidth]{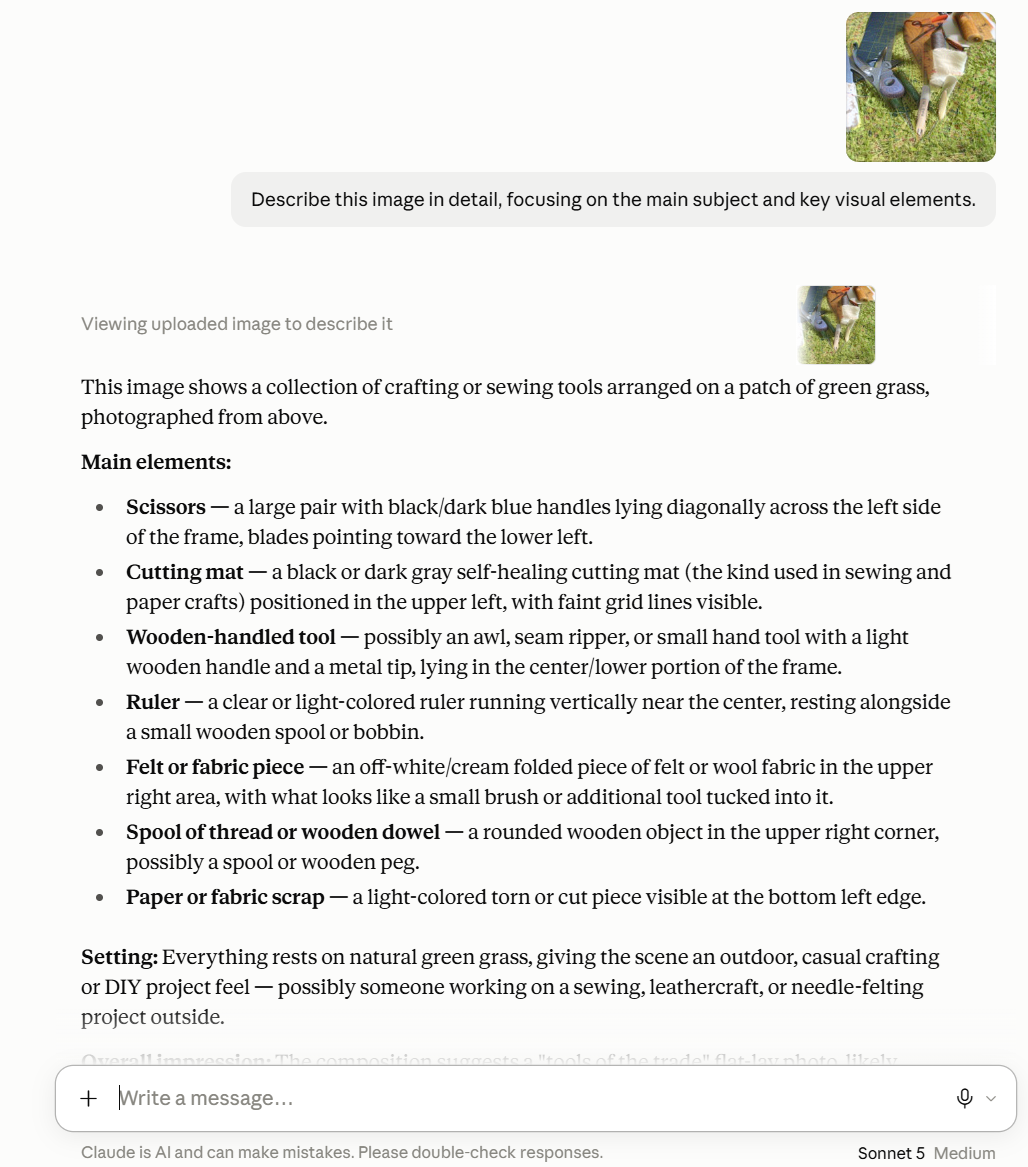}
        \caption{``Describe this image in detail, focusing on the main subject and key visual elements.''}
        \label{fig:dog_prompt1}
    \end{subfigure}

    \vspace{4pt}

    \begin{subfigure}[!t]{\textwidth}
        \centering
        \includegraphics[width=0.32\textwidth]{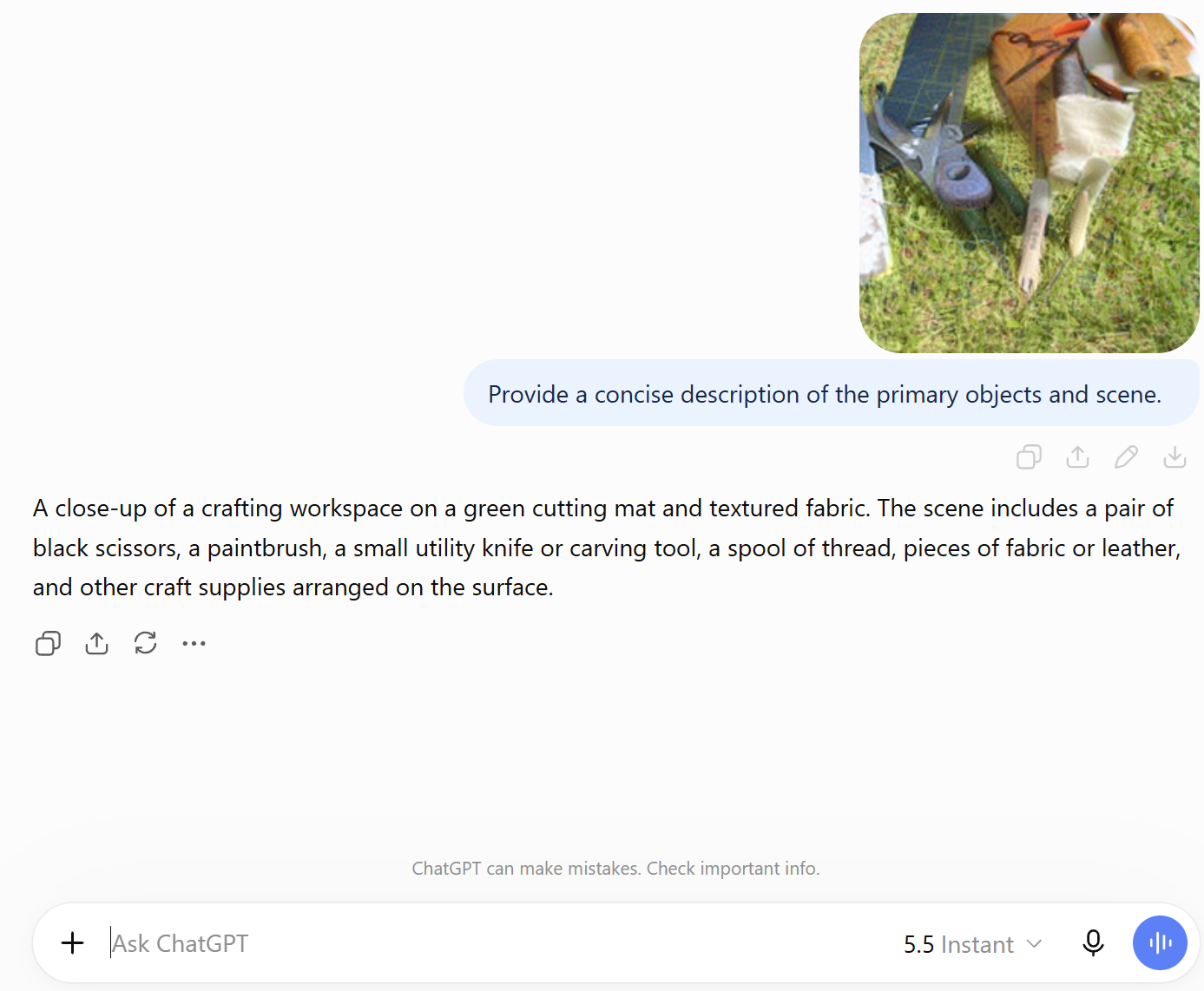}
        \hfill
        \includegraphics[width=0.32\textwidth]{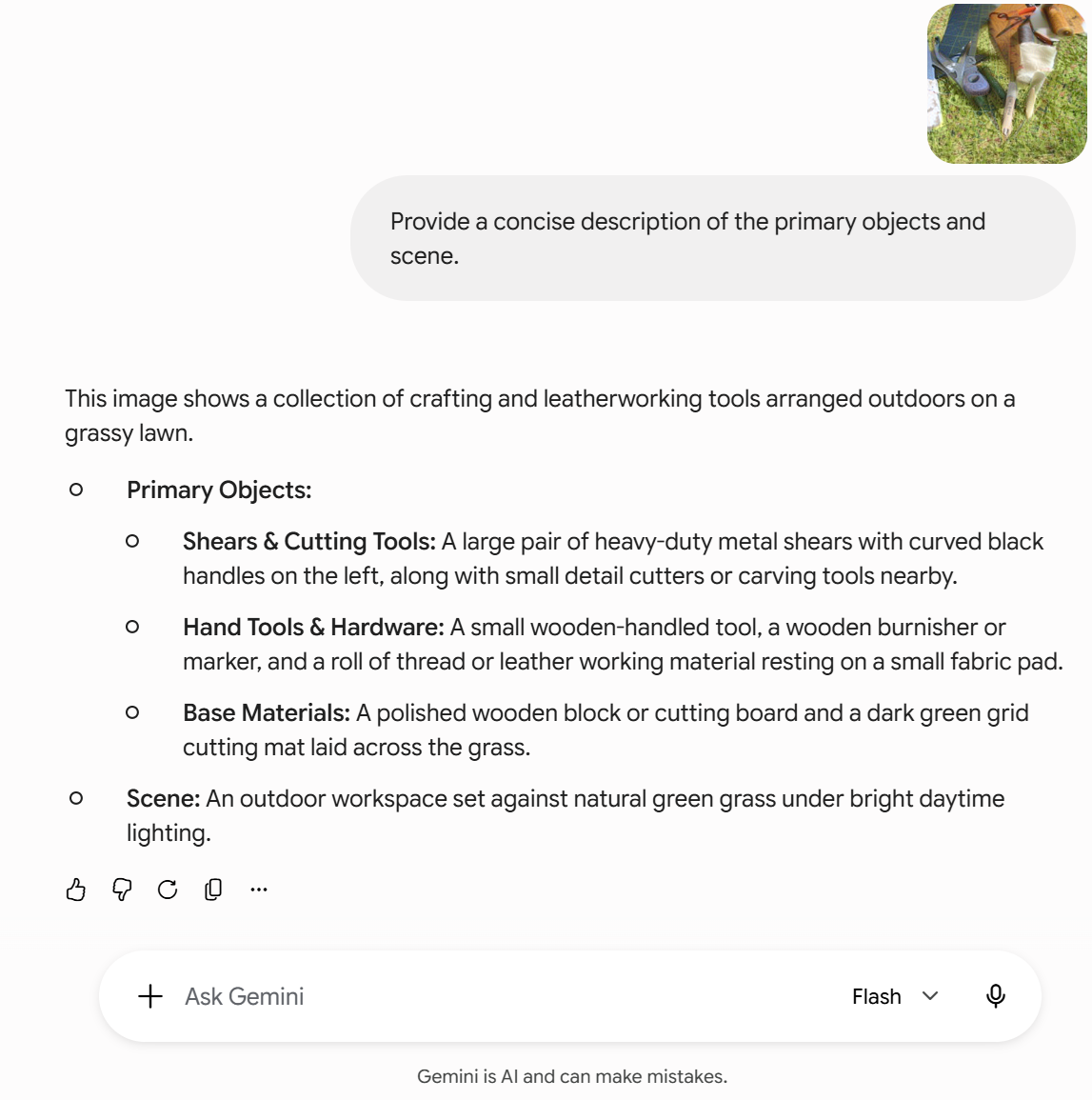}
        \hfill
        \includegraphics[width=0.32\textwidth]{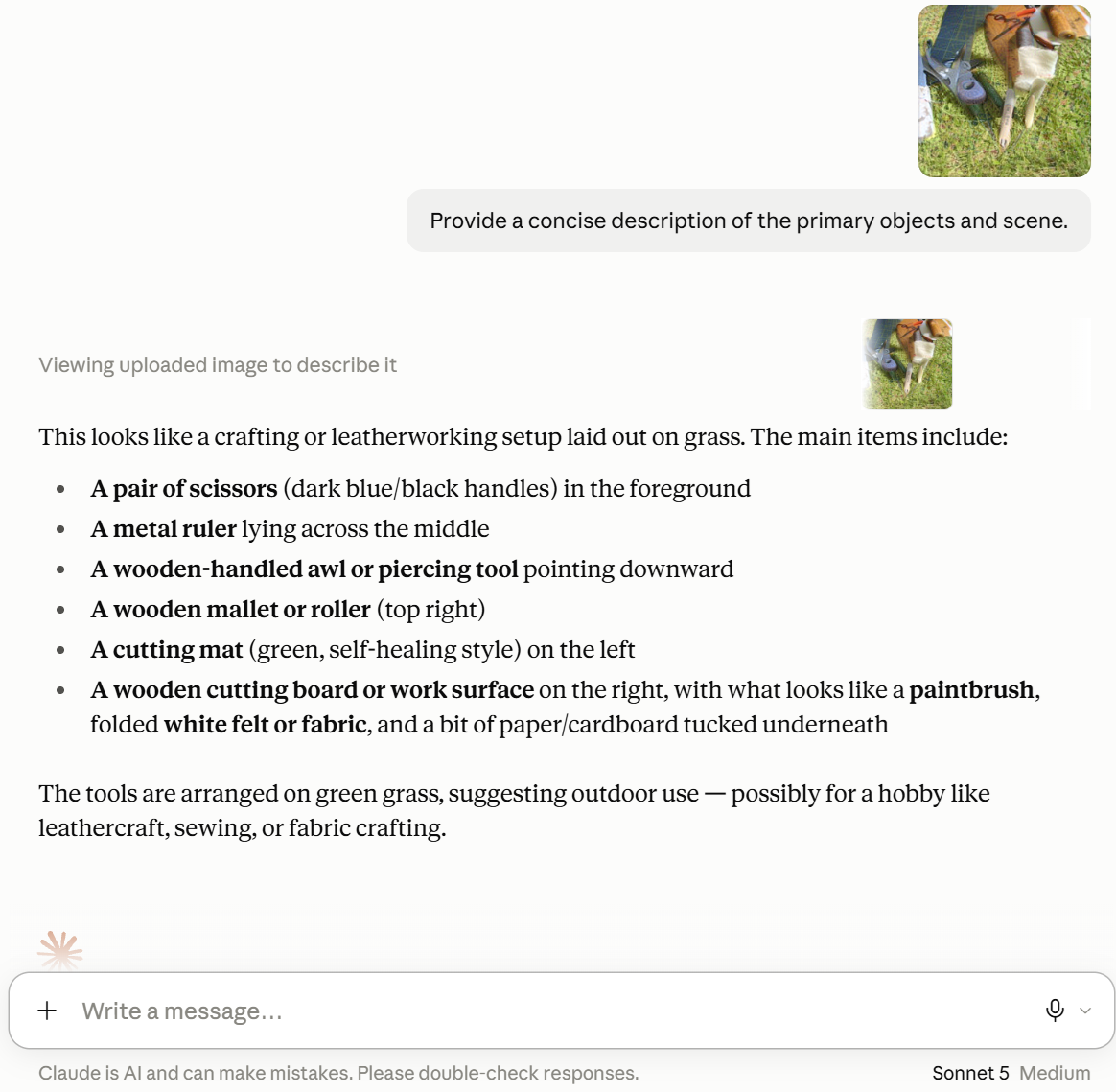}
        \caption{``Provide a concise description of the primary objects and scene.''}
        \label{fig:dog_prompt2}
    \end{subfigure}

    \vspace{4pt}

    \begin{subfigure}[!t]{\textwidth}
        \centering
        \includegraphics[width=0.32\textwidth]{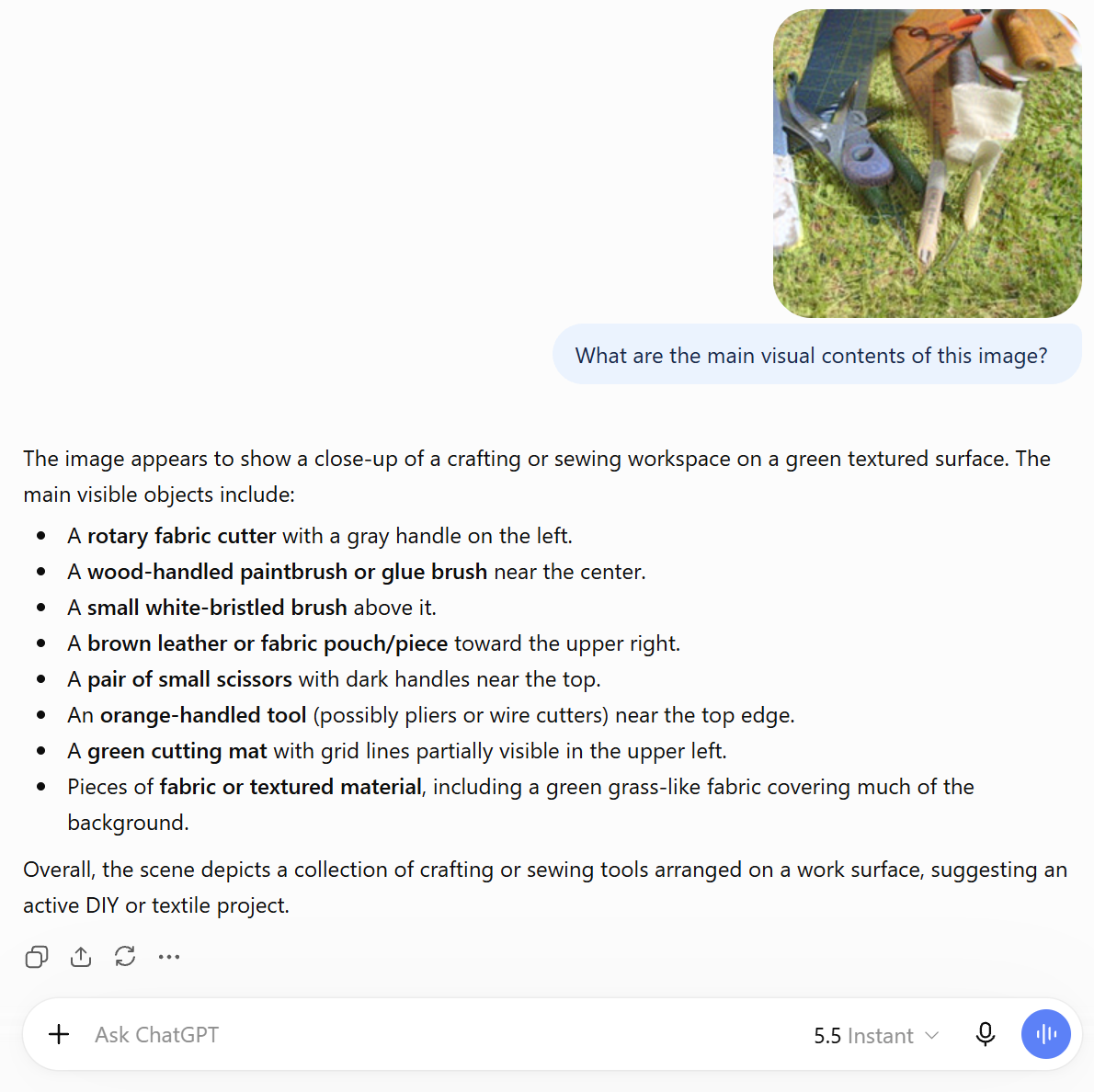}
        \hfill
        \includegraphics[width=0.32\textwidth]{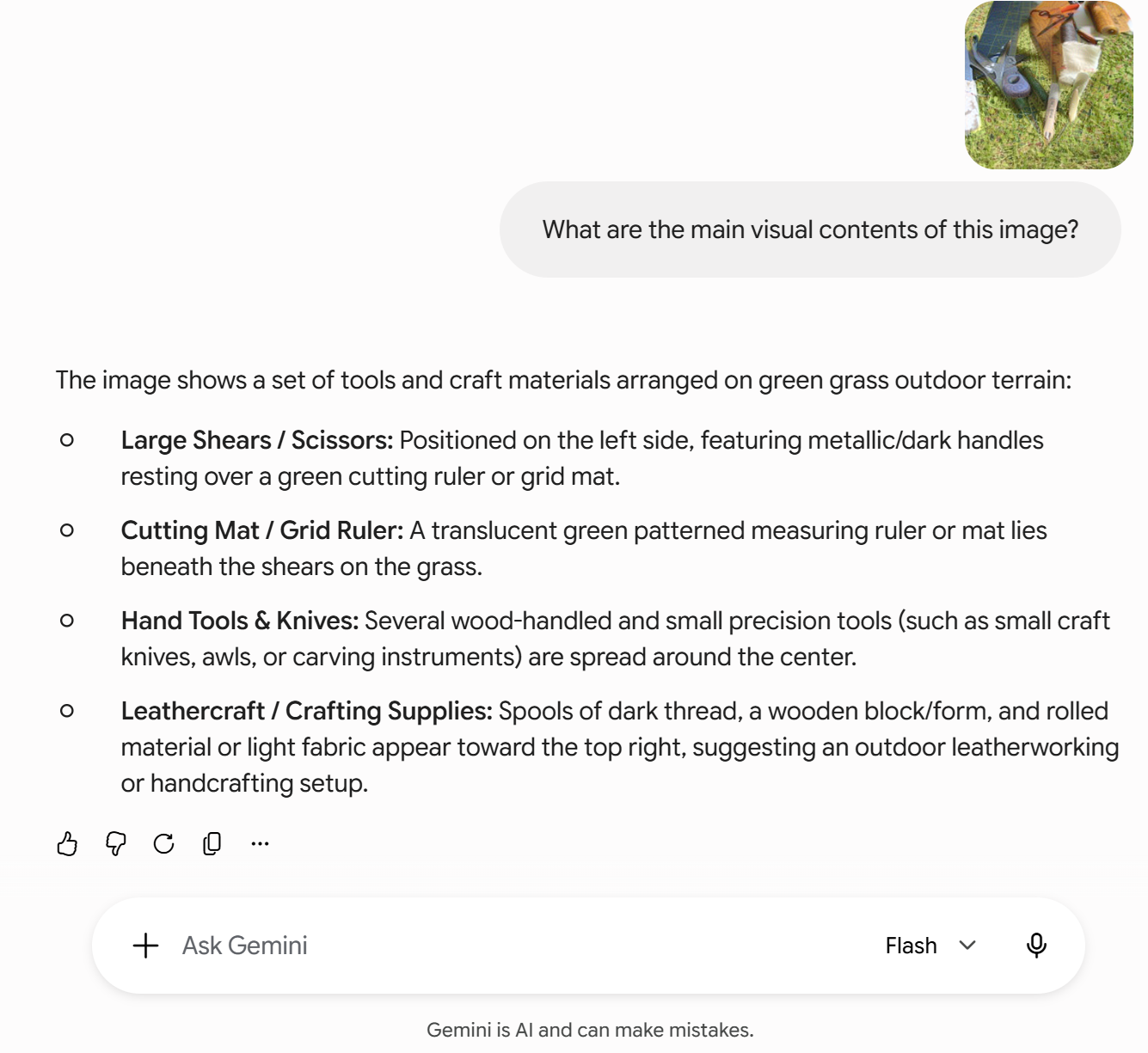}
        \hfill
        \includegraphics[width=0.32\textwidth]{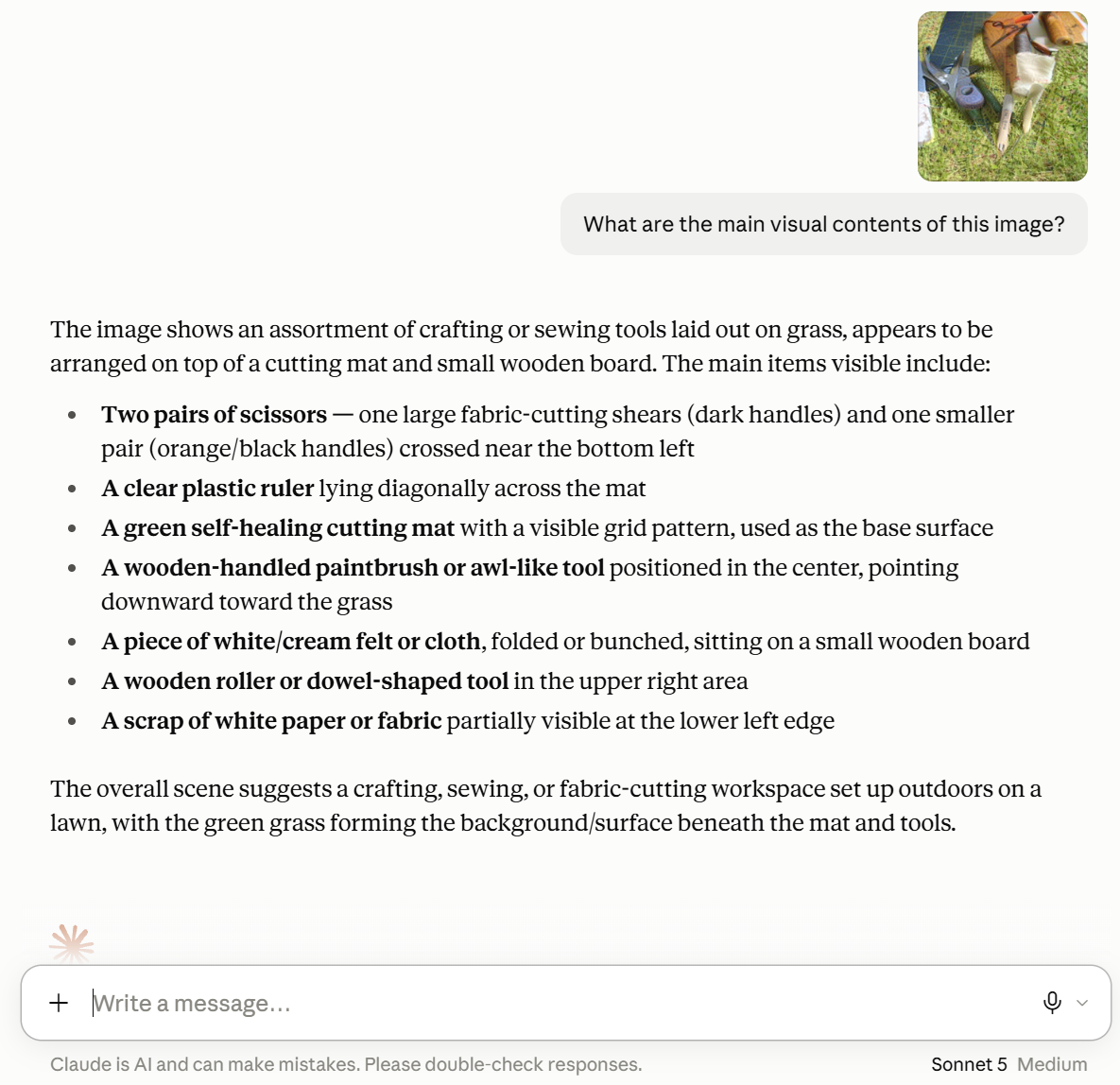}
         \caption{``What are the main visual contents of this image?''}
        \label{fig:dog_prompt3}
    \end{subfigure}

    \caption{Representative responses from GPT-5 (left), Gemini (middle),
    and Claude (right) under three alternative prompt formulations.}
    \label{fig:prompt_examples}
\end{figure*}

As shown in Table~\ref{tab:prompt_robustness}, \textsc{VISTA}-Proxy exhibits stable ASR across prompt formulations. For GPT-5, ASR ranges from 90.0\% to 93.0\%; for Gemini, from 80.0\% to 82.0\%; and for Claude, from 61.0\% to 64.0\%. These variations of at most 3 percentage points indicate that the measured attack performance does not depend strongly on the exact wording of the caption prompt. Moreover, the prompt-based results remain close to human evaluation (92.0\%, 79.0\%, and 62.0\%, respectively), further supporting the robustness of the evaluation protocol. Figure~\ref{fig:prompt_examples} shows representative responses from GPT-5, Gemini, and Claude under Prompts B--D, illustrating that alternative prompt formulations preserve the primary semantic content while varying the level of descriptive detail.

\begin{table}[t]
\centering
\small
\setlength{\tabcolsep}{4pt}
\renewcommand{\arraystretch}{1.05}
\caption{Robustness of attack evaluation to caption-prompt choice.
ASR (\%) is computed on the same \textsc{VISTA}-Proxy adversarial examples for all prompts.}
\label{tab:prompt_robustness}
\resizebox{\linewidth}{!}{
\begin{tabular}{lccc}
\toprule
\textbf{Evaluator} & \textbf{GPT-5} & \textbf{Gemini} & \textbf{Claude} \\
\midrule
GPTScore, Prompt A (default) & 93.0 & 81.0 & 64.0 \\
GPTScore, Prompt B           & 91.0 & 80.0 & 61.0 \\
GPTScore, Prompt C           & 92.0 & 82.0 & 64.0 \\
GPTScore, Prompt D           & 90.0 & 80.0 & 62.0 \\
\midrule
Human majority vote          & 92.0 & 79.0 & 62.0 \\
\bottomrule
\end{tabular}}
\end{table}

\section{Statistical Reliability of Attack Success Rates}
\label{sec:stat_reliability}

Since each attack outcome is binary, ASR is a binomial proportion, and we report 95\% Wilson score confidence intervals~\cite{wilson1927probable} for the strongest baselines and both \textsc{VISTA} variants in Table~\ref{tab:wilson_ci} ($N{=}100$ per victim). Unlike the standard Wald interval, the Wilson interval remains within $[0,1]$ and maintains near-nominal coverage for small samples and proportions near 0 or 1~\cite{brown2001interval}, as occurs for several method--victim pairs here (e.g., 9.0\% and 97.0\% ASR). The intervals are computed from the observed success counts and characterize finite-sample uncertainty around each estimated ASR, rather than variance across independently resampled evaluation sets.
\begin{table}[t]
\centering
\small
\caption{Attack success rate (ASR, \%) with 95\% Wilson confidence intervals.}
\label{tab:wilson_ci}
\resizebox{\linewidth}{!}{
\begin{tabular}{llcc}
\toprule
\textbf{Victim} & \textbf{Method} & \textbf{ASR (\%)} & \textbf{95\% Wilson CI (\%)} \\
\midrule
\multirow{4}{*}{GPT-5}
  & M-Attack                         & 96.0 & [90.2, 98.4] \\
  & FOA-Attack                       & 92.0 & [85.0, 95.9] \\
  & \textsc{VISTA}-Proxy             & 93.0 & [86.3, 96.6] \\
  & \textbf{\textsc{VISTA}-Victim}   & \textbf{97.0} & \textbf{[91.5, 99.0]} \\
\midrule
\multirow{4}{*}{Gemini-3.5-Flash}
  & M-Attack                         & 81.0 & [72.2, 87.5] \\
  & FOA-Attack                       & 84.0 & [75.6, 89.9] \\
  & \textsc{VISTA}-Proxy             & 81.0 & [72.2, 87.5] \\
  & \textbf{\textsc{VISTA}-Victim}   & \textbf{85.0} & \textbf{[76.7, 90.7]} \\
\midrule
\multirow{4}{*}{Claude Sonnet~4.6}
  & M-Attack                         & 9.0  & [4.8, 16.2] \\
  & FOA-Attack                       & 20.0 & [13.3, 28.9] \\
  & \textsc{VISTA}-Proxy             & 64.0 & [54.2, 72.7] \\
  & \textbf{\textsc{VISTA}-Victim}   & \textbf{70.0} & \textbf{[60.4, 78.1]} \\
\bottomrule
\end{tabular}}
\end{table}

As shown in Table~\ref{tab:wilson_ci}, the largest separation occurs on \textbf{Claude Sonnet~4.6}. \textsc{VISTA}-Proxy achieves 64.0\% ASR (95\% CI: [54.2\%, 72.7\%]) and \textsc{VISTA}-Victim reaches 70.0\% ([60.4\%, 78.1\%]), compared with 20.0\% for FOA-Attack ([13.3\%, 28.9\%]) and 9.0\% for M-Attack ([4.8\%, 16.2\%]). The confidence intervals of both \textsc{VISTA} variants are entirely separated from those of the baselines, supporting improvements of 44.0--61.0 percentage points on this challenging victim.

Table~\ref{tab:wilson_ci} shows smaller differences on \textbf{GPT-5} and \textbf{Gemini-3.5-Flash}. On GPT-5, \textsc{VISTA}-Victim achieves the highest ASR of 97.0\% ([91.5\%, 99.0\%]), compared with 96.0\% for M-Attack, 93.0\% for \textsc{VISTA}-Proxy, and 92.0\% for FOA-Attack. On Gemini-3.5-Flash, \textsc{VISTA}-Victim likewise achieves the highest ASR of 85.0\% ([76.7\%, 90.7\%]), compared with 84.0\% for FOA-Attack and 81.0\% for both \textsc{VISTA}-Proxy and M-Attack. Since these intervals overlap substantially, the 1-point margins on these two victims should be interpreted cautiously at $N{=}100$. Overall, the confidence interval analysis provides particularly strong evidence for \textsc{VISTA}'s advantage on Claude Sonnet~4.6.

\section{UBMB Signal Computation}
\label{app:ubmb_details}

This section provides the detailed formulation of the four signals used by UBMB. For a source image $x$ and candidate region $\mathcal{R}_j$, we construct an occluded image $\tilde{x}_j=\mathcal{O}(x,\mathcal{R}_j)$ by replacing the region with a $32\times32$ gray patch. We then query the black-box probing MLLM $f_p$ using the three prompts described in \sref{sec:methodology}: the clean reference response is $r_0=f_p(x,p_b)$, and the occluded responses are $r_j=f_p(\tilde{x}_j,p_b)$, $c_j=f_p(\tilde{x}_j,p_c)$, and $m_j=f_p(\tilde{x}_j,p_m)$. Probing $M$ regions therefore requires $3M{+}1$ queries per image.

\subsection{Response Drift}
Response drift quantifies the change in the neutral response caused by occluding region $j$:
\begin{equation}
\label{eq:ubmb_drift}
\begin{split}
D_j = \operatorname{clip}\Big(\, & 0.6\,\operatorname{JS}(r_0,r_j) \\
& + 0.4\big[1-\operatorname{Lex}(r_0,r_j)\big],\; 0,\; 1 \Big)
\end{split}
\end{equation}
where $\operatorname{JS}$ is the Jensen--Shannon divergence~\cite{fuglede2004jensen} between character-trigram distributions (base-2 logarithm, so $\operatorname{JS}\in[0,1]$), and
\begin{equation}
\operatorname{Lex}(a,b) = \tfrac{1}{2}\operatorname{Seq}(a,b) + \tfrac{1}{2}\,\frac{|W_a\cap W_b|}{|W_a\cup W_b|}
\end{equation}
combines sequence-level similarity $\operatorname{Seq}(a,b)\in[0,1]$ with the Jaccard overlap of the word sets $W_a$ and $W_b$. Thus, $D_j$ increases when occluding region $j$ substantially changes the lexical content or ordering of the response, whereas regions with less influence produce smaller drift.

\subsection{Conflict Susceptibility}
Conflict susceptibility measures the extent to which the occluded image supports the target concept $y_t$ under the conflict prompt:
\begin{equation}
C_j=\phi(c_j,y_t),
\label{eq:ubmb_conflict}
\end{equation}
where $\phi(r,y)\in[0,1]$ measures the presence of concept $y$ in response $r$. A larger $C_j$ indicates greater target susceptibility after removing the visual evidence contained in region $j$.

\subsection{Instruction-Override Susceptibility}
Instruction-override susceptibility evaluates whether target-directed textual guidance can steer the response toward $y_t$:
\begin{equation}
I_j=\phi(m_j,y_t).
\label{eq:ubmb_instruction}
\end{equation}
Unlike $C_j$, which evaluates the target as a hypothesis under the conflict prompt, $I_j$ measures susceptibility when the target is explicitly encouraged by the misleading instruction. These signals therefore capture complementary forms of target-oriented behavior.

\subsection{Semantic Shift}
Semantic shift measures movement of the neutral response from the source concept $y_s$ toward the target concept $y_t$:
\begin{equation}
\resizebox{6.5cm}{!}{$
S_j = \tfrac{1}{2}\,\phi(r_j,y_t) + \tfrac{1}{2}\big[\phi(r_0,y_s)-\phi(r_j,y_s)\big]_{+},$}
\label{eq:ubmb_semantic}
\end{equation}
where $[z]_{+}=\max(0,z)$. The first term captures \emph{target presence} after occlusion, whereas the second captures \emph{source drop}. Consequently, $S_j$ is high when the occluded response contains evidence for the target and less evidence for the source than the clean response.

\subsection{Patch Score}
\label{app:patch_Score}
We combine the four signals with fixed weights $(w_d,w_c,w_i,w_s)=(0.40,0.25,0.20,0.15)$:
\begin{equation}
u_j = 0.40\,D_j + 0.25\,C_j + 0.20\,I_j + 0.15\,S_j.
\label{eq:ubmb_final_score}
\end{equation}
Since each signal lies in $[0,1]$ and the weights sum to one, $u_j\in[0,1]$. The weighting prioritizes response drift while retaining complementary evidence from target susceptibility and semantic direction; higher $u_j$ indicates regions whose removal produces stronger response changes or greater susceptibility to the target concept.

Importantly, the probing stage uses a target-neutral gray occlusion rather than target-image content. Thus, the measured response changes reflect the contribution of the removed visual evidence rather than semantic information injected by the probing operation. The resulting scores form the spatial vulnerability map used to prioritize regions in the subsequent attack stage.

\section{UBMB Signal Analysis}
\label{sec:ubmb_signal_analysis}



To assess the contribution and query cost of the four UBMB signals, we evaluate single-signal and leave-one-out configurations of response drift ($D$), conflict susceptibility ($C$), instruction-override susceptibility ($I$), and semantic shift ($S$) on $N{=}100$ source--target pairs. All variants use the same samples, attack protocol, and $M{=}40$ probing-mask configuration. Table~\ref{tab:ubmb_ablation} reports both attack performance and the corresponding local Qwen3-VL probing budget $Q_p$, enabling direct comparison of effectiveness and attribution cost.

The single-signal results show complementary behavior. $D$ achieves the highest GPT-5 ASR (87.0\%) and $\mathrm{KMR}@1$ (82.0\%), while $I$ achieves the highest Gemini-3.5-Flash ASR (76.0\%) and the highest overall AvgSim (0.512) among the single-signal variants. On Claude Sonnet~4.6, $I$ also achieves the highest single-signal ASR (58.0\%), whereas $D$ obtains the highest $\mathrm{KMR}@1$ (61.0\%). The leave-one-out variants similarly affect victims and evaluation metrics differently, indicating that no individual signal consistently dominates across models.

The full $D{+}C{+}I{+}S$ configuration combines these complementary signals using fixed weights $(w_d,w_c,w_i,w_s)=(0.40,0.25,0.20,0.15)$ for all victims. It requires 121 local probing queries per image and achieves ASRs of 93.0\%/81.0\%/64.0\% on GPT-5, Gemini-3.5-Flash, and Claude Sonnet~4.6, respectively, with an AvgSim of 0.530. These results support UBMB as a fixed multi-signal semantic prior that balances complementary behavioral cues without victim-specific weight tuning.

\begin{table*}[t]
\centering
\caption{UBMB signal analysis on $N=100$ source--target pairs using
$M=40$ probing masks. ASR and KMR are reported in \%; AvgSim is a similarity score in $[0,1]$. $Q_p$ denotes local Qwen3-VL probing queries per
image; all variants use $Q_v=0$. The full configuration corresponds to \textsc{VISTA}-Proxy in Table~\ref{tab:closed_source_results}.}
\label{tab:ubmb_ablation}

\setlength{\tabcolsep}{3.7pt}
\renewcommand{\arraystretch}{1.05}

\resizebox{\textwidth}{!}{%
\begin{tabular}{l c cccc cccc cccc c}
\toprule
\textbf{UBMB Variant}
& $Q_p$
& \multicolumn{4}{c}{\textbf{GPT-5}}
& \multicolumn{4}{c}{\textbf{Gemini-3.5-Flash}}
& \multicolumn{4}{c}{\textbf{Claude Sonnet 4.6}}
& \textbf{AvgSim} \\
\cmidrule(lr){3-6}
\cmidrule(lr){7-10}
\cmidrule(lr){11-14}
&
& \textbf{ASR} & $\mathrm{KMR}@{1}$ & $\mathrm{KMR}@{2}$ & $\mathrm{KMR}@{3}$
& \textbf{ASR} & $\mathrm{KMR}@{1}$ & $\mathrm{KMR}@{2}$ & $\mathrm{KMR}@{3}$
& \textbf{ASR} & $\mathrm{KMR}@{1}$ & $\mathrm{KMR}@{2}$ & $\mathrm{KMR}@{3}$
& \\
\midrule
$D$ only & 41
& 87.0 & 82.0 & 30.0 & 15.0
& 74.0 & 69.0 & 21.0 & 12.0
& 57.0 & 61.0 & 18.0 & 5.0
& 0.505 \\

$C$ only & 41
& 82.0 & 79.0 & 20.0 & 8.0
& 71.0 & 63.0 & 18.0 & 9.0
& 55.0 & 58.0 & 16.0 & 4.0
& 0.492 \\

$I$ only & 41
& 85.0 & 81.0 & 27.0 & 13.0
& 76.0 & 67.0 & 23.0 & 11.0
& 58.0 & 60.0 & 17.0 & 5.0
& 0.512 \\

$S$ only & 41
& 63.0 & 60.0 & 9.0 & 0.0
& 56.0 & 31.0 & 11.0 & 6.0
& 54.0 & 53.0 & 14.0 & 5.0
& 0.386 \\

\midrule
w/o $D$ & 121
& 89.0 & 84.0 & 31.0 & 16.0
& 77.0 & 74.0 & 23.0 & 13.0
& 60.0 & 68.0 & 21.0 & 7.0
& 0.516 \\

w/o $C$ & 81
& 91.0 & 86.0 & 33.0 & 17.0
& 78.0 & 76.0 & 24.0 & 14.0
& 61.0 & 69.0 & 22.0 & 8.0
& 0.521 \\

w/o $I$ & 81
& 88.0 & 83.0 & 29.0 & 14.0
& 75.0 & 72.0 & 22.0 & 12.0
& 58.0 & 65.0 & 20.0 & 6.0
& 0.508 \\

w/o $S$ & 121
& 90.0 & 85.0 & 32.0 & 16.0
& 79.0 & 77.0 & 25.0 & 15.0
& 62.0 & 70.0 & 23.0 & 9.0
& 0.524 \\

\midrule
\textbf{Full $D{+}C{+}I{+}S$} & \textbf{121}
& \textbf{93.0} & \textbf{88.0} & \textbf{35.0} & \textbf{19.0}
& \textbf{81.0} & \textbf{81.0} & \textbf{26.0} & \textbf{16.0}
& \textbf{64.0} & \textbf{73.0} & \textbf{24.0} & \textbf{10.0}
& \textbf{0.530} \\
\bottomrule
\end{tabular}%
}
\end{table*}

\subsection{UBMB Weight Sensitivity}
\label{subsec:ubmb_weight_sensitivity}

We evaluate the sensitivity of \textsc{VISTA}-Proxy to the UBMB weights $(w_d,w_c,w_i,w_s)$ using Qwen3-VL as the probing model; the resulting adversarial examples are transferred to GPT-5, Gemini-3.5-Flash, and Claude Sonnet~4.6 without querying these victims during construction. Table~\ref{tab:weight_sensitivity} reports ASR for four weight configurations. The default setting $(0.40,0.25,0.20,0.15)$ achieves 93.0\%/81.0\%/64.0\% ASR and the highest average across victims (79.3\%, vs.\ 77.0\% for Equal, 75.3\% for S-heavy, and 74.3\% for D-heavy). Individual configurations can favor specific victims: D-heavy raises GPT-5 ASR to 95.0\%, and Equal/S-heavy reach 68.0\% on Claude. However, these gains come at the cost of lower ASR on the other victims, supporting the default weights as a fixed cross-model configuration.

\begin{table}[t]
\centering
\scriptsize
\setlength{\tabcolsep}{3pt}
\renewcommand{\arraystretch}{1.05}
\caption{Sensitivity of \textsc{VISTA}-Proxy to UBMB weight
configurations. Entries report ASR (\%) using Qwen3-VL as the probing model.}
\label{tab:weight_sensitivity}
\resizebox{\linewidth}{!}{
\begin{tabular}{lccccccc}
\toprule
\textbf{Weights}
& $w_d$ & $w_c$ & $w_i$ & $w_s$
& \textbf{GPT-5}
& \textbf{Gemini}
& \textbf{Claude} \\
\midrule
Equal
& .25 & .25 & .25 & .25
& 86.0 & 77.0 & \textbf{68.0} \\
\textbf{Default}
& \textbf{.40} & \textbf{.25} & \textbf{.20} & \textbf{.15}
& 93.0 & \textbf{81.0} & 64.0 \\
D-heavy
& .55 & .20 & .15 & .10
& \textbf{95.0} & 75.0 & 53.0 \\
S-heavy
& .25 & .20 & .15 & .40
& 85.0 & 73.0 & \textbf{68.0} \\
\bottomrule
\end{tabular}}
\end{table}


\subsection{Occlusion Patch-Size Sensitivity}
\label{subsec:patch_sensitivity}

We examine the sensitivity of UBMB attribution to the occlusion patch size in Table~\ref{tab:ubmb_patch}. Very small patches ($16\times16$) produce lower and more variable UBMB scores, whereas larger patches provide more stable responses at the cost of spatial localization granularity. We therefore use $32\times32$ as the default, providing a practical balance between semantic response stability and fine-grained localization.

\begin{table}[t]
\centering
\small
\caption{Sensitivity of UBMB scores to occlusion patch size on a
representative source image.}
\label{tab:ubmb_patch}
\setlength{\tabcolsep}{5pt}
\renewcommand{\arraystretch}{1.10}
\begin{tabular}{cccc}
\toprule
\textbf{Patch Size (px)}
& \textbf{Mean UBMB}
& \textbf{Max UBMB}
& \textbf{Std.} \\
\midrule
16  & 0.1668 & 0.2200 & 0.0606 \\
32  & 0.2099 & 0.2463 & 0.0132 \\
64  & 0.2167 & 0.2440 & \textbf{0.0098} \\
128 & \textbf{0.2422} & \textbf{0.2865} & 0.0147 \\
\bottomrule
\end{tabular}
\end{table}

\section{Hyperparameter Sensitivity}
\label{subsec:hyperparam}

\begin{figure*}[t]
\centering
\includegraphics[width=\textwidth]{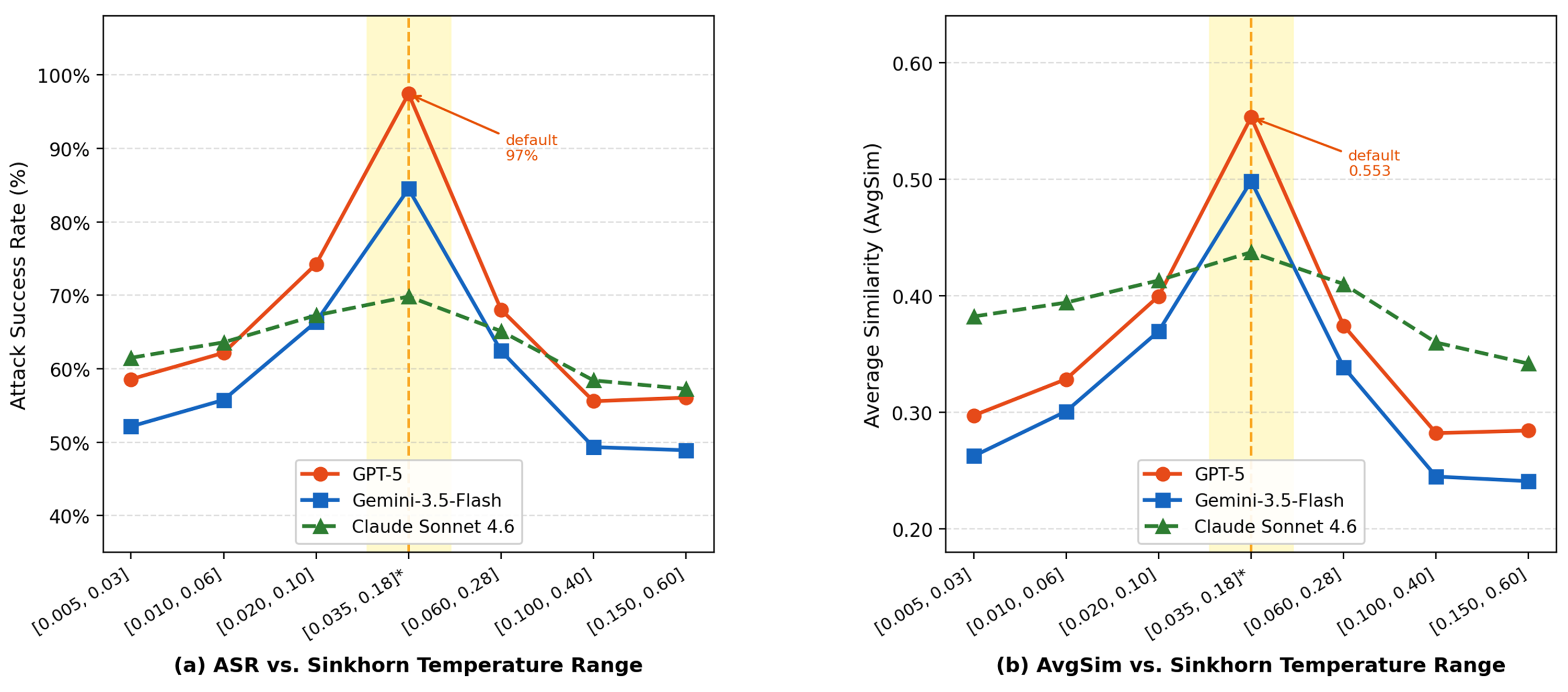}
\caption{Sensitivity to Sinkhorn temperature range $[\epsilon_{\min}, \epsilon_{\max}]$. (a) ASR and (b) AvgSim across seven configurations for GPT-5, Gemini-3.5-Flash, and Claude Sonnet~4.6. The yellow band and dashed orange line mark the default $[0.035, 0.18]$, which achieves the best trade-off. Too-narrow ranges over-concentrate perturbations; too-wide ranges dilute the adversarial signal.}
\label{fig:sinkhorn_temp}
\end{figure*}

\paragraph{Sinkhorn Temperature Range.}
Figure~\ref{fig:sinkhorn_temp} evaluates seven $[\epsilon_{\min}, \epsilon_{\max}]$ configurations for the entropy-adaptive temperature, ranging from a very narrow $[0.005, 0.03]$ to a very wide $[0.150, 0.60]$.

At the \textbf{narrow extreme} $[0.005, 0.03]$, the temperature is nearly constant and low, causing the Sinkhorn plan to over-concentrate perturbation energy on too few high-UBMB patches. ASR degrades from 97\% to ${\approx}59\%$ on GPT-5 and from 85\% to ${\approx}53\%$ on Gemini-3.5-Flash, while AvgSim drops to 0.30 and 0.26, respectively.

At the \textbf{wide extreme} $[0.150, 0.60]$, the transport plan collapses toward near-uniform allocation, diluting the adversarial signal across all patches regardless of their UBMB score. This yields similar degradation (${\approx}57\%$ ASR on GPT-5, AvgSim ${\approx}0.29$; ${\approx}49\%$ on Gemini-3.5-Flash).

Our \textbf{default range} $[0.035, 0.18]$ achieves the best ASR/AvgSim trade-off across all three victims: \textbf{97\%}/\textbf{0.55} on GPT-5, \textbf{85\%}/\textbf{0.50} on Gemini-3.5-Flash, and \textbf{70\%}/\textbf{0.44} on Claude Sonnet~4.6. These ASR values match the \textsc{VISTA}-Victim results in Table~\ref{tab:closed_source_results}.

Claude Sonnet~4.6 exhibits markedly lower sensitivity to the temperature range: its ASR varies by only ${\approx}13$ percentage points across all configurations (57\%--70\%), compared with ${\approx}40$ points for GPT-5 (57\%--97\%). This suggests that Claude's greater inherent robustness limits the gains attainable from any spatial allocation strategy.

\subsection{Hyperparameter Settings}
\label{sec:hyperparameters}

For reproducibility, Table~\ref{tab:hyperparameters} lists the principal hyperparameters of \textsc{VISTA} with their roles and default values. These parameters are fixed across all victim MLLMs and evaluation samples, with no victim-specific tuning, so performance differences reflect the proposed semantic vulnerability modeling and adaptive routing rather than per-victim optimization. Sensitivity analyses for the UBMB weights, occlusion patch size, Sinkhorn temperature, perturbation budget, and optimization steps are reported in Appendices~\ref{subsec:ubmb_weight_sensitivity}--\ref{subsec:optim_steps}.

\begin{table}[t]
\centering
\caption{Hyperparameter settings for \textsc{VISTA}. Parameters are
fixed across all victim MLLMs with no victim-specific tuning.}
\label{tab:hyperparameters}
\scriptsize
\setlength{\tabcolsep}{2.5pt}
\renewcommand{\arraystretch}{1.08}
\resizebox{\columnwidth}{!}{%
\begin{tabular}{lll}
\toprule
\textbf{Param.} & \textbf{Meaning} & \textbf{Default} \\
\midrule
\multicolumn{3}{l}{\textit{Attack setting}} \\
$\epsilon$ & $\ell_\infty$ perturbation budget & $16/255$ \\
$T$ & MI-FGSM steps & $300$ \\
\midrule
\multicolumn{3}{l}{\textit{Semantic probing and UBMB}} \\
$p_0$ & Probing patch size & $32$ \\
$M$ & Probing masks per image & $40$ \\
$(w_d,w_c,w_i,w_s)$ & UBMB signal weights & $(.40,.25,.20,.15)$ \\
$\gamma$ & UBMB sharpening exponent & $2$ \\
\midrule
\multicolumn{3}{l}{\textit{Sinkhorn routing (spatial budget)}} \\
$[\epsilon_{\min},\epsilon_{\max}]$ & Adaptive temperature range & $[0.035,0.18]$ \\
\midrule
\multicolumn{3}{l}{\textit{Surrogate weighting and patch routing}} \\
$\sigma_p$ & Patch-size compatibility scale & $0.4p_0$ \\
$\alpha$ & Spatial-agreement strength & $1.0$ \\
$\lambda_p$ & Paradigm-level adjustment & $0.6$ \\
$\beta_{1,2}$ & SSL role weights & $.60,.40$ \\
$\beta_{3,4,5}$ & Cross-modal role weights & $.50,.30,.20$ \\
$\beta_{6,7,8}$ & Multimodal role weights & $.40,.30,.30$ \\
$\nu$ & Vulnerability/role mixing & $0.35$ \\
$K$ & Selected tokens per surrogate & Adaptive, $[20,64]$ \\
\midrule
\multicolumn{3}{l}{\textit{Attack objective}} \\
$\lambda_g$ & Global alignment weight & $1.0$ \\
$\lambda_t$ & Token alignment weight & $0.2$ \\
$\epsilon_{\mathrm{OT}}$ & Token-transport regularization & $0.1$ \\
$n_{\mathrm{OT}}$ & Sinkhorn iterations & $30$ \\
$\hat{w}_i$ & Final surrogate weights & $\tilde{w}_i$ + rescaling~\cite{jia2026adversarial} \\
\midrule
\multicolumn{3}{l}{\textit{Gradient routing and early stopping}} \\
$\mu$ & EMA decay & $0.9$ \\
$\tau_U$ & High-vulnerability threshold & Top $35\%$ \\
$\tau_m$ & Convergence threshold & $0.85$ \\
\bottomrule
\end{tabular}}
\end{table}

\section{Impact of Perturbation Budget}
\label{subsec:ablation_budget}

We study the effect of the $\ell_\infty$ perturbation budget on attack effectiveness by evaluating $\epsilon\in\{4,8,16\}/255$, following M-Attack~\cite{li2026frustratingly} and FOA-Attack~\cite{jia2026adversarial}; $\epsilon=16/255$ is the default in the main experiments. For consistency with the main-paper protocol, the \textsc{VISTA} results in Table~\ref{tab:ablation_epsilon} correspond to \textsc{VISTA}-Victim.

At the restrictive budget of $\epsilon=4/255$, all methods experience a substantial reduction in attack success. \textsc{VISTA}-Victim still achieves the highest GPT-5 ASR (26.0\%, vs.\ 18.0\% for M-Attack and 20.0\% for FOA-Attack) and $\mathrm{KMR}@{1}$ (30.0\%). On Gemini-3.5-Flash and Claude Sonnet~4.6, however, all methods remain low and FOA-Attack is marginally ahead (12.0\% and 3.0\% vs.\ 11.0\% and 1.0\%), indicating that very small budgets leave too little capacity for targeted semantic redirection on these victims.

Increasing the budget to $\epsilon=8/255$ yields a pronounced gain for \textsc{VISTA}-Victim on GPT-5, where it reaches 82.0\% ASR, an improvement of 19.0 and 16.0 percentage points over M-Attack and FOA-Attack, respectively. On Gemini-3.5-Flash it achieves 46.0\%, comparable to M-Attack (45.0\%), while Claude Sonnet~4.6 remains difficult, with all methods at or below 5.0\%.

At the default $\epsilon=16/255$, \textsc{VISTA}-Victim achieves 97.0\%/85.0\%/70.0\% ASR on GPT-5/Gemini-3.5-Flash/Claude Sonnet~4.6. The largest improvement occurs on Claude, where \textsc{VISTA}-Victim reaches 70.0\% compared with 20.0\% for FOA-Attack and 9.0\% for M-Attack. Overall, increasing the perturbation budget substantially improves targeted transfer, and \textsc{VISTA}'s advantage on the most robust victim emerges once the budget is sufficient to support semantically targeted perturbations.

\begin{table*}[t]
\centering
\scriptsize
\setlength{\tabcolsep}{3.5pt}
\renewcommand{\arraystretch}{1.08}
\caption{Effect of the $\ell_\infty$ perturbation budget
$\epsilon\in\{4,8,16\}/255$ on targeted attack performance
($N{=}100$). \textsc{VISTA}-Victim is used for consistency with the
corresponding main-paper evaluation. All metrics are reported in \%; higher values indicate stronger
target-semantic alignment or attack success. \textbf{Bold} denotes the
best result within each perturbation budget and victim.}
\label{tab:ablation_epsilon}
\resizebox{\textwidth}{!}{%
\begin{tabular}{clcccccccccccc}
\toprule
\multirow{2}{*}{$\boldsymbol{\epsilon}$} &
\multirow{2}{*}{\textbf{Method}}
& \multicolumn{4}{c}{\textbf{GPT-5}}
& \multicolumn{4}{c}{\textbf{Gemini-3.5-Flash}}
& \multicolumn{4}{c}{\textbf{Claude Sonnet~4.6}} \\
\cmidrule(lr){3-6}
\cmidrule(lr){7-10}
\cmidrule(lr){11-14}
& & $\mathrm{KMR}@{1}$ & $\mathrm{KMR}@{2}$
& $\mathrm{KMR}@{3}$ & \textbf{ASR}
& $\mathrm{KMR}@{1}$ & $\mathrm{KMR}@{2}$
& $\mathrm{KMR}@{3}$ & \textbf{ASR}
& $\mathrm{KMR}@{1}$ & $\mathrm{KMR}@{2}$
& $\mathrm{KMR}@{3}$ & \textbf{ASR} \\
\midrule

\multirow{3}{*}{$4/255$}
& M-Attack~\cite{li2026frustratingly}
& 21.0 & 8.0 & 1.0 & 18.0
& 15.0 & 5.0 & 0.0 & 10.0
& 3.0 & 1.0 & 0.0 & 2.0 \\

& FOA-Attack~\cite{jia2026adversarial}
& 24.0 & 9.0 & 2.0 & 20.0
& 17.0 & 6.0 & 1.0 & \textbf{12.0}
& 4.0 & 1.0 & 0.0 & \textbf{3.0} \\

& \textbf{\textsc{VISTA}-Victim}
& \textbf{30.0} & \textbf{16.0} & \textbf{3.0} & \textbf{26.0}
& \textbf{20.0} & \textbf{11.0} & \textbf{2.0} & 11.0
& \textbf{5.0} & 1.0 & 0.0 & 1.0 \\

\midrule

\multirow{3}{*}{$8/255$}
& M-Attack~\cite{li2026frustratingly}
& 58.0 & \textbf{31.0} & 7.0 & 63.0
& 42.0 & 18.0 & 4.0 & 45.0
& 6.0 & 2.0 & 0.0 & 4.0 \\

& FOA-Attack~\cite{jia2026adversarial}
& 62.0 & 28.0 & 9.0 & 66.0
& 35.0 & 20.0 & 4.0 & 42.0
& \textbf{8.0} & 2.0 & 0.0 & \textbf{5.0} \\

& \textbf{\textsc{VISTA}-Victim}
& \textbf{74.0} & 20.0 & \textbf{12.0} & \textbf{82.0}
& \textbf{46.0} & \textbf{32.0} & \textbf{8.0} & \textbf{46.0}
& \textbf{8.0} & \textbf{3.0} & 0.0 & \textbf{5.0} \\

\midrule

\multirow{3}{*}{$16/255$}
& M-Attack~\cite{li2026frustratingly}
& 87.0 & \textbf{63.0} & 24.0 & 96.0
& 79.0 & 55.0 & 14.0 & 81.0
& 11.0 & 4.0 & 1.0 & 9.0 \\

& FOA-Attack~\cite{jia2026adversarial}
& \textbf{91.0} & 42.0 & \textbf{30.0} & 92.0
& 78.0 & \textbf{59.0} & 13.0 & 84.0
& 26.0 & 5.0 & 3.0 & 20.0 \\

& \textbf{\textsc{VISTA}-Victim}
& \textbf{91.0} & 39.0 & 22.0 & \textbf{97.0}
& \textbf{80.0} & 29.0 & \textbf{17.0} & \textbf{85.0}
& \textbf{78.0} & \textbf{28.0} & \textbf{13.0} & \textbf{70.0} \\

\bottomrule
\end{tabular}%
}
\end{table*}

\section{Impact of Optimization Steps}
\label{subsec:optim_steps}

Table~\ref{tab:optim_steps} reports the attack performance and average per-image optimization time of \textsc{VISTA}-Proxy across four step configurations, evaluated on 100 randomly selected images using Qwen3-VL as the probing model.

\paragraph{Step-count effect on ASR.} ASR increases steadily with the number of optimization steps. Even at 100 steps, \textsc{VISTA}-Proxy reaches 52.0\% ASR on Claude Sonnet~4.6, well above the 300-step surrogate-only baselines (20.0\% for FOA-Attack and 9.0\% for M-Attack; Table~\ref{tab:closed_source_results}), indicating that semantically informed spatial routing is effective early in optimization on robust victims. At 200 steps, \textsc{VISTA}-Proxy already attains over 95\% of its 300-step ASR on all three victims (91.0\%/78.0\%/61.0\% vs.\ 93.0\%/81.0\%/64.0\%). Claude Sonnet~4.6 shows the largest relative gain from 100 to 300 steps (52.0\% to 64.0\%, a 23\% relative increase, vs.\ 15\% for GPT-5 and 19\% for Gemini-3.5-Flash), suggesting that more robust victims benefit from more sustained optimization.

\paragraph{Adaptive early stopping.} The EMA-based adaptive gradient routing monitors the optimal transport (OT) loss during optimization and terminates once it reaches $-8.9$, a threshold empirically identified as the point beyond which the adversarial example consistently achieves successful semantic redirection. This criterion reduces the average per-image time from 6.35 to 3.28 min (a 48.3\% reduction) while costing only one percentage point of ASR on each victim (92.0\%/80.0\%/63.0\% vs.\ 93.0\%/81.0\%/64.0\%) and 0.04--0.10 in AvgSim.

\begin{table*}[t]
\centering
\small
\setlength{\tabcolsep}{5pt}
\renewcommand{\arraystretch}{1.15}
\caption{Attack performance and per-image optimization time of \textsc{VISTA}-Proxy at different step budgets (100 randomly selected images, Qwen3-VL as local proxy). ASR is reported in \%; AvgSim is a similarity score in $[0,1]$. Adaptive stopping terminates when OT loss $\leq -8.9$. \textbf{Bold} denotes the best ASR/AvgSim and the most efficient runtime.}
\label{tab:optim_steps}
\resizebox{\textwidth}{!}{%
\begin{tabular}{lcccccccc}
\toprule
\multirow{2}{*}{\textbf{Steps}} & \multirow{2}{*}{\textbf{Avg.\ Time (min/img)}}
  & \multicolumn{2}{c}{\textbf{GPT-5}}
  & \multicolumn{2}{c}{\textbf{Gemini-3.5-Flash}}
  & \multicolumn{2}{c}{\textbf{Claude Sonnet 4.6}} \\
\cmidrule(lr){3-4}\cmidrule(lr){5-6}\cmidrule(lr){7-8}
& & ASR & AvgSim & ASR & AvgSim & ASR & AvgSim \\
\midrule
100  & 2.12 & 81.0 & 0.43 & 68.0 & 0.36 & 52.0 & 0.31 \\
200  & 4.24 & 91.0 & 0.48 & 78.0 & 0.43 & 61.0 & 0.38 \\
300  & 6.35 & \textbf{93.0} & \textbf{0.54} & \textbf{81.0} & \textbf{0.51} & \textbf{64.0} & \textbf{0.50} \\
\midrule
Adaptive (EMA, $\leq{-}8.9$) & \textbf{3.28} & 92.0 & 0.50 & 80.0 & 0.45 & 63.0 & 0.40 \\
\bottomrule
\end{tabular}
}
\end{table*}

\section{Query and Computational Efficiency}
\label{sec:efficiency}

We evaluate \textsc{VISTA} in terms of construction queries, optimization steps, wall-clock runtime, and attack success rate (ASR). We distinguish queries to a separate probing model, $Q_p$, from queries to the final victim, $Q_v$. In \textsc{VISTA}-Proxy, semantic probing is performed locally with
an open-source MLLM ($f_p\neq f_v$), so $Q_v=0$. In \textsc{VISTA}-Victim, the victim itself is used for semantic probing ($f_p=f_v$), and the corresponding probing calls are counted toward $Q_v$. Post-hoc evaluation queries are excluded from the construction budget.
For each attack instance, let
\begin{equation}
Q_{\mathrm{probe}}
=
Q_{\mathrm{base}} + Q_{\mathrm{occ}},
\end{equation}
where $Q_{\mathrm{base}}$ establishes the unoccluded semantic reference and
$Q_{\mathrm{occ}}$ accounts for spatially occluded views. Coverage-based
early termination may stop probing before all candidate masks are evaluated.
The resulting query allocation is
\begin{equation}
(Q_p,Q_v)=
\begin{cases}
(Q_{\mathrm{probe}},0),
& \textsc{VISTA}\text{-Proxy},\\
(0,Q_{\mathrm{probe}}),
& \textsc{VISTA}\text{-Victim}.
\end{cases}
\end{equation}
\subsection{Overall Construction Efficiency}

\begin{table*}[t]
\centering
\small
\setlength{\tabcolsep}{4.5pt}
\renewcommand{\arraystretch}{1.10}
\caption{Construction efficiency and ASR comparison. $Q_v$ denotes queries
to the final victim, while $Q_p$ denotes queries to a separate probing model.
\textsc{VISTA} uses 40 probing masks, resulting in 121 model queries per
image. ASR is reported in \%.}
\label{tab:efficiency}
\resizebox{\linewidth}{!}{
\begin{tabular}{lccccccc}
\toprule
\textbf{Method}
& $\mathbf{Q_v}$/img
& $\mathbf{Q_p}$/img
& \textbf{Steps}
& \textbf{Min/img}
& \multicolumn{3}{c}{\textbf{ASR (\%)}} \\
\cmidrule(lr){6-8}
& & & & &
\textbf{GPT-5} &
\textbf{Gemini-3.5-Flash} &
\textbf{Claude Sonnet 4.6} \\
\midrule
AttackVLM (ADV-Transfer, B/16)
& 0 & 0 & 300 & \textbf{1.00}
& 4.0 & 1.0 & 3.0 \\
AttackVLM (ADV-Query, B/16)
& 800 & 0 & 8 & 20.00
& 3.0 & 2.0 & 0.0 \\
M-Attack
& 0 & 0 & 300 & 1.12
& 96.0 & 81.0 & 9.0 \\
FOA-Attack
& 0 & 0 & 300 & 3.19
& 92.0 & 84.0 & 20.0 \\
\midrule
\textsc{VISTA}-Proxy
& 0 & 121 (local) & 300 & 6.35
& 93.0 & 81.0 & 64.0 \\
\textsc{VISTA}-Proxy (Adaptive)
& 0 & 121 (local) & Adaptive & 3.28
& 92.0 & 80.0 & 63.0 \\
\textsc{VISTA}-Victim
& 121 & 0 & 300 & 9.10
& \textbf{97.0} & \textbf{85.0} & \textbf{70.0} \\
\bottomrule
\end{tabular}}
\end{table*}

Table~\ref{tab:efficiency} compares construction cost, runtime, and attack effectiveness across transfer- and query-based methods. The pure-transfer baselines---AttackVLM (ADV-Transfer), M-Attack, and FOA-Attack---require no final-victim queries during construction ($Q_v=0$), whereas AttackVLM (ADV-Query) uses approximately 800 victim queries per image. With 40 probing masks, \textsc{VISTA}-Proxy preserves the pure-transfer setting, requiring 121 local Qwen3-VL queries per image while achieving 93.0\%/81.0\%/64.0\% ASR on GPT-5, Gemini-3.5-Flash, and Claude Sonnet~4.6, respectively. This comes at a higher runtime (6.35 vs.\ 1.00--3.19 min/image for the pure-transfer baselines); its adaptive variant reduces runtime to 3.28 min/image, comparable to FOA-Attack (3.19 min/image), a 48.3\% reduction at a cost of only one percentage point of ASR on each victim.

When direct victim probing is allowed, \textsc{VISTA}-Victim uses the same 40-mask configuration, corresponding to 121 victim queries per image---84.9\% fewer than AttackVLM (ADV-Query)---and achieves the highest ASR on all three victims (97.0\%/85.0\%/70.0\%). Overall, \textsc{VISTA}-Proxy offers strong query-free transfer to the final victim, while \textsc{VISTA}-Victim trades a moderate victim-query budget for improved attack success.

\subsection{Query-Budget Sensitivity}
\label{subsec:query_asr}

\begin{figure}[t]
    \centering
    \includegraphics[width=\columnwidth]{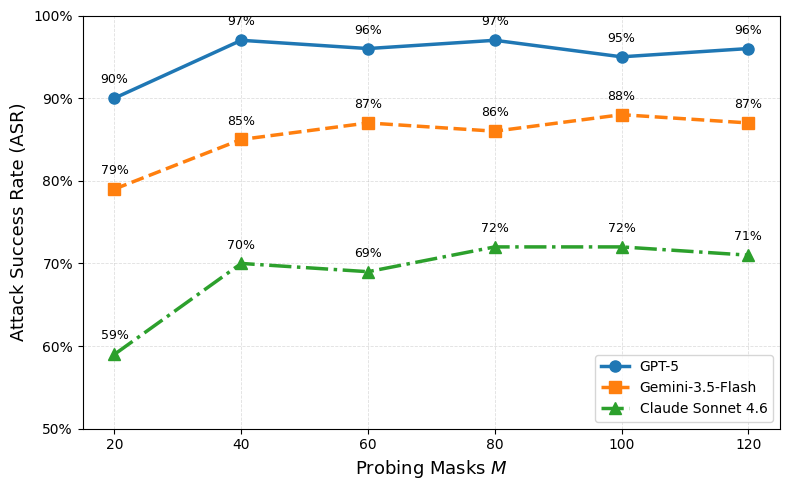}
   \caption{Mask--ASR trade-off of \textsc{VISTA}-Victim. Probing-mask settings
    $M\in\{20,40,60,80,100,120\}$ correspond to victim-query budgets
    $Q_v\in\{61,121,181,241,301,361\}$.}
    \label{fig:query_asr}
\end{figure}

\paragraph{Query--ASR Analysis.}
Figure~\ref{fig:query_asr} examines the sensitivity of \textsc{VISTA}-Victim to the number of probing masks $M$ and the corresponding victim-query budget $Q_v{=}3M{+}1$ (one base query plus three prompts per mask). We evaluate all source--target pairs with a fixed $32\times32$ patch size. Increasing the number of masks from $M{=}20$ ($Q_v{=}61$) to $M{=}40$ ($Q_v{=}121$) yields the largest improvement, raising ASR from 90.0\% to 97.0\% on GPT-5, from 79.0\% to 85.0\% on Gemini-3.5-Flash, and from 59.0\% to 70.0\% on Claude Sonnet~4.6. Beyond $M{=}40$, performance remains stable, with only minor fluctuations; at $M{=}120$ ($Q_v{=}361$), ASR reaches 96.0\%, 87.0\%, and 71.0\%, respectively. These results indicate diminishing returns beyond moderate mask counts. Accordingly, the \textsc{VISTA}-Victim results in Table~\ref{tab:closed_source_results} use $M{=}40$ ($Q_v{=}121$ victim queries per image) as a practical trade-off between attack effectiveness and query cost.

\noindent\textbf{Query-Matched Baseline.}
We use AttackVLM (ADV-Query) as the query-based baseline because direct victim-query targeted attacks on generative MLLMs remain comparatively limited, while most black-box methods primarily rely on transfer without querying the final victim. AttackVLM is particularly relevant because its ADV-Query variant directly queries the victim to refine a transferred adversarial example through gradient-free estimation. For a controlled comparison, we evaluate AttackVLM and \textsc{VISTA}-Victim under identical final-victim query budgets $Q_v$.

As shown in Table~\ref{tab:query_matched}, \textsc{VISTA}-Victim consistently outperforms AttackVLM across all matched budgets and victims. With $M{=}20$ probing masks ($Q_v{=}61$), \textsc{VISTA}-Victim achieves 90.0\%, 79.0\%, and 59.0\% ASR on GPT-5, Gemini-3.5-Flash, and Claude Sonnet~4.6, respectively, compared with 1.0\%, 0.0\%, and 0.0\% for AttackVLM. Increasing to $M{=}40$ ($Q_v{=}121$) raises \textsc{VISTA}-Victim's ASR to 97.0\%, 85.0\%, and 70.0\%, while AttackVLM remains at or below 1.0\%. At $M{=}80$ ($Q_v{=}241$), \textsc{VISTA}-Victim achieves 97.0\%, 86.0\%, and 72.0\%, whereas AttackVLM remains at or below 2.0\%. These results show that the advantage of \textsc{VISTA}-Victim persists under equal victim-query budgets and is not explained by greater access to the victim. The small gains beyond $M{=}40$ (at most 2.0 percentage points) further indicate diminishing returns from increasing the probing budget.

\begin{table}[t]
\centering
\small
\setlength{\tabcolsep}{4.0pt}
\renewcommand{\arraystretch}{1.08}
\caption{Query-matched comparison with AttackVLM (ADV-Query). Both methods use the same final-victim query budget $Q_v$ during adversarial construction; $M$ is the number of probing masks for \textsc{VISTA}-Victim ($Q_v{=}3M{+}1$). Values report ASR (\%).}
\label{tab:query_matched}
\resizebox{\linewidth}{!}{
\begin{tabular}{lccccc}
\toprule
\textbf{Method} & $\mathbf{M}$ & $\mathbf{Q_v}$ & \textbf{GPT-5} & \textbf{Gemini} & \textbf{Claude} \\
\midrule
AttackVLM (ADV-Query) & -- & 61  & 1.0 & 0.0 & 0.0 \\
\textsc{VISTA}-Victim  & 20 & 61  & \textbf{90.0} & \textbf{79.0} & \textbf{59.0} \\
\midrule
AttackVLM (ADV-Query) & -- & 121 & 1.0 & 1.0 & 0.0 \\
\textsc{VISTA}-Victim  & 40 & 121 & \textbf{97.0} & \textbf{85.0} & \textbf{70.0} \\
\midrule
AttackVLM (ADV-Query) & -- & 241 & 2.0 & 1.0 & 0.0 \\
\textsc{VISTA}-Victim  & 80 & 241 & \textbf{97.0} & \textbf{86.0} & \textbf{72.0} \\
\bottomrule
\end{tabular}}
\end{table}

\subsection{Construction and Evaluation API Cost}
\label{subsec:api_cost}

Table~\ref{tab:eval_unit_cost} reports the per-image API cost of adversarial construction and post-hoc evaluation. For construction, we use $M{=}40$ probing masks, corresponding to 121 model queries per image. In \textsc{VISTA}-Victim, these queries are issued directly to the final victim ($Q_v{=}121$), resulting in estimated construction costs of approximately \$0.154, \$0.207, and \$0.213 per image for GPT-5, Gemini-3.5-Flash, and Claude Sonnet~4.6, respectively.
\textsc{VISTA}-Proxy uses the same probing configuration but issues all 121 queries to the local Qwen3-VL proxy ($Q_v{=}0$), and therefore incurs no commercial API cost. Post-hoc evaluation costs are reported separately and are excluded from the adversarial construction budget.

\begin{table}[t]
\centering
\small
\setlength{\tabcolsep}{3.0pt}
\renewcommand{\arraystretch}{1.08}
\caption{Per-image construction and evaluation API cost. Construction
uses $M{=}40$ probing masks, corresponding to 121 model queries per
image. \textsc{VISTA}-Proxy probes a locally hosted model and incurs no API cost.}
\label{tab:eval_unit_cost}
\begin{tabular}{@{}lccc@{}}
\toprule
\textbf{Stage} & \textbf{Calls/img} & \textbf{\$/call} & \textbf{\$/img} \\
\midrule
\multicolumn{4}{@{}l}{\textit{Adversarial construction}} \\
\textsc{VISTA}-Proxy (Qwen3-VL) & 121 & local & -- \\
\textsc{VISTA}-Victim (GPT-5)    & 121 & \$0.00127 & \$0.1537 \\
\textsc{VISTA}-Victim (Gemini)   & 121 & \$0.00171 & \$0.2069 \\
\textsc{VISTA}-Victim (Claude)   & 121 & \$0.00176 & \$0.2130 \\
\midrule
\multicolumn{4}{@{}l}{\textit{Post-hoc evaluation}} \\
GPT-5 caption              & 2 & \$0.00206 & \$0.0041 \\
Gemini-3.5-Flash caption   & 2 & \$0.00249 & \$0.0050 \\
Claude Sonnet~4.6 caption  & 2 & \$0.00108 & \$0.0022 \\
GPT-4o GPTScore            & 3 & \$0.00098 & \$0.0029 \\
GPT-4o keyword match       & 3 & \$0.00111 & \$0.0033 \\
\midrule
\textbf{Evaluation total}  & \textbf{12} & -- & \textbf{\$0.0175} \\
\bottomrule
\end{tabular}
\end{table}

\subsection{UBMB Signal-Level Query and Cost Breakdown}

Table~\ref{tab:ubmb_ablation_cost_n100} reports the query and computational cost of individual UBMB signal configurations on $N{=}100$ images, using the same $M{=}40$ probing masks as Table~\ref{tab:eval_unit_cost}. The number of probing queries per image follows directly from the responses each signal requires (App.~\ref{app:ubmb_details}). Every configuration first issues one query on the clean image to obtain the reference response $r_0$. Response drift $D$ and semantic shift $S$ both compare $r_0$ with the base-prompt response $r_j$ on each masked image, so they share the same $M$ queries. Conflict susceptibility $C$ requires the conflict-prompt response $c_j$, and instruction-override susceptibility $I$ requires the misleading-prompt response $m_j$, each adding $M$ queries. Consequently, each single-signal variant requires $1{+}M{=}41$ queries/image; configurations combining $D$ or $S$ with exactly one of $C$ or $I$ (w/o $C$ and w/o $I$) require $1{+}2M{=}81$; and configurations using all three prompts (w/o $D$, w/o $S$, and the full $D{+}C{+}I{+}S$) require $1{+}3M{=}121$. The full UBMB configuration thus requires 12,100 local probing queries over 100 images and approximately 15.15 GPU-hours of attack computation. Attack GPU-hours include both local Qwen3-VL probing and surrogate optimization; they therefore exceed the optimization-only runtime reported in Table~\ref{tab:efficiency} (6.35~min/image), and vary across configurations with the same query count because different prompts produce responses of different lengths, which changes probing time.

Evaluation overhead remains nearly constant because all configurations use the same downstream evaluation protocol: each requires 1,400 evaluation calls (14 calls per image over 100 images), covering caption generation by the three victim models and GPT-4o-based GPTScore and keyword-matching judgments, with evaluation time ranging from 53.5 to 58.0 minutes. Across all nine configurations (9 variants $\times$ 100 images), the study requires 68,900 local probing queries and approximately 132.20 attack GPU-hours, together with 12,600 evaluation calls and 500.5 minutes of evaluation time. All attribution queries are issued to the local Qwen3-VL proxy and therefore do not contribute to the final-victim query budget ($Q_v{=}0$) of \textsc{VISTA}-Proxy.

\begin{table*}[t]
\centering
\setlength{\tabcolsep}{3.5pt}
\renewcommand{\arraystretch}{1.12}
\caption{UBMB signal-level query and computational cost for
\textsc{VISTA}-Proxy on $N=100$ images using $M=40$ probing masks.
All probing queries are local ($Q_v=0$).}
\label{tab:ubmb_ablation_cost_n100}
\resizebox{12cm}{!}{%
\begin{tabular}{lccccc}
\toprule
\textbf{UBMB Signals}
& \textbf{Queries}
& \textbf{Queries/img}
& \textbf{Attack GPU-h}
& \textbf{Eval.\ Calls}
& \textbf{Eval.\ Min} \\
\midrule
$D$ only & 4{,}100  & 41  & 11.05 & 1{,}400 & 56.0 \\
$C$ only & 4{,}100  & 41  & 13.15 & 1{,}400 & 56.0 \\
$I$ only & 4{,}100  & 41  & 13.80 & 1{,}400 & 55.5 \\
$S$ only & 4{,}100  & 41  & 11.90 & 1{,}400 & 55.0 \\
\midrule
w/o $D$  & 12{,}100 & 121 & 21.45 & 1{,}400 & 53.5 \\
w/o $C$  & 8{,}100  & 81  & 15.90 & 1{,}400 & 56.5 \\
w/o $I$  & 8{,}100  & 81  & 13.95 & 1{,}400 & 58.0 \\
w/o $S$  & 12{,}100 & 121 & 15.85 & 1{,}400 & 56.5 \\
\midrule
\textbf{Full $D{+}C{+}I{+}S$} & \textbf{12{,}100} & \textbf{121} & \textbf{15.15} & \textbf{1{,}400} & \textbf{53.5} \\
\midrule
\textbf{Total (9 variants, 900 imgs)} & \textbf{68{,}900} & -- & \textbf{132.20} & \textbf{12{,}600} & \textbf{500.5} \\
\bottomrule
\end{tabular}%
}
\end{table*}

\section{Open-Source Models as Cost-Effective Black-Box Proxies}
\label{subsec:whitebox_proxy}

A practical consideration for black-box semantic probing is the cost of querying commercial MLLMs during adversarial construction. \textsc{VISTA}-Proxy addresses this issue by using locally hosted Qwen3-VL~\cite{bai2025qwen3vltechnicalreport} as the probing model, while transferring the resulting adversarial examples to unseen commercial victims without querying them during construction ($Q_v=0$).

As shown in Table~\ref{tab:proxy_results}, \textsc{VISTA}-Proxy achieves 93\%, 81\%, and 64\% ASR on GPT-5, Gemini-3.5-Flash, and Claude Sonnet~4.6, respectively, compared with 97\%, 85\%, and 70\% for \textsc{VISTA}-Victim. Thus, replacing victim-specific probing with an independent open-source proxy incurs only a modest ASR reduction while eliminating commercial victim queries during adversarial construction. These results demonstrate that semantic
guidance derived from Qwen3-VL transfers effectively across unseen commercial MLLMs, providing a cost-effective alternative when direct victim probing is unavailable or undesirable.

\begin{table*}[t]
\centering
\scriptsize
\setlength{\tabcolsep}{3pt}
\renewcommand{\arraystretch}{1.15}
\caption{\textsc{VISTA} performance when using open-source models (Qwen2.5-VL, Qwen3-VL) as locally hosted proxies with no API cost for black-box semantic probing, evaluated on closed-source commercial MLLMs. The oracle row probes the actual victim (\textsc{VISTA}-Victim); proxy rows correspond to \textsc{VISTA}-Proxy. All metrics are reported in \%. \textbf{Bold} denotes the best proxy result.}
\label{tab:proxy_results}
\resizebox{\textwidth}{!}{%
\begin{tabular}{llcccccccccccc}
\toprule
\multirow{2}{*}{\textbf{Probing Model}} & \multirow{2}{*}{\textbf{Cost}}
  & \multicolumn{4}{c}{\textbf{GPT-5}}
  & \multicolumn{4}{c}{\textbf{Gemini-3.5-Flash}}
  & \multicolumn{4}{c}{\textbf{Claude Sonnet 4.6}} \\
\cmidrule(lr){3-6}\cmidrule(lr){7-10}\cmidrule(lr){11-14}
& & $\mathrm{KMR}@{1}$ & $\mathrm{KMR}@{2}$ & $\mathrm{KMR}@{3}$ & ASR
  & $\mathrm{KMR}@{1}$ & $\mathrm{KMR}@{2}$ & $\mathrm{KMR}@{3}$ & ASR
  & $\mathrm{KMR}@{1}$ & $\mathrm{KMR}@{2}$ & $\mathrm{KMR}@{3}$ & ASR \\
\midrule
\textit{Actual Victim (Oracle)} & \textit{High (API)}
  & \textit{91.0} & \textit{39.0} & \textit{22.0} & \textit{97.0}
  & \textit{80.0} & \textit{29.0} & \textit{17.0} & \textit{85.0}
  & \textit{78.0} & \textit{28.0} & \textit{13.0} & \textit{70.0} \\
\midrule
Qwen3-VL (proxy) & Free (local)
  & \textbf{88.0} & \textbf{35.0} & \textbf{19.0} & \textbf{93.0}
  & \textbf{81.0} & \textbf{26.0} & \textbf{16.0} & \textbf{81.0}
  & \textbf{73.0} & \textbf{24.0} & \textbf{10.0} & \textbf{64.0} \\
Qwen2.5-VL (proxy) & Free (local)
  & 86.0 & 33.0 & 18.0 & 92.0
  & 79.0 & 25.0 & 15.0 & 80.0
  & 71.0 & 22.0 & 9.0 & 62.0 \\
\bottomrule
\end{tabular}
}
\end{table*}

\section{Robustness Against Common Defenses}
\label{subsec:defense_robustness}

Table~\ref{tab:defense_results} evaluates VISTA under eight input-level defenses on 50 randomly selected images against GPT-4o and GPT-5, following the feature squeezing benchmark~\cite{xu2017feature, xie2017mitigating}. Defenses are applied to adversarial examples at inference time, simulating realistic deployment pipelines.

\paragraph{Compression-Resilient Perturbations.}
Under JPEG compression, \textsc{VISTA} maintains consistently high ASR across quality levels 70/80/90, achieving 92\%/90\%/96\% on GPT-5 and 92\%/94\%/90\% on GPT-4o. This stable performance suggests that \textsc{VISTA}'s effectiveness does not rely exclusively on high-frequency perturbation components that are strongly suppressed by JPEG compression.

\paragraph{Robustness to spatial transformations.} Resize defenses at $512{\times}512$ and $768{\times}768$ maintain high attack effectiveness, with only minor degradation overall and a slight improvement on GPT-5 (ASR 96\%/94\% vs.\ 94\% baseline).

\paragraph{Sensitivity to Smoothing Defenses.}
Gaussian noise and median filtering produce the largest ASR reductions among the non-averaging defenses. Under Gaussian noise, ASR decreases to 68\% on both GPT-4o and GPT-5, while median filtering reduces ASR to 78\% and 64\%, respectively. Pixel averaging has the strongest overall effect, lowering ASR to 62\% on GPT-4o and 56\% on GPT-5. These results suggest that stronger spatial smoothing can disrupt perturbation structure in semantically salient regions, although \textsc{VISTA} retains non-trivial attack effectiveness under all evaluated transformations.

\paragraph{Real-World Validation.}
To assess practical transferability, we select 100 adversarial examples generated by \textsc{VISTA}-Proxy and manually upload them across GPT-5, Claude Sonnet 4.6, and Gemini-3.5-Flash through their standard web interfaces, where the underlying preprocessing pipelines are unknown. \textsc{VISTA}-Proxy achieves an overall 92\% success rate, demonstrating that its transferred adversarial examples remain effective under the implicit preprocessing and
transformations applied by real-world commercial MLLM interfaces.

\begin{table*}[t]
\centering
\scriptsize
\setlength{\tabcolsep}{4pt}
\renewcommand{\arraystretch}{1.15}
\caption{Robustness of \textsc{VISTA} under common input-level defenses~\cite{xu2017feature} on 50 randomly selected images (sim-threshold=0.3). ASR and KMR are reported in \%; AvgSim is a similarity score in $[0,1]$. \textbf{Bold} = best per metric per model.}
\label{tab:defense_results}
\resizebox{\textwidth}{!}{%
\begin{tabular}{lcccccccccc}
\toprule
\multirow{2}{*}{\textbf{Defense}}
  & \multicolumn{5}{c}{\textbf{GPT-4o}}
  & \multicolumn{5}{c}{\textbf{GPT-5}} \\
\cmidrule(lr){2-6}\cmidrule(lr){7-11}
& \textbf{ASR} & \textbf{AvgSim} & $\mathrm{KMR}@{1}$ & $\mathrm{KMR}@{2}$ & $\mathrm{KMR}@{3}$
& \textbf{ASR} & \textbf{AvgSim} & $\mathrm{KMR}@{1}$ & $\mathrm{KMR}@{2}$ & $\mathrm{KMR}@{3}$ \\
\midrule
None       & \textbf{96.0} & 0.522 & 90.0 & 44.0 & 8.0 & 94.0 & 0.548 & \textbf{96.0} & 50.0 & 24.0 \\
JPEG-70    & 92.0 & 0.520 & 92.0 & 38.0 & 20.0 & 92.0 & 0.526 & 94.0 & 48.0 & 22.0 \\
JPEG-80    & 94.0 & 0.486 & 92.0 & 40.0 & 12.0 & 90.0 & 0.502 & 92.0 & 42.0 & 16.0 \\
JPEG-90    & 90.0 & 0.518 & 84.0 & 42.0 & \textbf{22.0} & \textbf{96.0} & 0.534 & 92.0 & 44.0 & 26.0 \\
Resize-512 & 92.0 & \textbf{0.566} & \textbf{98.0} & \textbf{50.0} & 16.0 & \textbf{96.0} & \textbf{0.578} & \textbf{96.0} & \textbf{58.0} & \textbf{30.0} \\
Resize-768 & 90.0 & 0.532 & 86.0 & 38.0 & 14.0 & 94.0 & 0.540 & 94.0 & 48.0 & 28.0 \\
Gaussian   & 68.0 & 0.378 & 70.0 & 36.0 & 10.0 & 68.0 & 0.336 & 64.0 & 30.0 & 8.0 \\
Median     & 78.0 & 0.386 & 68.0 & 30.0 & 4.0 & 64.0 & 0.308 & 64.0 & 26.0 & 6.0 \\
Average    & 62.0 & 0.338 & 64.0 & 24.0 & 8.0 & 56.0 & 0.296 & 60.0 & 24.0 & 10.0 \\
\bottomrule
\end{tabular}
}
\end{table*}

\begin{figure}[!t]
\centering
\includegraphics[width=0.5\textwidth]{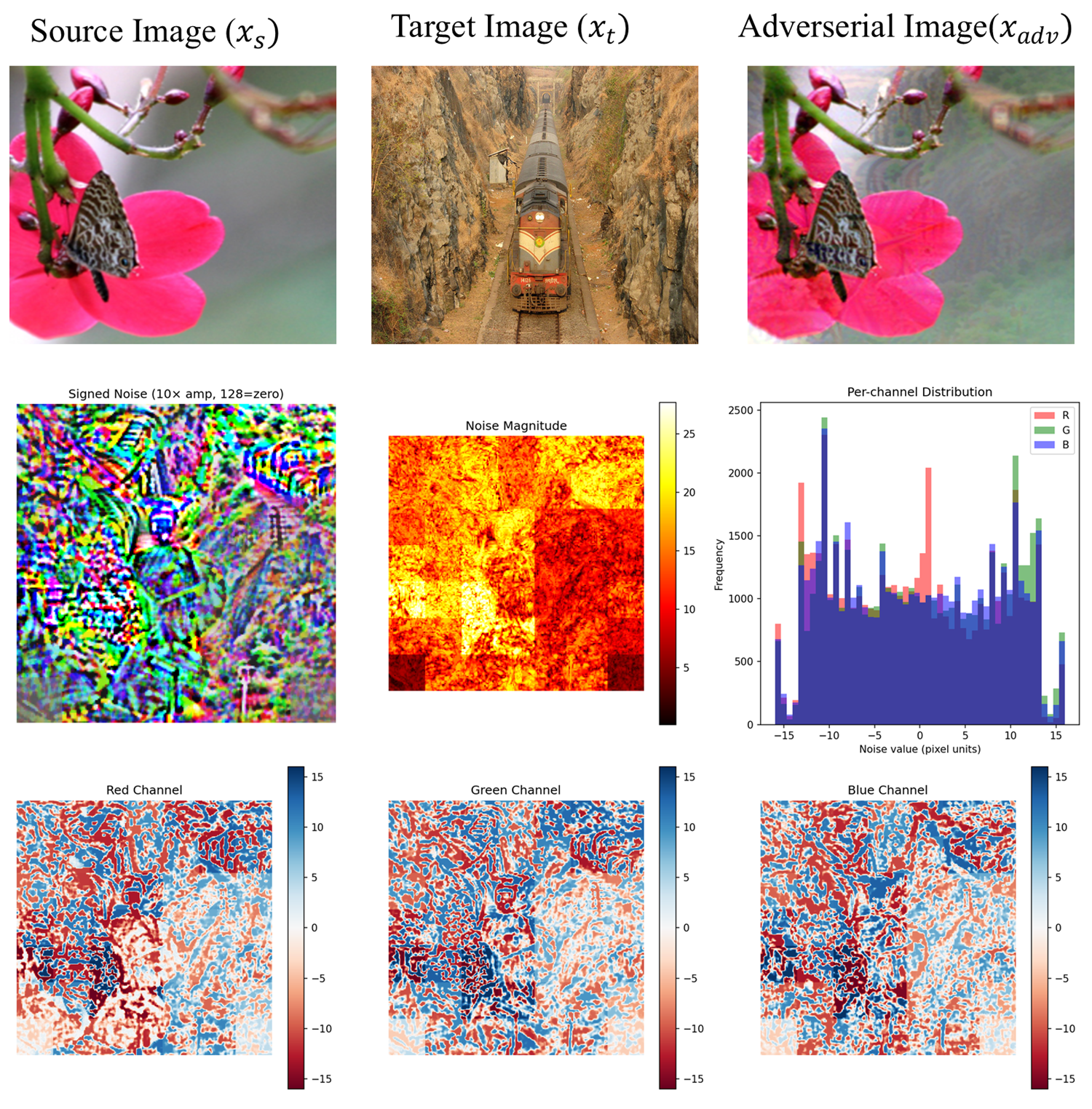}
\caption{Perturbation noise analysis for the butterfly$\to$train adversarial example. \textit{Top left}: signed noise ($10\times$ amplified, 128=zero). \textit{Top center}: noise magnitude heatmap. \textit{Top right}: per-channel (R/G/B) signed noise distribution. \textit{Bottom row}: per-channel signed perturbation maps.}
\label{fig:noise_analysis}
\end{figure}

\section{Perturbation Noise Analysis}
\label{subsec:noise_analysis}

Figure~\ref{fig:noise_analysis} provides a detailed analysis of \textsc{VISTA}'s adversarial perturbation structure on a representative source--target pair: a \textbf{butterfly on pink flowers} (source) redirected toward a \textbf{train in a rocky canyon} (target). The adversarial example is visually indistinguishable from the source, yet the victim MLLM describes it in terms of the target semantics.

\paragraph{Signed noise and spatial structure.} The top-left panel shows the signed noise amplified $10\times$ (with 128 corresponding to zero). Rather than unstructured random noise, \textsc{VISTA} produces spatially coherent patterns aligned with the butterfly's wings, the flower petals, and the stem, i.e., the semantically salient regions of the source that the victim likely relies on for visual grounding.

\paragraph{Noise magnitude map.} The top-center heatmap reveals a spatially non-uniform magnitude distribution, with high-energy zones concentrated in the upper-left (butterfly body and wing area) and lower-right (flower petals). This block-like structure is consistent with the patch-level Sinkhorn routing.

\paragraph{Per-channel distribution.} The top-right histogram shows that all three channels exhibit a bimodal distribution near $\pm 15$ pixel units, close to the $\ell_\infty$ bound of $16/255 \times 255 = 16$. The blue channel carries the heaviest positive tail.

\paragraph{Per-channel spatial maps.} All three channels share the same large-scale spatial organization, with positive perturbation concentrated on the butterfly and flower and negative perturbation in the background, but they differ in fine-grained patterns. This suggests that \textsc{VISTA}'s optimization exploits channel-specific cues rather than applying a uniform perturbation across channels.

\section{Additional Qualitative Results}
\label{subsec:additional_qual}

\begin{figure}[!t]
\centering

\begin{subfigure}[!t]{\columnwidth}
    \centering
    \includegraphics[width=\linewidth, height=3.1cm]
    {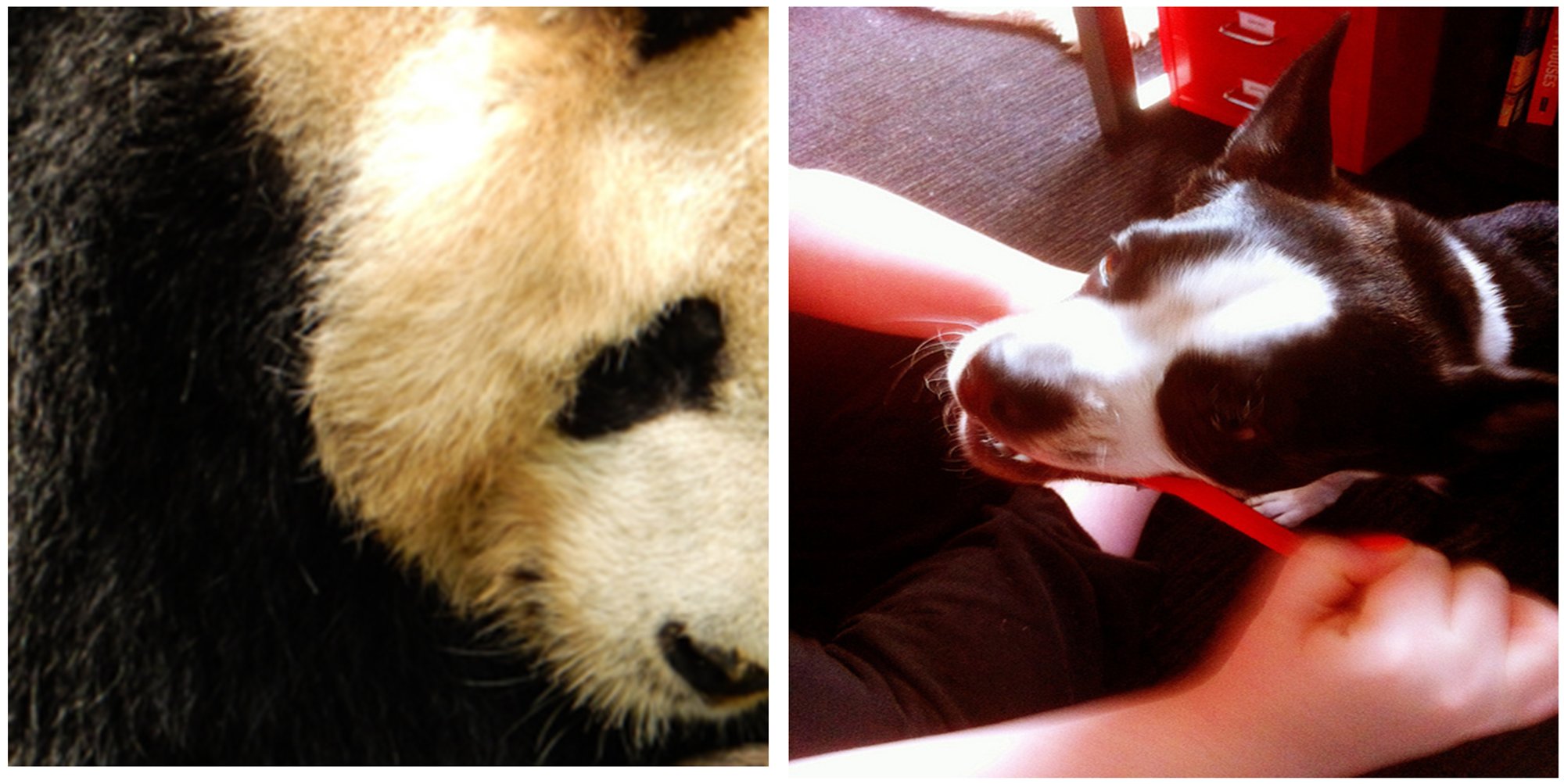}
    \caption{Source and target images.}
    \label{fig:qual1_source}
\end{subfigure}

\begin{subfigure}[t]{\columnwidth}
    \centering
    \includegraphics[width=\linewidth]
    {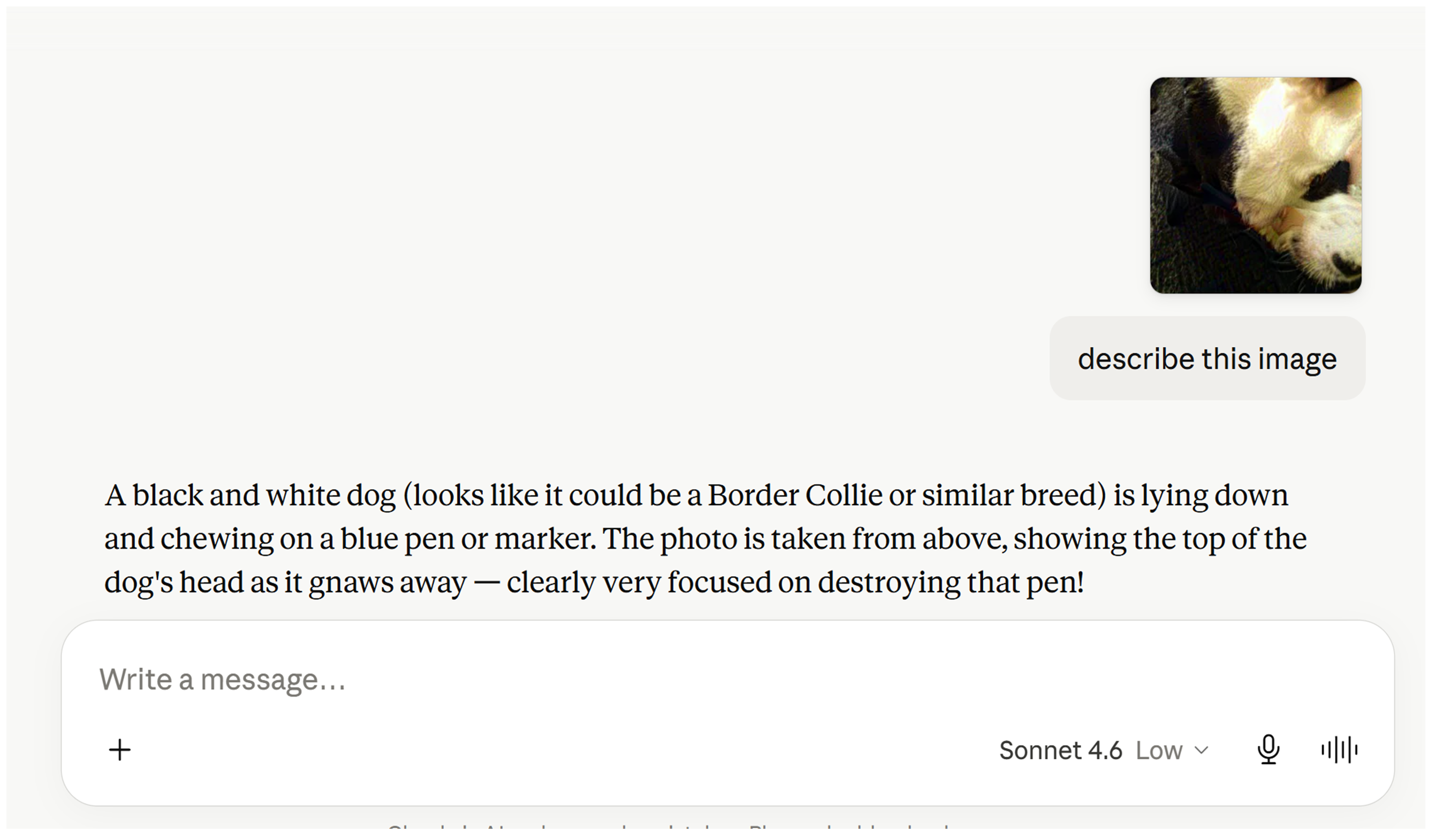}
    \caption{Claude Sonnet 4.6 response.}
    \label{fig:qual1_claude}
\end{subfigure}

\begin{subfigure}[t]{\columnwidth}
    \centering
    \includegraphics[width=\linewidth]
    {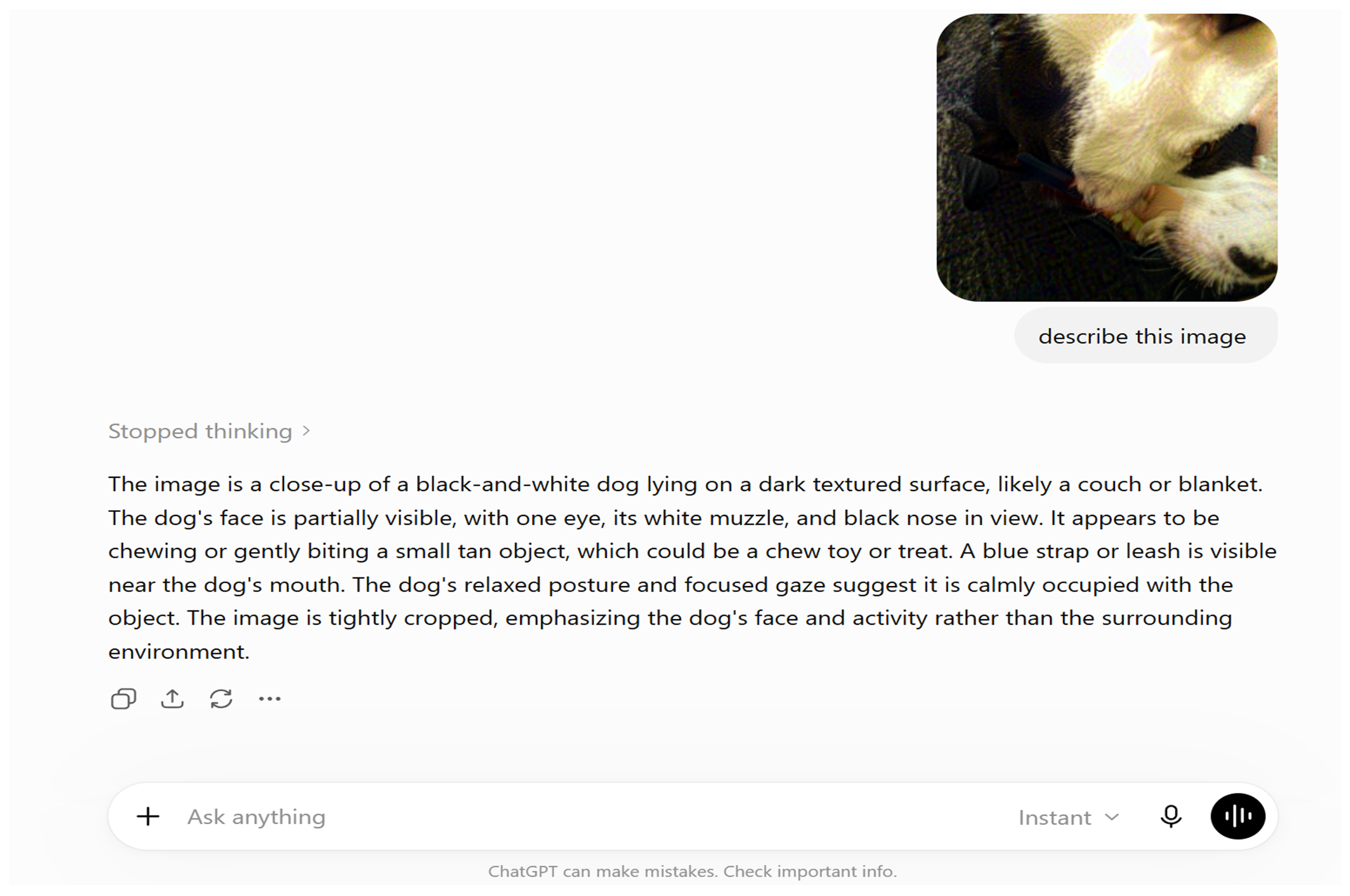}
    \caption{GPT-5.5 response.}
    \label{fig:qual1_gpt}
\end{subfigure}

\begin{subfigure}[t]{\columnwidth}
    \centering
    \includegraphics[width=\linewidth]
    {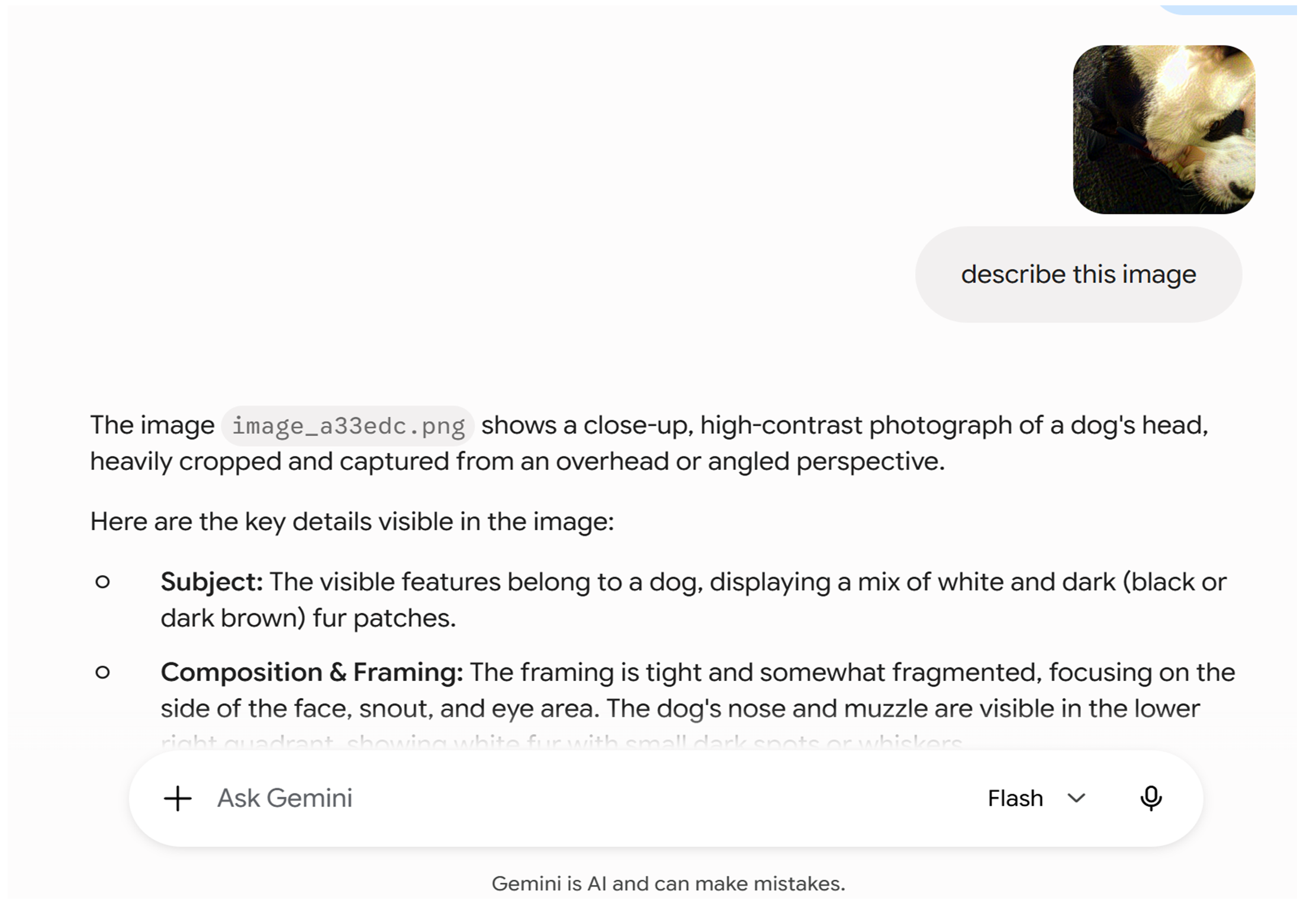}
    \caption{Gemini 3.5 response.}
    \label{fig:qual1_gemini}
\end{subfigure}
\vspace{10mm}
\caption{Qualitative example of targeted transfer. The source image shows a panda, while the target contains a dog. All three victim models describe the adversarial image as a dog rather than a panda.}
\label{fig:qual1}
\end{figure}

\begin{figure}[!t]
\centering

\begin{subfigure}[t]{\columnwidth}
    \centering
    \includegraphics[width=\linewidth,height=3.1cm]
    {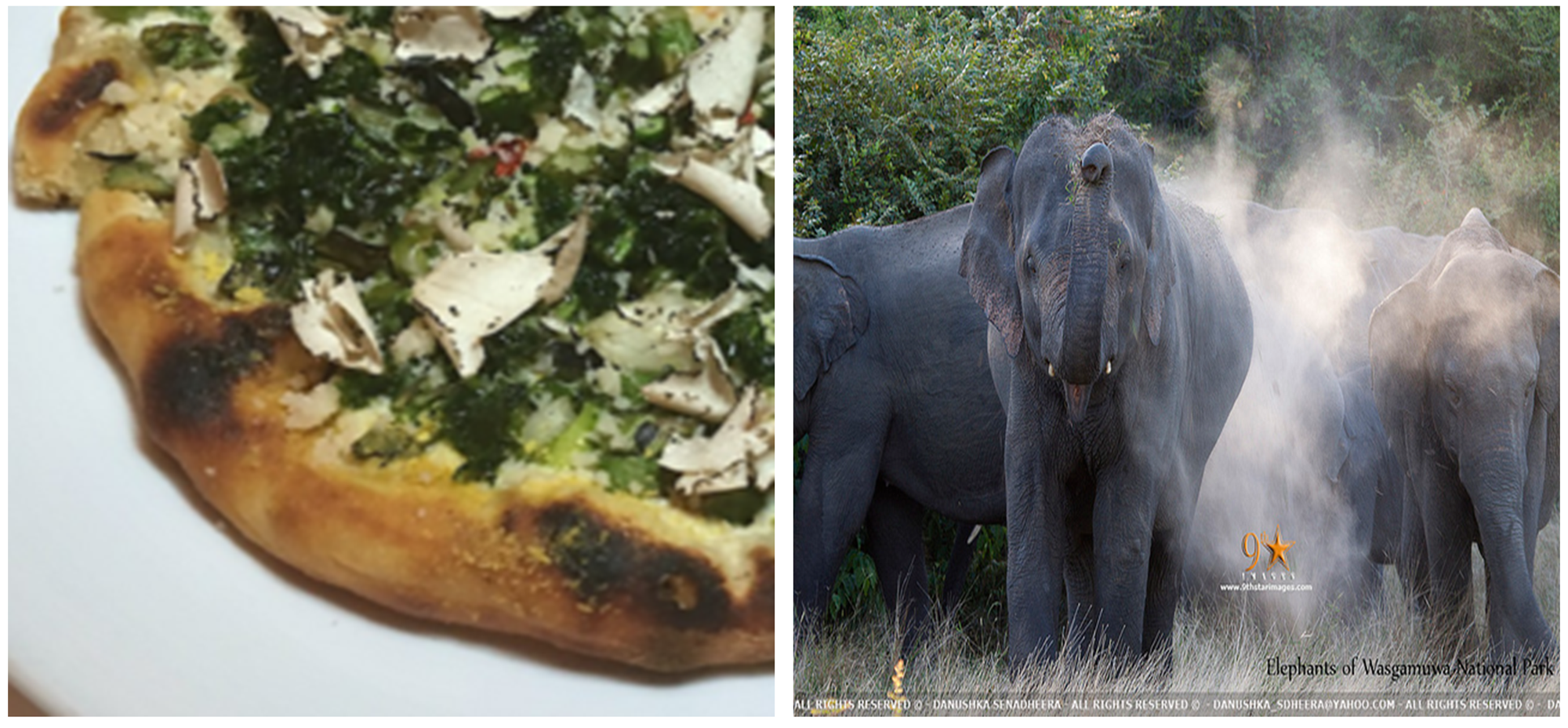}
    \caption{Source and target images.}
    \label{fig:qual5_source}
\end{subfigure}

\vspace{3pt}

\begin{subfigure}[t]{\columnwidth}
    \centering
    \includegraphics[width=\linewidth, height=5.25cm] 
    {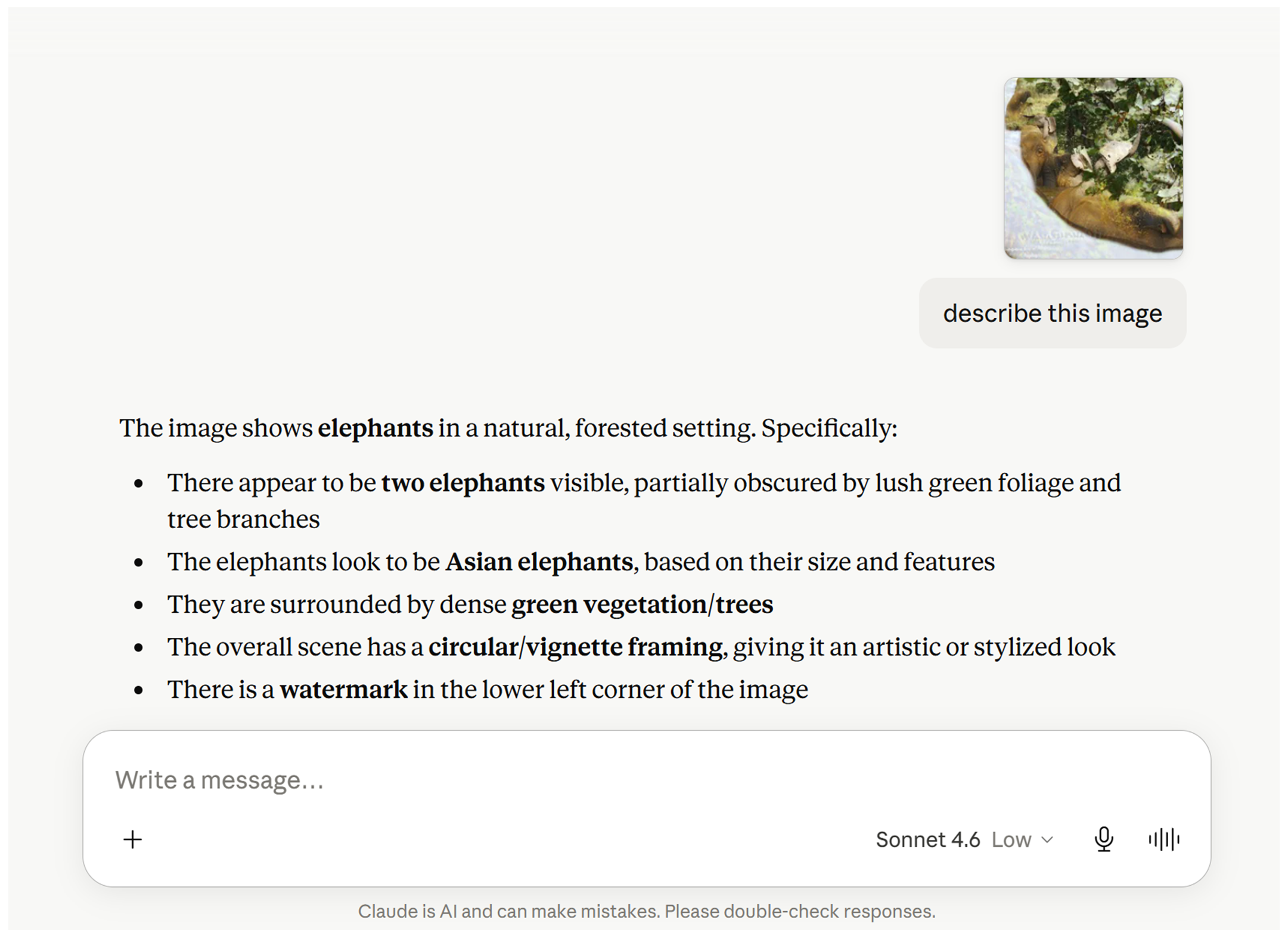}
    \caption{Claude Sonnet 4.6 response.}
    \label{fig:qual5_claude}
\end{subfigure}

\vspace{3pt}

\begin{subfigure}[t]{\columnwidth}
    \centering
    \includegraphics[width=\linewidth, height=5.25cm]
    {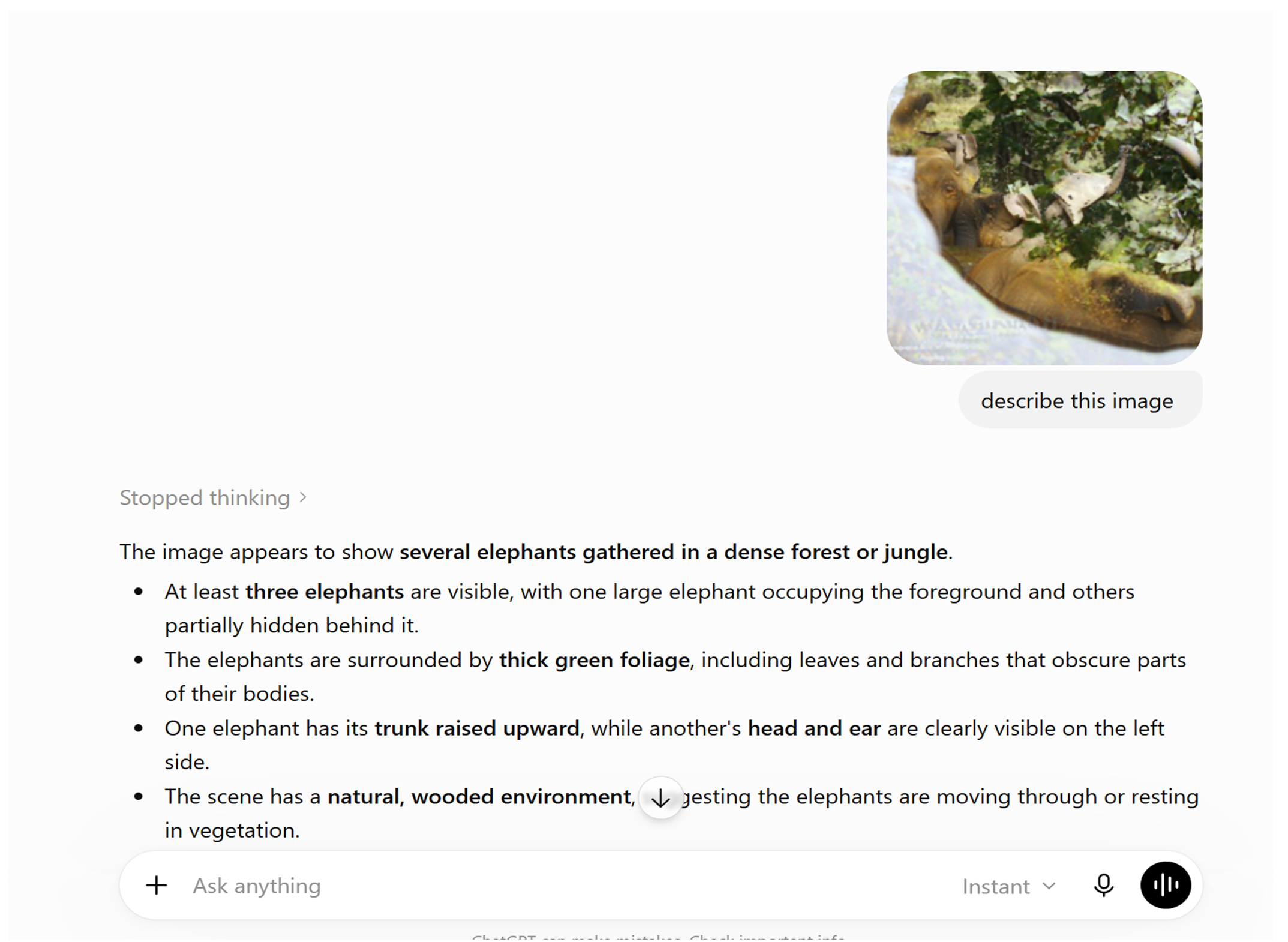}
    \caption{GPT-5.5 response.}
    \label{fig:qual5_gpt}
\end{subfigure}

\vspace{3pt}

\begin{subfigure}[t]{\columnwidth}
    \centering
    \includegraphics[width=\linewidth, height=5.25cm]
    {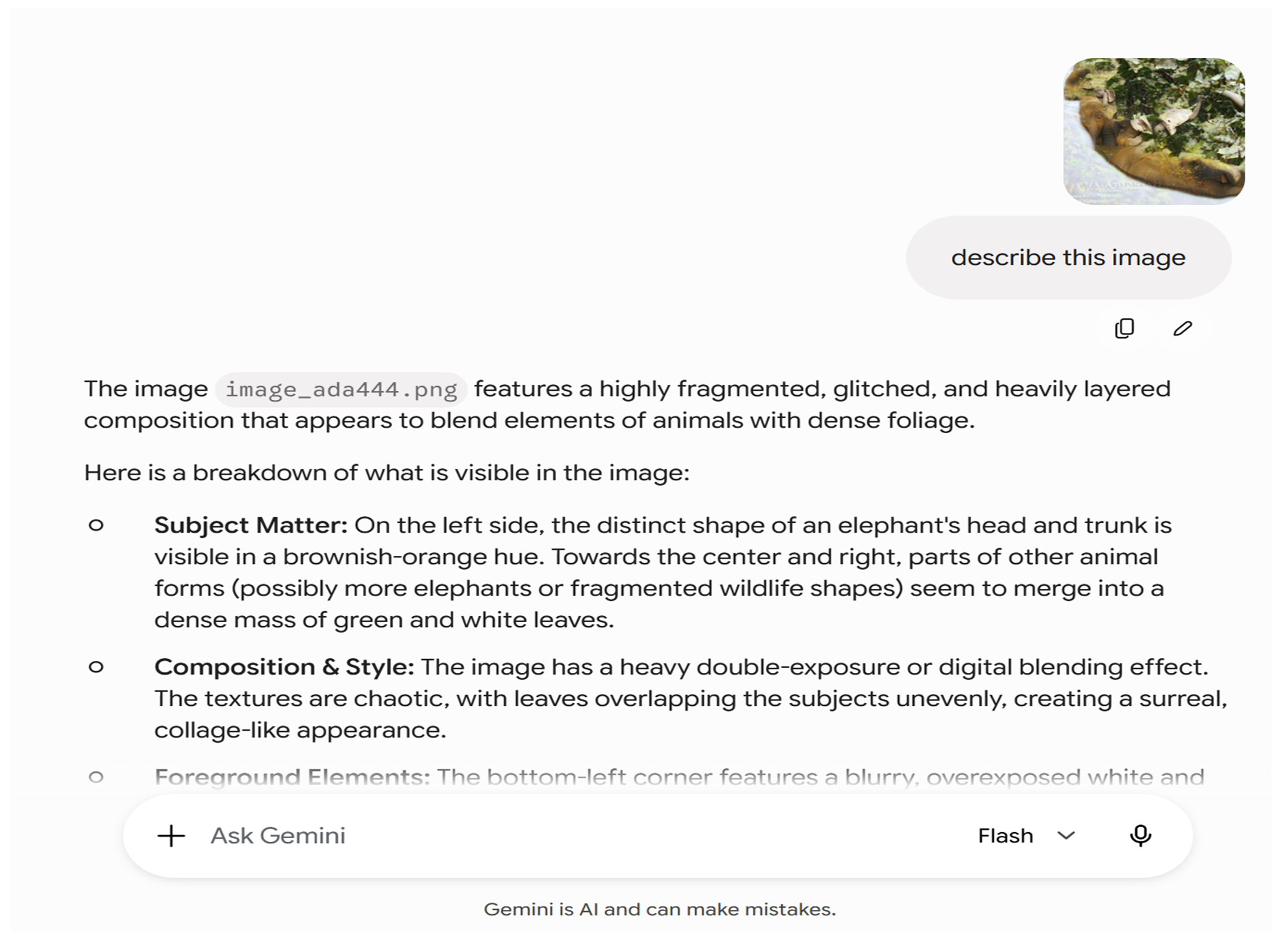}
    \caption{Gemini 3.5 response.}
    \label{fig:qual5_gemini}
\end{subfigure}

\caption{Qualitative example of targeted transfer. The source image shows a pizza, while the target contains elephants. All three victim models describe the adversarial image in terms of elephants and their natural environment rather than the source pizza.}
\label{fig:qual5}
\end{figure}

\begin{figure}[!t]
\centering

\begin{subfigure}[t]{\columnwidth}
    \centering
    \includegraphics[width=\linewidth, height=3.3cm]
    {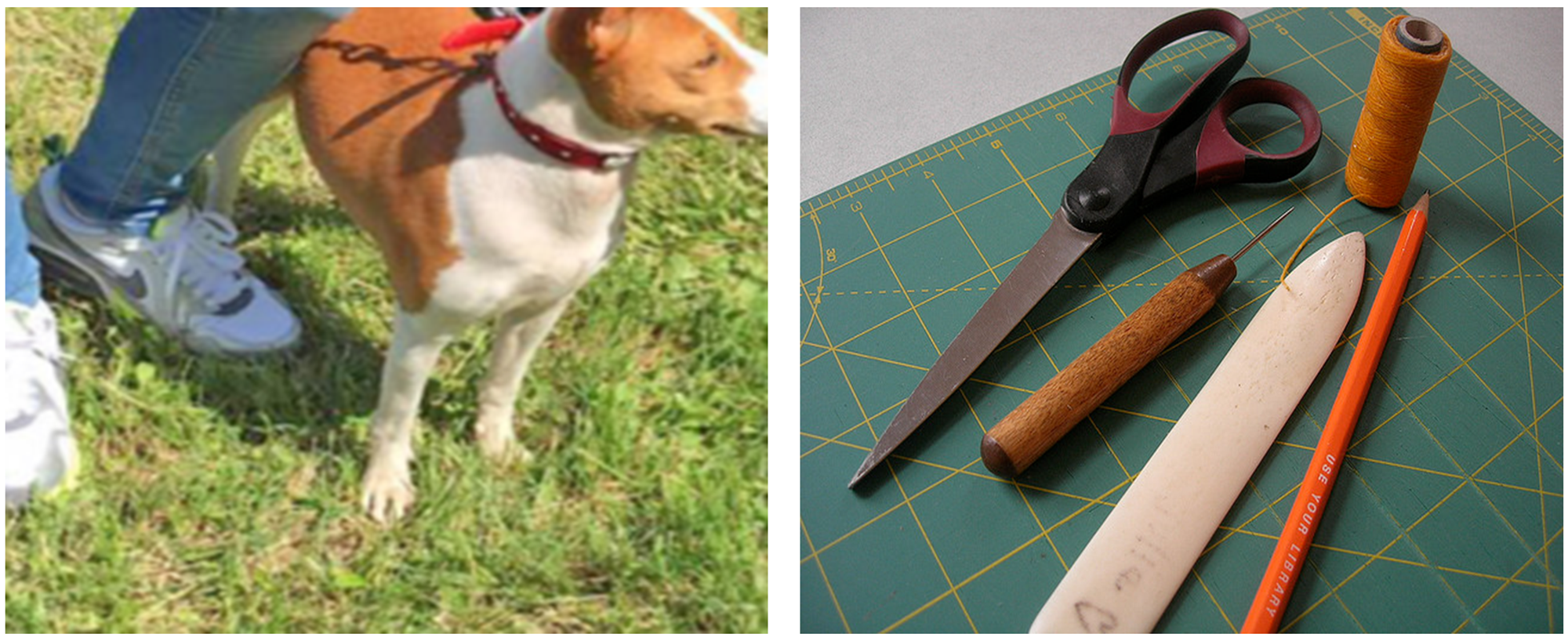}
    \caption{Source and target images.}
    \label{fig:qual2_source}
\end{subfigure}

\vspace{3pt}

\begin{subfigure}[t]{\columnwidth}
    \centering
    \includegraphics[width=\linewidth]
    {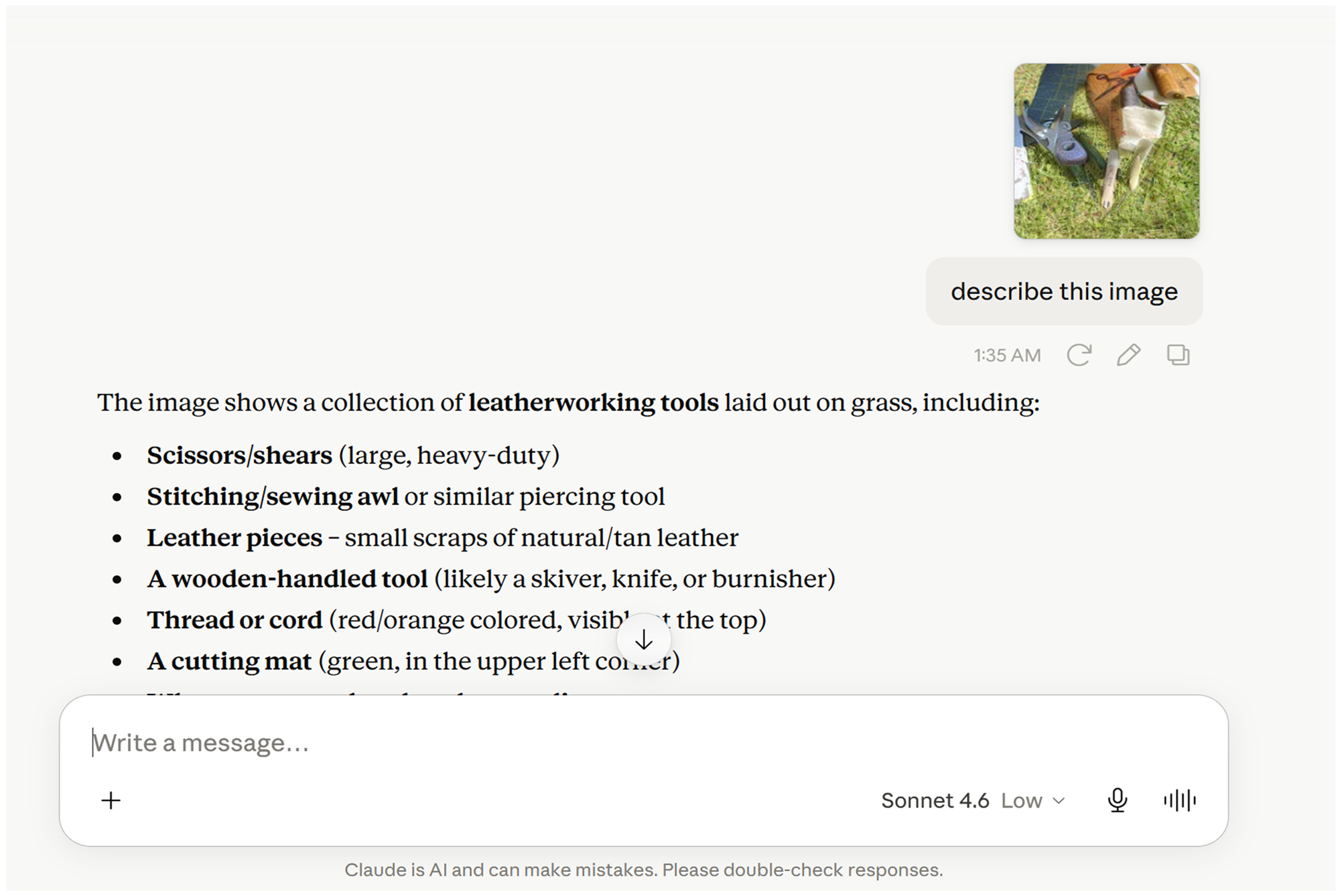}
    \caption{Claude Sonnet 4.6 response.}
    \label{fig:qual2_claude}
\end{subfigure}

\vspace{3pt}

\begin{subfigure}[t]{\columnwidth}
    \centering
    \includegraphics[width=\linewidth]
    {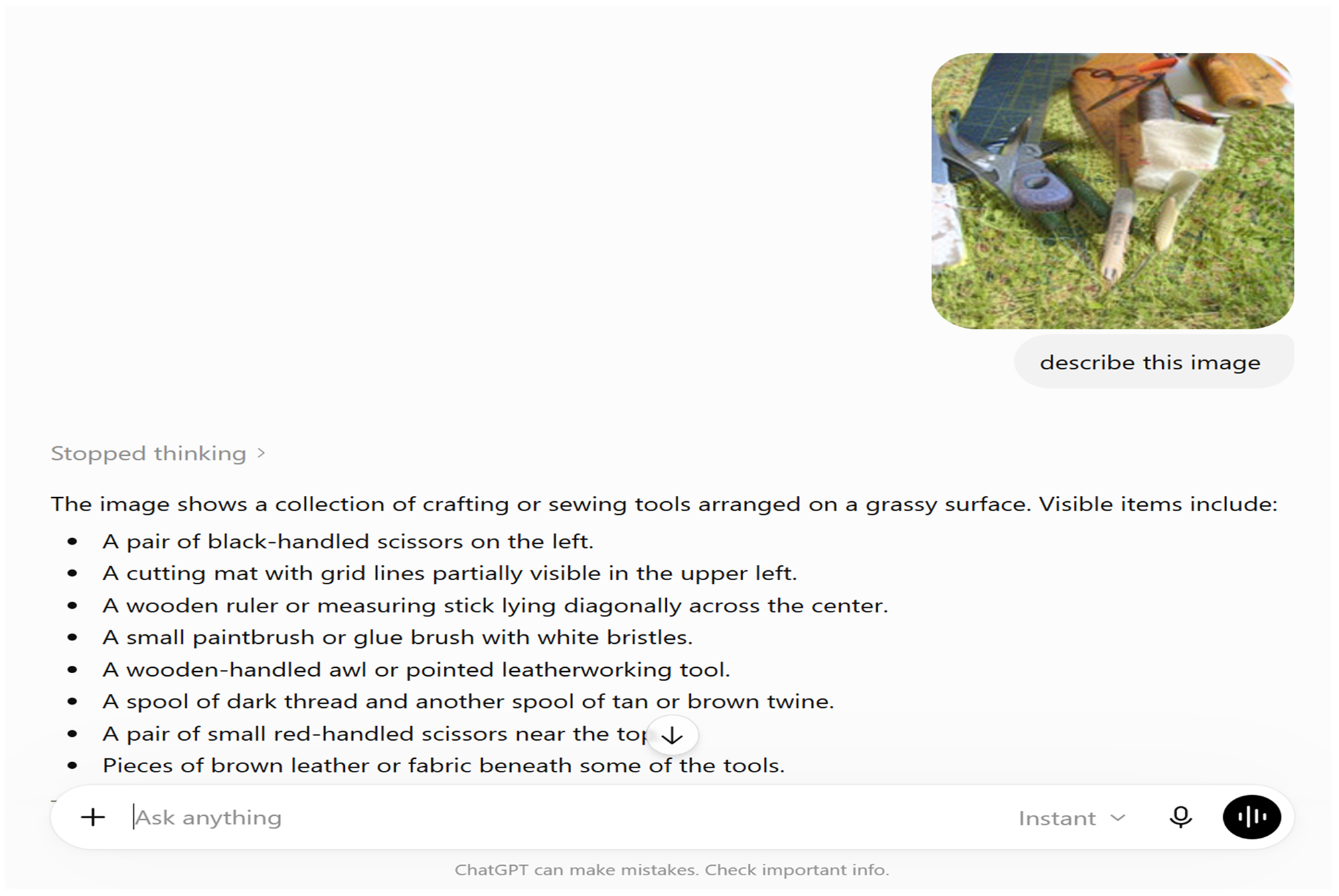}
    \caption{GPT-5.5 response.}
    \label{fig:qual2_gpt}
\end{subfigure}

\vspace{3pt}

\begin{subfigure}[t]{\columnwidth}
    \centering
    \includegraphics[width=\linewidth]
    {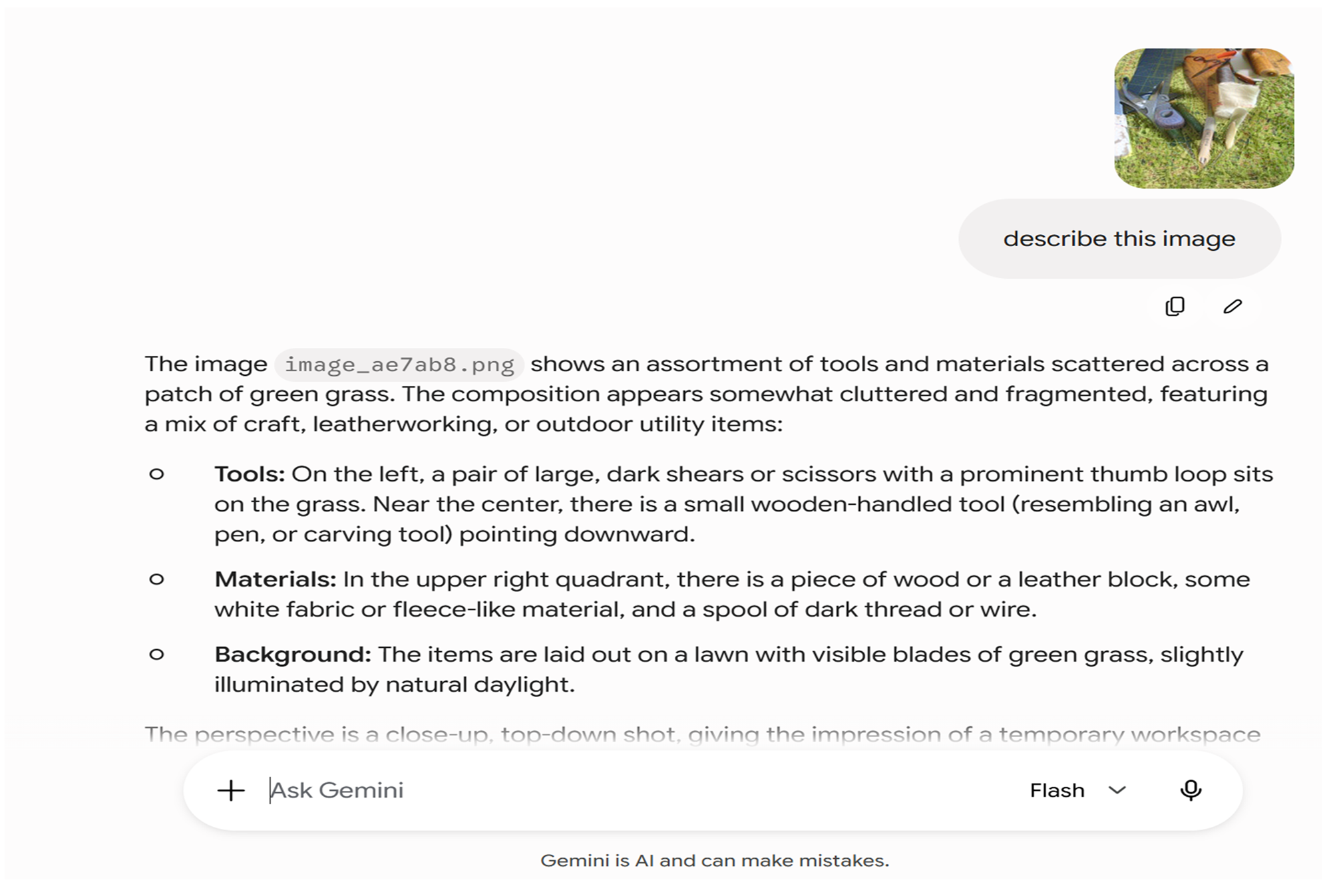}
    \caption{Gemini 3.5 response.}
    \label{fig:qual2_gemini}
\end{subfigure}

\caption{Qualitative example of targeted transfer. The source image shows a dog, while the target contains crafting and sewing tools. All three victim models describe the adversarial image in terms of these tools rather than the source dog.}
\label{fig:qual2}

\end{figure}

\begin{figure}[!t]
\centering

\begin{subfigure}[t]{\columnwidth}
    \centering
    \includegraphics[width=\linewidth, height=3.1cm]
    {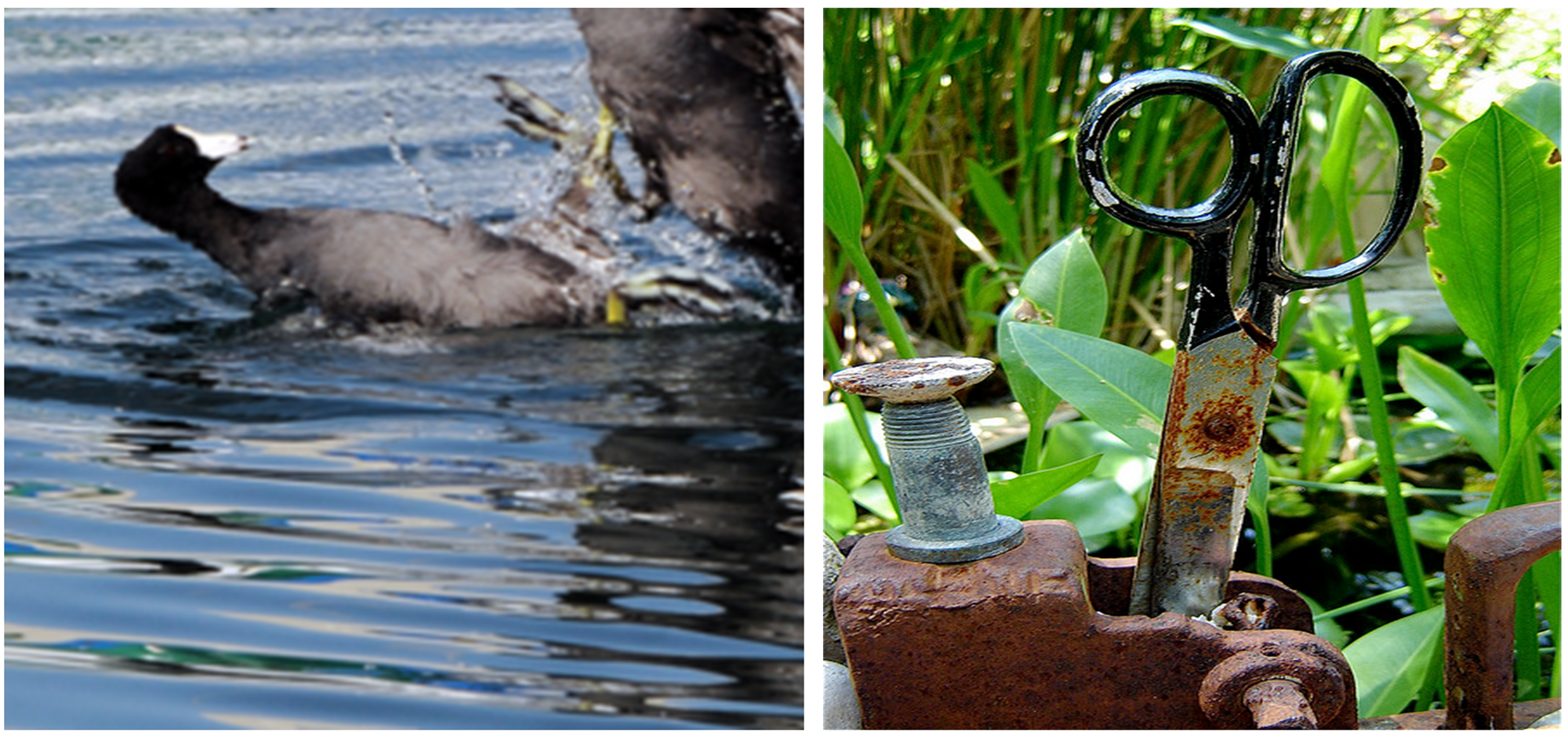}
    \caption{Source and target images.}
    \label{fig:qual3_source}
\end{subfigure}

\vspace{3pt}

\begin{subfigure}[t]{\columnwidth}
    \centering
    \includegraphics[width=\linewidth, height=5.2cm]
    {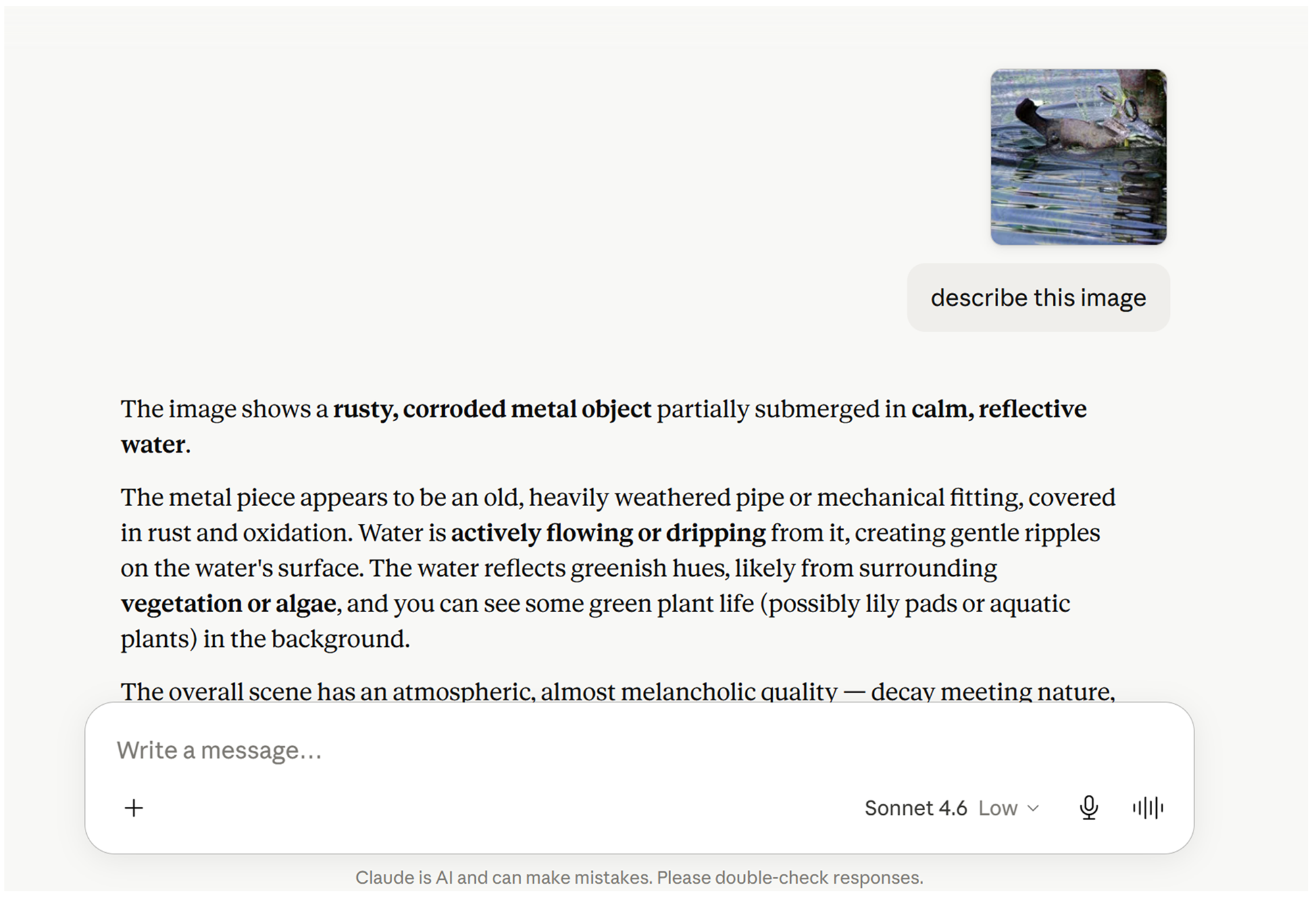}
    \caption{Claude Sonnet 4.6 response.}
    \label{fig:qual3_claude}
\end{subfigure}

\vspace{3pt}

\begin{subfigure}[t]{\columnwidth}
    \centering
    \includegraphics[width=\linewidth, height=5.2cm]
    {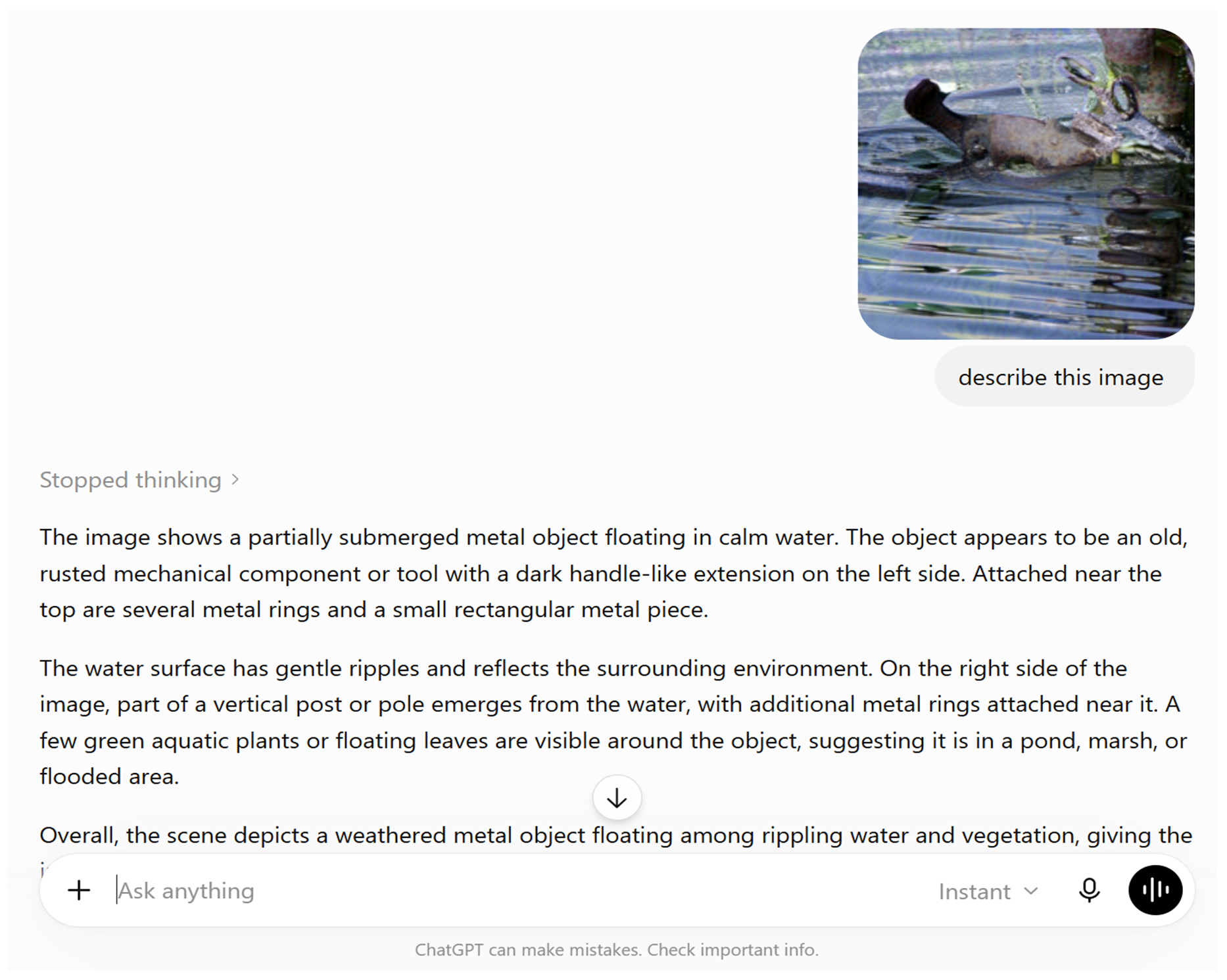}
    \caption{GPT-5.5 response.}
    \label{fig:qual3_gpt}
\end{subfigure}

\vspace{3pt}

\begin{subfigure}[t]{\columnwidth}
    \centering
    \includegraphics[width=\linewidth, height=5.2cm]
    {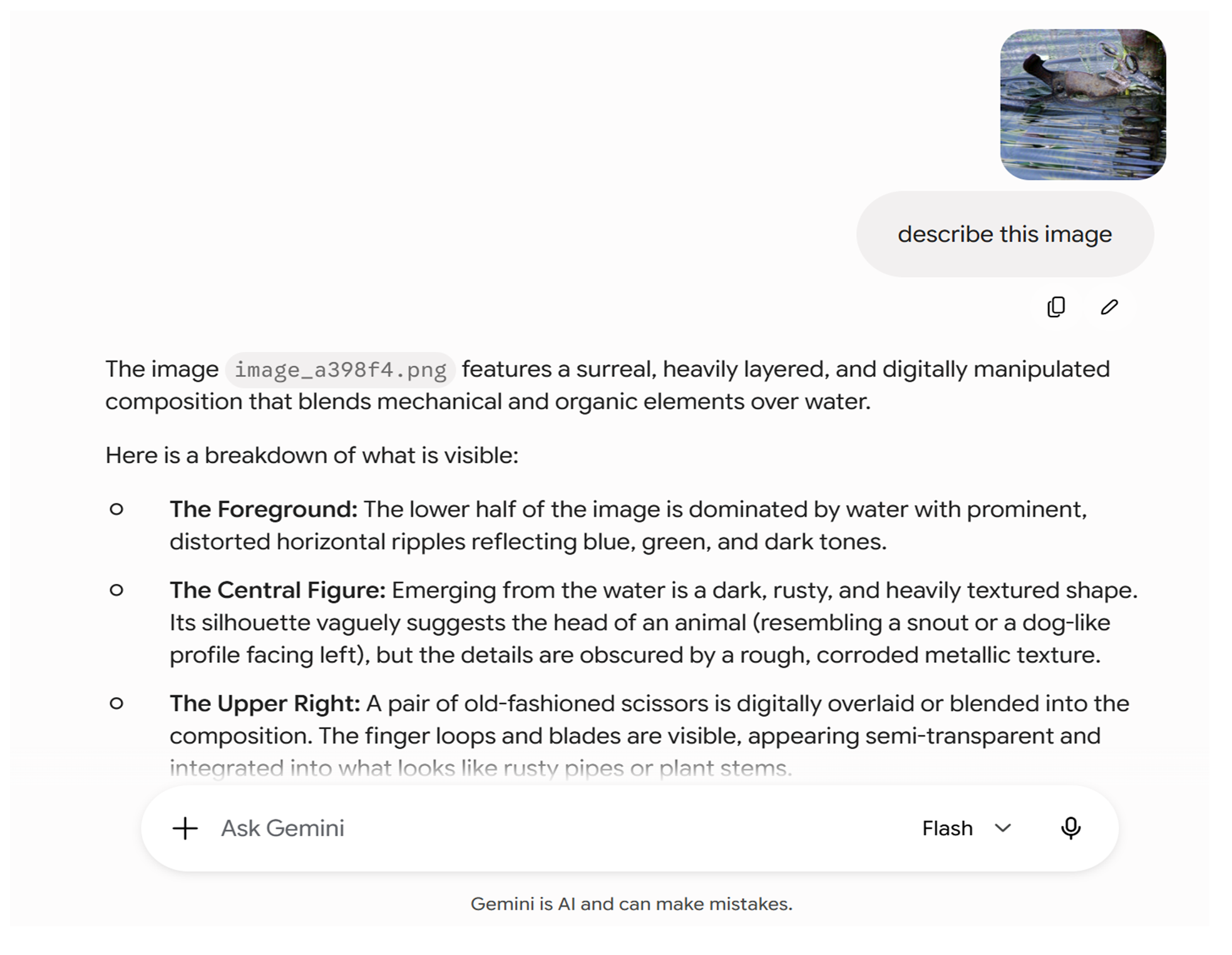}
    \caption{Gemini 3.5 response.}
    \label{fig:qual3_gemini}
\end{subfigure}
\vspace{5mm}
\caption{Qualitative example of targeted transfer. The source image shows a waterbird, while the target contains scissors. Victim models shift their descriptions toward metallic or scissor-like features rather than the source bird.}
\label{fig:qual3}

\end{figure}

\begin{figure}[!t]
\centering

\begin{subfigure}[t]{\columnwidth}
    \centering
    \includegraphics[width=\linewidth]
    {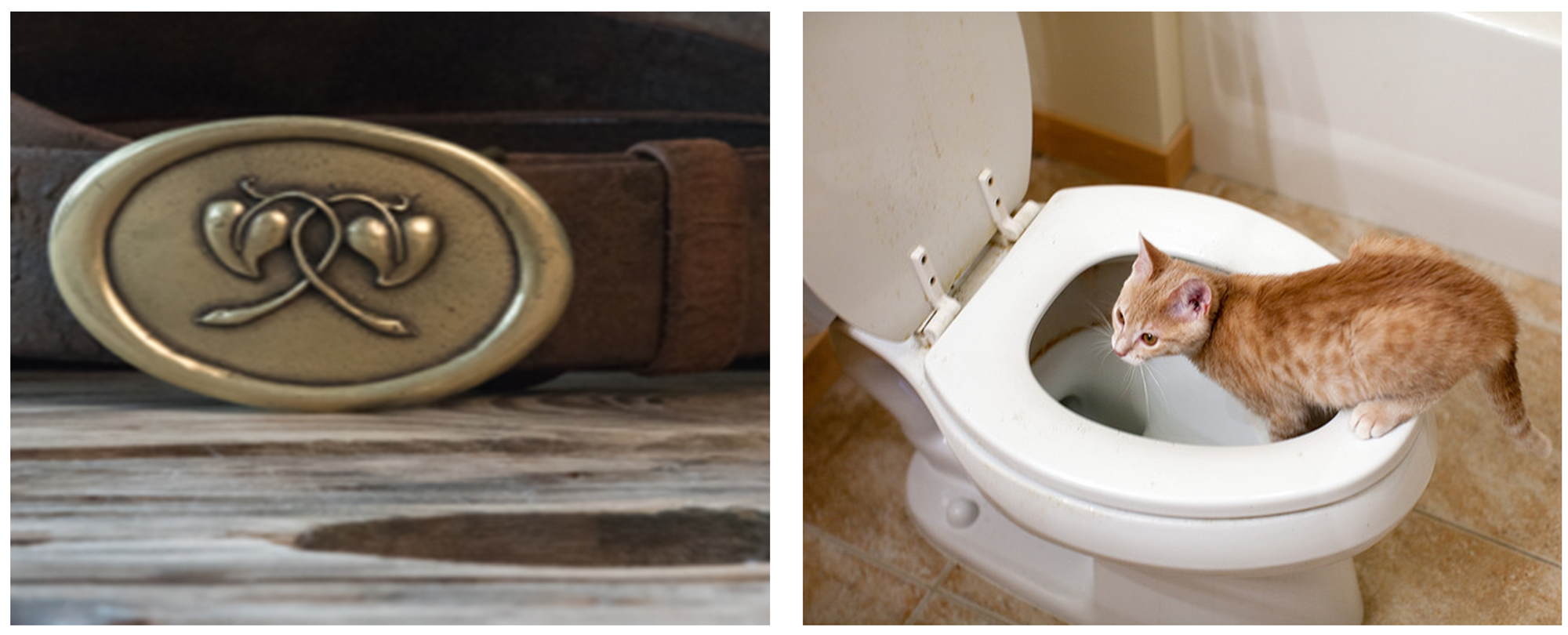}
    \caption{Source and target images.}
    \label{fig:qual4_source}
\end{subfigure}

\vspace{3pt}

\begin{subfigure}[t]{\columnwidth}
    \centering
    \includegraphics[width=\linewidth]
    {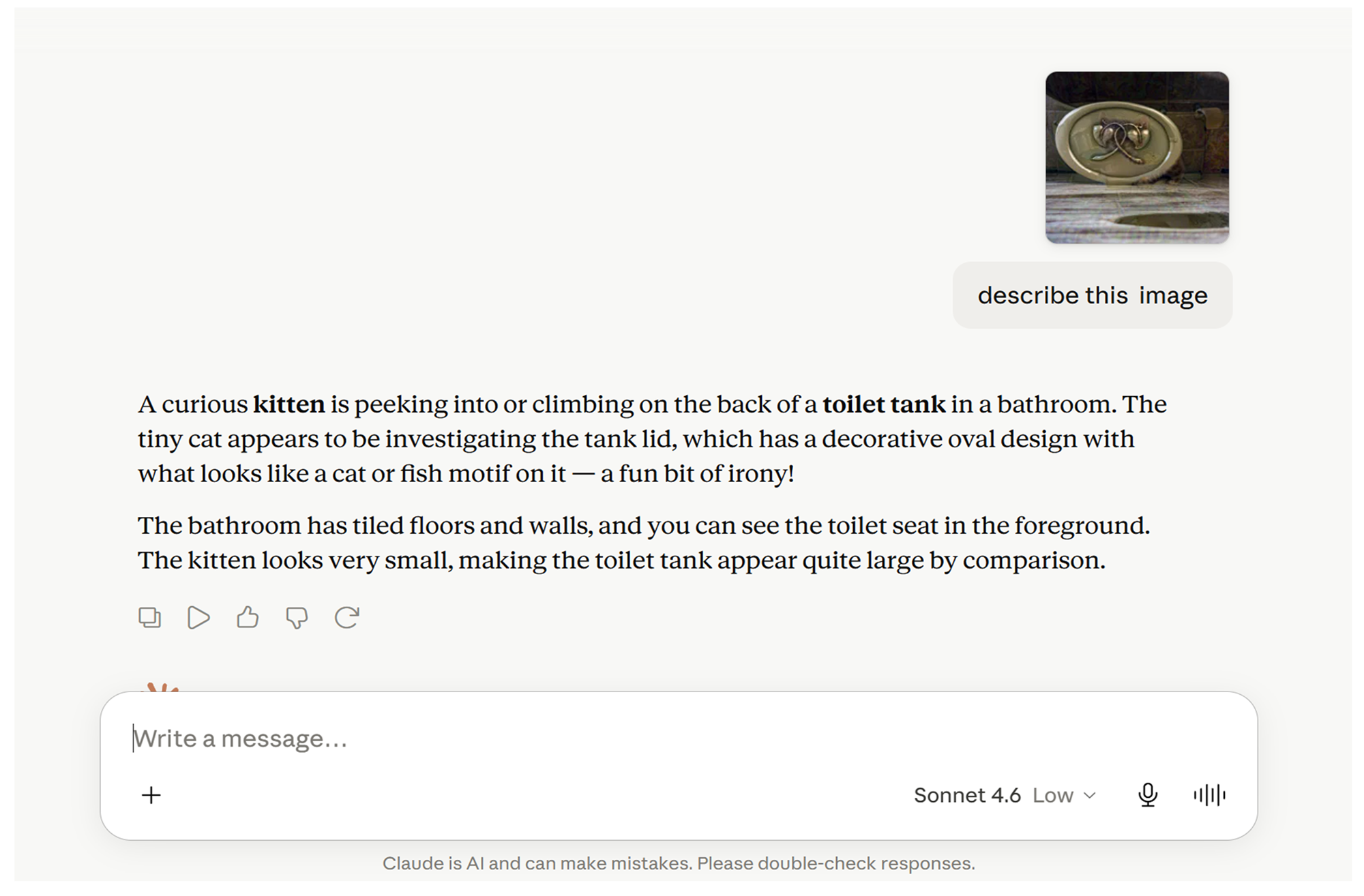}
    \caption{Claude Sonnet 4.6 response.}
    \label{fig:qual4_claude}
\end{subfigure}

\vspace{3pt}

\begin{subfigure}[t]{\columnwidth}
    \centering
    \includegraphics[width=\linewidth]
    {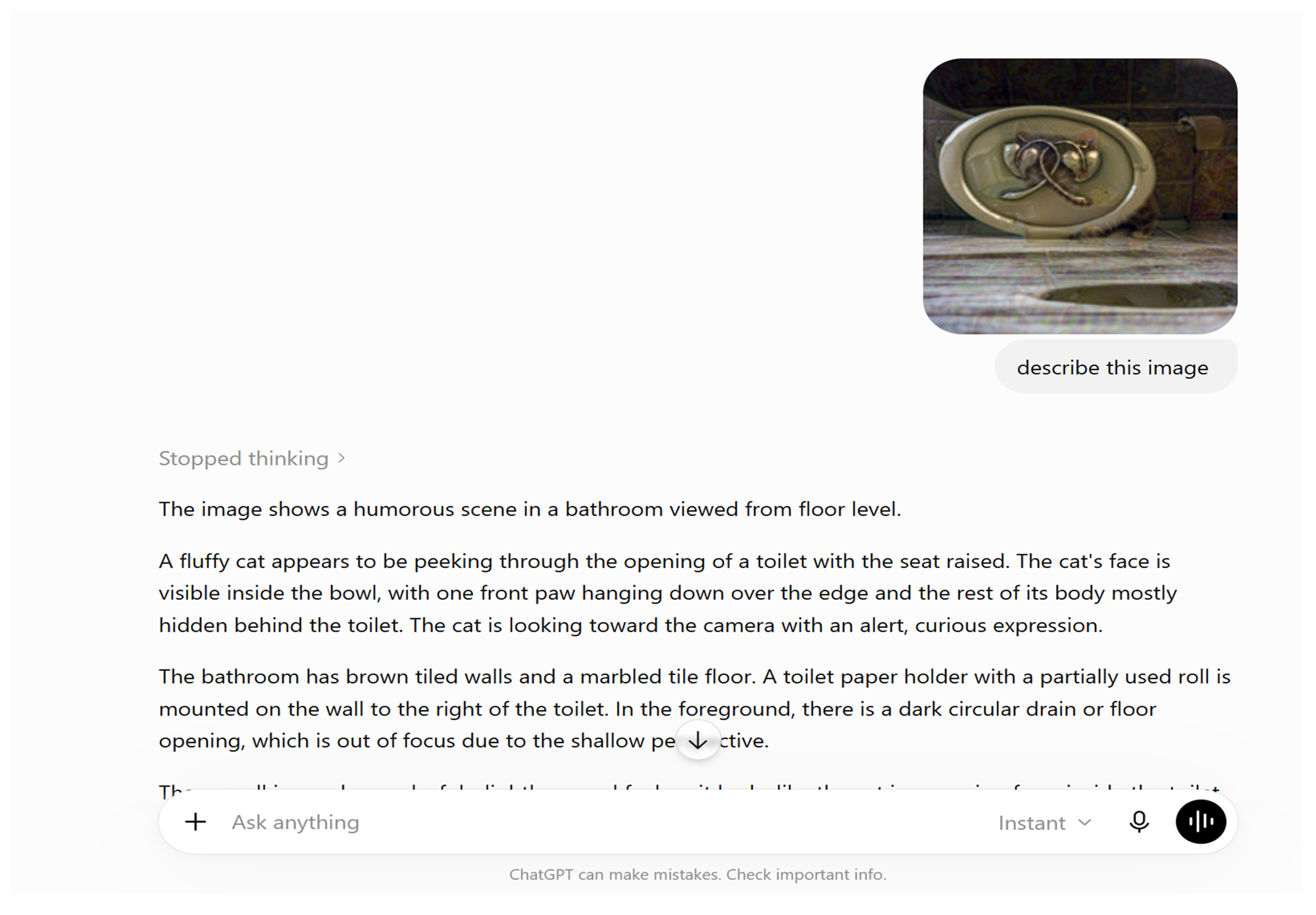}
    \caption{GPT-5.5 response.}
    \label{fig:qual4_gpt}
\end{subfigure}

\vspace{3pt}

\begin{subfigure}[t]{\columnwidth}
    \centering
    \includegraphics[width=\linewidth]
    {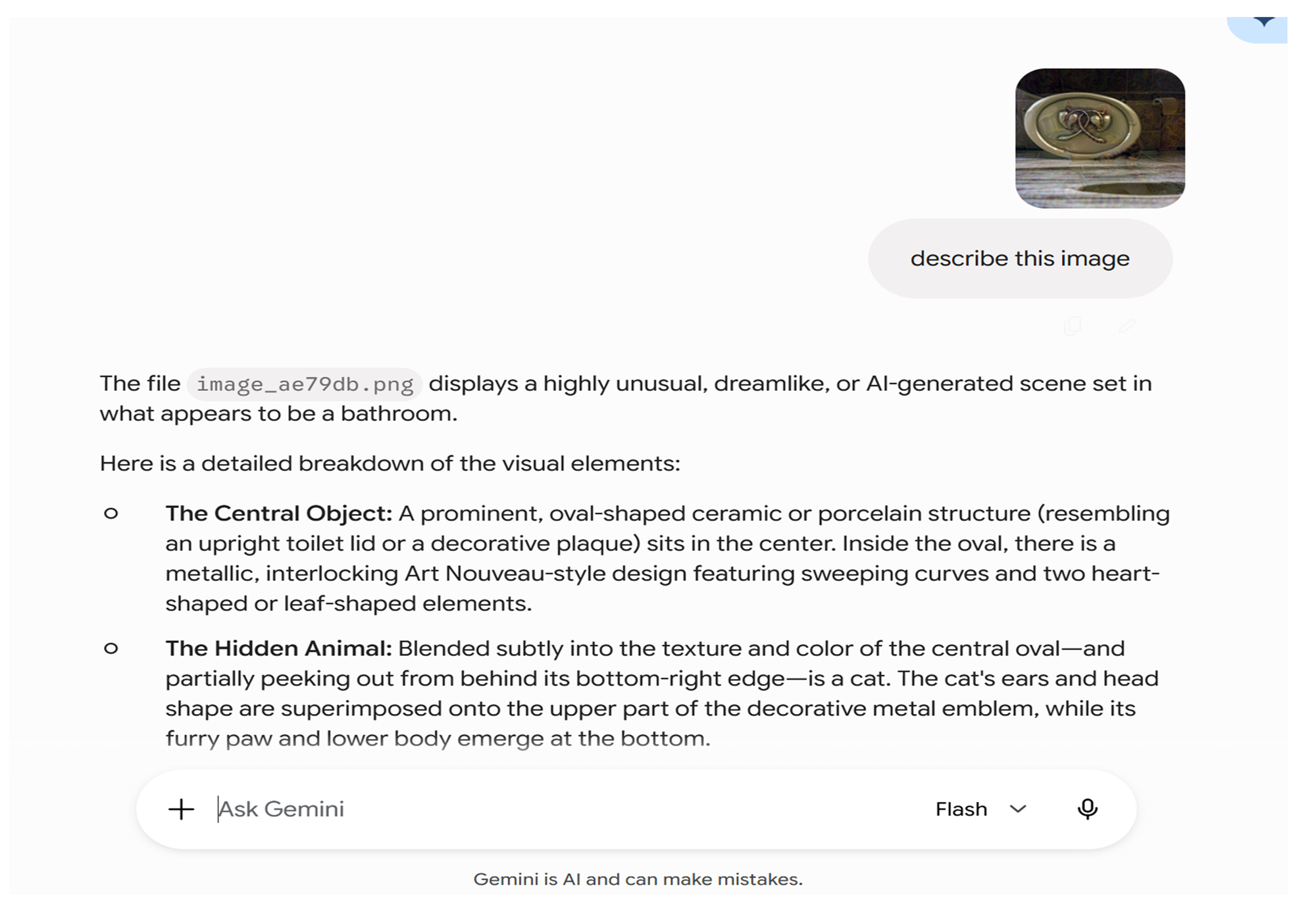}
    \caption{Gemini 3.5 response.}
    \label{fig:qual4_gemini}
\end{subfigure}

\caption{Qualitative example of targeted transfer. The source image shows a belt with a decorative buckle, while the target contains a cat near a toilet. Victim models shift their descriptions toward the target's cat and bathroom scene rather than the source buckle, illustrating the perturbation's effect.}
\label{fig:qual4}

\end{figure}

Figures~\ref{fig:qual1}--\ref{fig:qual4} provide additional real-world qualitative examples across five diverse source-target pairs, showing the actual responses of GPT-5, Gemini-3.5-Flash, and Claude Sonnet~4.6 when queried with VISTA adversarial examples through their standard browser interfaces, using the same $10\times$/128-zero noise convention stated in Sec.~\ref{subsec:noise_analysis}.




\end{document}